\documentclass{article}

\usepackage[final, nonatbib]{neurips_2021} %to compile a preprint version, e.g., for submission to
\pdfoutput=1

\usepackage[utf8]{inputenc} % allow utf-8 input
\usepackage[T1]{fontenc}    % use 8-bit T1 fonts
\usepackage{hyperref}       % hyperlinks
\usepackage{url}            % simple URL typesetting
\usepackage{booktabs}       % professional-quality tables
\usepackage{amsfonts}       % blackboard math symbols
\usepackage{nicefrac}       % compact symbols for 1/2, etc.
\usepackage{microtype}      % microtypography
\usepackage{xcolor}         % colors
\usepackage{textcomp}

\usepackage{hyperref}
\usepackage{rotating}
\usepackage{titlesec}
\usepackage{amsmath}
\usepackage{bbm}
\usepackage{amsthm}
\usepackage{amssymb}
\usepackage{mathtools}
\usepackage{xspace}
\usepackage{enumitem}
\usepackage[hang,flushmargin]{footmisc}
\usepackage{algorithm}% http://ctan.org/pkg/algorithm
\usepackage{algpseudocode}% http://ctan.org/pkg/algorithmicx
\usepackage{subfigure}
\usepackage[
hypcap=false,
width=0.9\textwidth,
skip=0pt,
]{caption}
\setlist{leftmargin=*}
\setlist[itemize]{noitemsep}
\setlist[enumerate]{noitemsep}
\newcommand{\methodname}{LVD\xspace}
\newcommand{\embedder}{\mathbf{f}}
\newcommand{\validationset}{\mathcal{S}_{\text{conformal}}} 
\newcommand{\fullset}{\mathcal{S}_{\text{train}}} 
\newcommand{\trainingset}{\mathcal{S}_{\text{embed}}} 
\newcommand{\defeq}{\vcentcolon=}
\DeclarePairedDelimiter\ceil{\lceil}{\rceil}
\DeclarePairedDelimiter\floor{\lfloor}{\rfloor}

\newtheorem{theorem}{Theorem}[section]

\newtheorem{definition}{Definition}

\usepackage{xcolor}

\def \AlgoMain{
\begin{algorithm}[htb]
   \caption{\methodname}
   \label{alg:main}
\textbf{Input}:\\
    $\fullset$: A set of observations $\{Z_i=(X_i,Y_i)\}_{i=1}^N$\\
    $\alpha$: Parameter specifying (local) target coverage rate\\
    $X_{N+1}$: Unseen data point\\
\textbf{Output}: 
    A locally valid PI, $\hat{C}_{\alpha}(X_{N+1})$. \\
\textbf{Training}:
\begin{algorithmic}
    \State [Optional] Randomly split $\fullset$ into $\trainingset$ and $\validationset$. Denote $\validationset$ as $\{Z_{n+i}\}_{i=1}^{m}$
    \State [Optional] Train a NN regression model $\hat{\mu}^{NN}$ on $\trainingset$. 
    \State Remove the last layer of $\hat{\mu}^{NN}$ to get an embedding function $\embedder$
    \State Train $\mathbf{A}$ on $\embedder(\trainingset)$ in a Nadaraya-Watson kernel regression setting, with kernel $K_\embedder(x_1,x_2) = e^{-||\mathbf{A}(\embedder(x_1)-\embedder(x_2))||^2}$
    \State Collect residuals $R_i = |y_{n+i}-\hat{y}_{n+i}|$ for $i\in[m]$
\end{algorithmic}
\textbf{Inference}:
\begin{algorithmic}
    \State Compute PI as $\displaystyle \hat{C}_{\alpha}(X_{N+1}) = \left\{y\in\mathbb{R}: |y - \hat{y}_{N+1}| \leq Q\left(1-\alpha, w_{N+1}\delta_{\infty} + \sum_{i=1}^m w_{n+i} \delta_{R_i}\right)\right\}$
    \State where $\displaystyle w_j = \frac{K_\embedder(x_{j}, x_{N+1})}{K_\embedder(x_{N+1}, x_{N+1})+\sum_{i=1}^m K_\embedder(x_{n+i}, x_{N+1})}$
\end{algorithmic}
\end{algorithm} 
}

\def \TableCounts{
\begin{table}[ht]
\vspace{-15pt}
\captionof{table}{
Size of each dataset. Size of the test set is in parenthesis. 
}
\begin{center}
\begin{footnotesize}
\setlength\tabcolsep{2pt}
\begin{tabular}{lcccccccr}
\toprule
Yacht & Housing  & Energy & Bike   &  Kin8nm        &  Concrete &  QM8 &  QM9\\
\midrule
308 (62)  & 506 (101) & 768 (154) & 17379(3476) & 8192 (1638) & 1030 (206) & 21786 (4357) & 133719 (26744)\\
\bottomrule
\end{tabular}
\label{table:exp:real:counts}
\end{footnotesize}
\end{center}
\vspace{-5pt}
\end{table}
}

\def \TableCoverage{
\begin{table}[ht]
\vspace{-10pt}
\captionof{table}{
Marginal coverage rate (MCR) and tail coverage rate (TCR) (coverage rate for left and right 10\% tail for test label) with target at 90\%.
``\textendash'' represents not-applicable models (see Section \ref{sec:exp:real}).
Coverage rates not significantly lower than target at $p=0.05$ are in bold (good).
Note that the too high is not better. For example, MCDP either \textit{greatly} over- or under-covers with MCR either 100\% or well below 90\%. 
}
\begin{center}
\begin{scriptsize}
\setlength\tabcolsep{6pt}
\begin{tabular}{l|cccccccr}
\toprule
MCR        & \methodname & MADSplit   & CQR        & DJ         & DE         & MCDP       & PBP       \\
\midrule
Yacht      & \textbf{96.8}$\pm$2.2 & 82.4$\pm$7.1 & \textbf{91.5}$\pm$4.7 & \textbf{95.0}$\pm$2.1 & 22.7$\pm$6.0 & 87.4$\pm$4.2 & 80.2$\pm$10.8\\
Housing    & \textbf{96.8}$\pm$2.9 & \textbf{90.6}$\pm$3.5 & \textbf{91.7}$\pm$3.3 & \textbf{97.6}$\pm$1.4 & \textbf{96.0}$\pm$1.9 & \textbf{100.0}$\pm$0.0 & 8.1$\pm$4.3\\
Energy     & \textbf{94.0}$\pm$1.6 & \textbf{90.3}$\pm$2.5 & \textbf{90.3}$\pm$2.2 & \textbf{96.2}$\pm$1.8 & \textbf{98.0}$\pm$2.7 & \textbf{100.0}$\pm$0.0 & 7.2$\pm$5.9\\
Bike       & \textbf{90.4}$\pm$0.8 & \textbf{89.9}$\pm$0.6 & \textbf{89.8}$\pm$0.7 & \textbf{95.2}$\pm$0.6 & \textbf{100.0}$\pm$0.0 & 71.9$\pm$0.7 & 0.6$\pm$0.2\\
Kin8nm     & \textbf{98.0}$\pm$0.6 & \textbf{90.0}$\pm$0.8 & \textbf{90.2}$\pm$0.6 & \textbf{94.7}$\pm$0.4 & \textbf{100.0}$\pm$0.1 & \textbf{100.0}$\pm$0.0 & \textbf{100.0}$\pm$0.0\\
Concrete   & \textbf{97.4}$\pm$1.3 & \textbf{88.8}$\pm$2.8 & 88.5$\pm$2.3 & \textbf{98.0}$\pm$1.6 & \textbf{97.8}$\pm$1.0 & \textbf{100.0}$\pm$0.0 & 3.3$\pm$0.8\\
QM8* & \textbf{92.6}$\pm$0.9 & \textbf{90.0}$\pm$0.7 & \textbf{90.0}$\pm$0.6 & \textendash & \textbf{100.0}$\pm$0.0 & \textendash & \textendash\\
QM9* & \textbf{90.3}$\pm$0.6 & \textbf{90.0}$\pm$0.2 & \textbf{90.0}$\pm$0.3 & \textendash & 60.7$\pm$46.8 & \textendash & \textendash\\
\bottomrule

\toprule
TCR        & \methodname & MADSplit   & CQR        & DJ         & DE         & MCDP       & PBP       \\
\midrule
Yacht      & \textbf{98.5}$\pm$3.2 & 65.4$\pm$23.8 & 77.7$\pm$12.3 & 76.2$\pm$9.9 & 1.5$\pm$4.9 & 50.0$\pm$9.8 & 70.0$\pm$14.3\\
Housing    & \textbf{96.2}$\pm$4.4 & \textbf{87.6}$\pm$8.8 & 82.9$\pm$8.2 & \textbf{90.0}$\pm$5.2 & 81.9$\pm$9.7 & \textbf{100.0}$\pm$0.0 & 1.0$\pm$3.0\\
Energy     & \textbf{86.8}$\pm$5.8 & 78.4$\pm$10.9 & 73.5$\pm$12.0 & \textbf{90.0}$\pm$6.5 & \textbf{95.8}$\pm$6.3 & \textbf{100.0}$\pm$0.0 & 9.7$\pm$12.7\\
Bike       & \textbf{90.2}$\pm$1.7 & \textbf{89.2}$\pm$3.5 & 58.7$\pm$7.3 & 85.6$\pm$3.3 & \textbf{100.0}$\pm$0.0 & 49.9$\pm$0.0 & 0.0$\pm$0.0\\
Kin8nm     & \textbf{97.2}$\pm$1.6 & 86.4$\pm$2.6 & 85.2$\pm$2.2 & 88.1$\pm$1.8 & \textbf{99.9}$\pm$0.3 & \textbf{100.0}$\pm$0.0 & \textbf{100.0}$\pm$0.0\\
Concrete   & \textbf{97.1}$\pm$3.4 & 83.9$\pm$7.3 & 85.4$\pm$6.2 & \textbf{95.6}$\pm$3.6 & \textbf{91.7}$\pm$4.8 & \textbf{100.0}$\pm$0.0 & 3.4$\pm$5.7\\
QM8* & \textbf{90.8}$\pm$1.9 & 86.3$\pm$2.4 & 80.0$\pm$5.9 & \textendash & \textbf{100.0}$\pm$0.0 & \textendash & \textendash\\
QM9* & \textbf{89.7}$\pm$2.5 & 86.1$\pm$3.0 & 79.7$\pm$8.9 & \textendash & 60.3$\pm$46.5 & \textendash & \textendash\\
\bottomrule
\end{tabular}
\label{table:exp:real:coverage}
\end{scriptsize}
\end{center}
\vspace{-10pt}
\end{table}
}

\def \TableDisc{
\begin{table}[ht]
\vspace{-10pt}
\captionof{table}{
At $p=0.05$, AUROCs (in predicting error being greater than 50\% percentile) that are significantly higher than 50\%, and mean absolute deviations (MAD) significantly lower than the second-best, are in bold.
\methodname, MADSplit and CQR are consistently discriminative, but CQR sometimes incurs high MAD.
DJ is not discriminative, whereas other methods occasionally demonstrate discrimination but usually have high MADs as well.
}
\begin{center}
\begin{scriptsize}
\setlength\tabcolsep{3pt}
\begin{tabular}{l|cccccccr}
\toprule
AUROC      & \methodname & MADSplit   & CQR        & DJ         & DE         & MCDP       & PBP       \\
\midrule
Yacht      & \textbf{83.5}$\pm$5.8 & \textbf{77.7}$\pm$9.0 & \textbf{84.9}$\pm$4.6 & 50.0$\pm$10.5 & \textbf{59.8}$\pm$6.4 & 47.2$\pm$7.7 & \textbf{82.8}$\pm$8.8\\
Housing    & \textbf{59.2}$\pm$8.5 & \textbf{62.0}$\pm$8.3 & \textbf{62.5}$\pm$6.7 & 49.6$\pm$5.9 & \textbf{60.0}$\pm$6.8 & 42.5$\pm$7.8 & 47.4$\pm$3.6\\
Energy     & \textbf{73.5}$\pm$6.3 & \textbf{72.9}$\pm$5.6 & \textbf{72.1}$\pm$8.2 & \textbf{57.5}$\pm$8.1 & 56.1$\pm$11.0 & \textbf{54.6}$\pm$5.5 & 48.2$\pm$2.6\\
Bike       & \textbf{68.2}$\pm$11.0 & \textbf{71.7}$\pm$8.5 & \textbf{84.8}$\pm$33.5 & 45.8$\pm$6.2 & \textbf{86.2}$\pm$12.5 & \textbf{94.3}$\pm$1.0 & 48.3$\pm$1.0\\
Kin8nm     & \textbf{60.3}$\pm$1.1 & \textbf{60.4}$\pm$1.9 & \textbf{60.0}$\pm$2.1 & 49.3$\pm$2.2 & 50.5$\pm$2.6 & \textbf{54.1}$\pm$2.5 & \textbf{53.6}$\pm$4.8\\
Concrete   & \textbf{64.0}$\pm$6.1 & \textbf{63.8}$\pm$5.7 & \textbf{66.0}$\pm$7.1 & 46.2$\pm$4.9 & \textbf{55.9}$\pm$6.2 & 51.9$\pm$3.5 & 49.7$\pm$3.9\\
QM8* & \textbf{71.3}$\pm$9.4 & \textbf{73.5}$\pm$6.8 & \textbf{65.5}$\pm$10.3 & \textendash & \textbf{91.7}$\pm$16.9 & \textendash & \textendash\\
QM9* & \textbf{62.7}$\pm$3.6 & \textbf{64.9}$\pm$3.5 & \textbf{55.0}$\pm$14.4 & \textendash & \textbf{56.8}$\pm$28.4 & \textendash & \textendash\\
\bottomrule

\toprule
MAD        & \methodname & MADSplit   & CQR        & DJ         & DE         & MCDP       & PBP       \\
\midrule
Yacht      & 1.90$\pm$0.48 & 1.90$\pm$0.48 & 3.55$\pm$0.85 & 10.15$\pm$0.84 & 11.25$\pm$0.81 & 10.92$\pm$0.73 & 1.80$\pm$0.30\\
Housing    & 3.31$\pm$0.53 & 3.31$\pm$0.53 & 3.44$\pm$0.33 & 3.69$\pm$0.33 & 4.42$\pm$0.39 & 6.04$\pm$0.54 & 7.94$\pm$1.97\\
Energy     & 2.99$\pm$0.75 & 2.99$\pm$0.75 & 3.44$\pm$1.04 & 3.19$\pm$0.51 & 3.79$\pm$0.31 & 8.12$\pm$0.59 & 11.66$\pm$2.24\\
Bike       & 0.04$\pm$0.03 & 0.04$\pm$0.03 & 7.34$\pm$3.51 & 0.05$\pm$0.03 & 3.37$\pm$2.90 & 124.57$\pm$2.68 & 162.21$\pm$2.58\\
Kin8nm     & \textbf{0.07}$\pm$0.00 & \textbf{0.07}$\pm$0.00 & 0.08$\pm$0.01 & 0.09$\pm$0.01 & 0.19$\pm$0.01 & 0.18$\pm$0.00 & 0.22$\pm$0.12\\
Concrete   & 5.44$\pm$0.53 & 5.44$\pm$0.53 & 6.21$\pm$1.05 & 5.58$\pm$0.58 & 7.22$\pm$0.76 & 13.75$\pm$0.69 & 20.59$\pm$3.57\\
QM8* & \textbf{0.01}$\pm$0.01 & \textbf{0.01}$\pm$0.01 & 0.03$\pm$0.02 & \textendash & 3.28$\pm$5.12 & \textendash & \textendash\\
QM9* & \textbf{3.69}$\pm$9.09 & \textbf{3.69}$\pm$9.09 & 32.11$\pm$50.42 & \textendash & 268.32$\pm$357.01 & \textendash & \textendash\\
\bottomrule
\end{tabular}
\label{table:exp:real:discriminative}
\end{scriptsize}
\end{center}
\vspace{-5pt}
\end{table}
}

\def \TabFigMethodFeaturesAndSyntheticData{
\begin{figure}
    \centering
    \captionof{table}{
Features of different methods.
CQR and MADSplit achieve strict finite-sample \textit{marginal} coverage. %, and DJ would be close in practice \js{can we rephrase or drop DJ would be close? It gives a contradicting impression as we just state DJ is not marginally valid, now we seem to say it is actually in practice.}(See more discussion in the Appendix). 
Unlike \methodname, no baseline is locally valid.
\methodname, MADSplit and CQR are discriminative, DJ is not, and DE/CMDP/PBP are supposed to be but usually fail to in our experiment. 
Among the post-hoc methods, \methodname and MADSplit have reasonable overhead, DJ's overhead is usually $O(N)$ where $N$ is the number of training data (and extremely large memory consumption).
CQR is not post-hoc because its PI may not contain $\hat{\mu}^{NN}(X)$. 
}
\begin{center}
\begin{scriptsize}
\begin{tabular}{c|ccccccc}
& \methodname & MADSplit & CQR & DJ & DE & MCDP & PBP\\
\midrule
Valid & Local & Marginal & Marginal & no guarantee & $\times$ & $\times$ & $\times$ \\
Discriminative& \checkmark & \checkmark & \checkmark & $\times$ & sometimes & sometimes & sometimes \\
Post-hoc & \checkmark & \checkmark & $\times$ & \checkmark & $\times$ & $\times$ & $\times$ \\
Overhead (if post-hoc) & Low & Low & N/A & Very High & N/A & N/A & N/A
\end{tabular} 
\label{table:exp:method_features}
\end{scriptsize}
\end{center}

\includegraphics[width=1\textwidth]{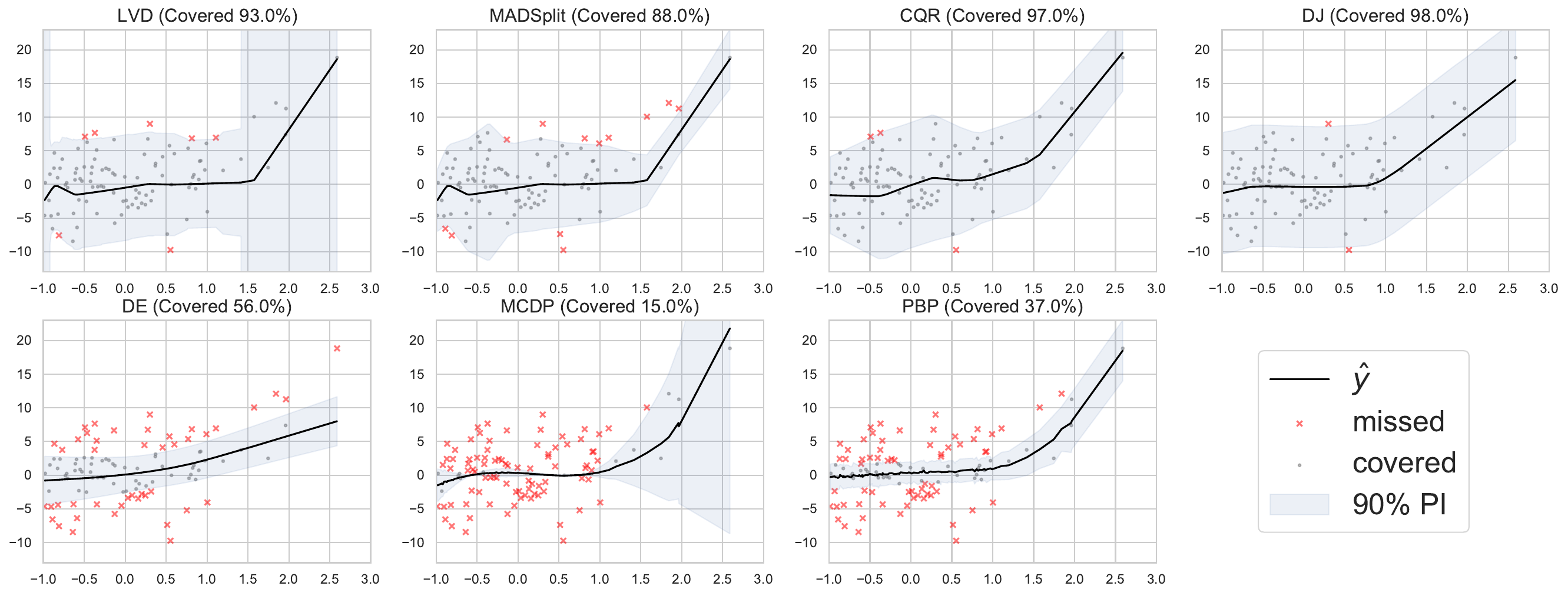}
\captionof{figure}[ Synthetic Data]
{
\methodname, CQR and MADSplit achieve marginal coverage.
Although without theoretical guarantee, DJ usually marginally covers in practice with near constant width (not discriminative).
DE, MCDP and PBP do not show validity as expected.
Among valid PIs, only \methodname tries to capture the less representative points by wider PIs, arguably showing the most useful discriminative pattern.
\label{fig:exp:syn}
}
\vspace{-20pt}
\end{figure}
}

\def \TableCoverageVariants{
\begin{table}[ht]
\captionof{table}{
MCR and TCR for different variants of \methodname.
}
\begin{center}
\begin{tiny}
\begin{tabular}{l|cccccccr|}

\toprule
& \multicolumn{4}{c|}{$\hat{y}^{KR}$} & \multicolumn{4}{c|}{$\hat{y}^{NN}$} \\
& \multicolumn{2}{c|}{No-smooth} & \multicolumn{2}{c|}{Smooth} & \multicolumn{2}{c|}{No-smooth} & \multicolumn{2}{c|}{Smooth} \\
\midrule
MCR        & MN & base & MN & base& MN & base& MN & \boxed{base}\\
\midrule
Yacht      & \textbf{96.6}$\pm$4.5 & \textbf{97.4}$\pm$2.0 & \textbf{95.2}$\pm$4.6 & \textbf{95.5}$\pm$2.3 & \textbf{96.1}$\pm$4.9 & \textbf{97.9}$\pm$1.7 & \textbf{94.7}$\pm$5.0 & \textbf{96.8}$\pm$2.2\\
Housing    & \textbf{96.8}$\pm$3.4 & \textbf{97.1}$\pm$2.8 & \textbf{96.0}$\pm$3.9 & \textbf{95.7}$\pm$3.5 & \textbf{97.3}$\pm$2.2 & \textbf{97.8}$\pm$2.1 & \textbf{96.1}$\pm$2.6 & \textbf{96.8}$\pm$2.9\\
Energy     & \textbf{92.8}$\pm$2.7 & \textbf{92.4}$\pm$2.9 & \textbf{92.5}$\pm$2.6 & \textbf{92.4}$\pm$2.9 & \textbf{94.0}$\pm$1.7 & \textbf{94.1}$\pm$1.6 & \textbf{93.9}$\pm$1.7 & \textbf{94.0}$\pm$1.6\\
Bike       & \textbf{91.6}$\pm$1.2 & \textbf{93.8}$\pm$0.8 & \textbf{91.6}$\pm$1.2 & \textbf{94.1}$\pm$0.8 & \textbf{90.6}$\pm$0.5 & \textbf{90.5}$\pm$0.8 & \textbf{90.4}$\pm$0.6 & \textbf{90.4}$\pm$0.8\\
Kin8nm     & \textbf{100.0}$\pm$0.0 & \textbf{100.0}$\pm$0.0 & \textbf{97.9}$\pm$0.7 & \textbf{97.9}$\pm$0.8 & \textbf{100.0}$\pm$0.0 & \textbf{100.0}$\pm$0.0 & \textbf{97.9}$\pm$0.6 & \textbf{98.0}$\pm$0.6\\
Concrete   & \textbf{99.6}$\pm$0.7 & \textbf{99.6}$\pm$0.8 & \textbf{96.7}$\pm$2.3 & \textbf{97.0}$\pm$1.1 & \textbf{99.7}$\pm$0.6 & \textbf{99.6}$\pm$0.7 & \textbf{97.0}$\pm$2.1 & \textbf{97.4}$\pm$1.3\\
QM8* & \textbf{94.9}$\pm$1.4 & \textbf{95.3}$\pm$1.3 & \textbf{92.3}$\pm$0.8 & \textbf{92.9}$\pm$0.9 & \textbf{94.7}$\pm$1.5 & \textbf{95.1}$\pm$1.4 & \textbf{92.0}$\pm$0.9 & \textbf{92.6}$\pm$0.9\\
QM9* & \textbf{94.0}$\pm$1.7 & \textbf{94.1}$\pm$1.6 & \textbf{90.6}$\pm$0.4 & \textbf{90.4}$\pm$0.5 & \textbf{93.5}$\pm$1.5 & \textbf{93.7}$\pm$1.5 & \textbf{90.3}$\pm$0.4 & \textbf{90.3}$\pm$0.6\\
\bottomrule%=============================MCR=========================
\toprule%================================TCR=========================
TCR & \\
\midrule
Yacht      & \textbf{96.9}$\pm$4.0 & \textbf{96.9}$\pm$5.4 & \textbf{93.1}$\pm$7.6 & \textbf{94.6}$\pm$5.2 & \textbf{95.4}$\pm$7.4 & \textbf{99.2}$\pm$2.4 & \textbf{93.8}$\pm$7.1 & \textbf{98.5}$\pm$3.2\\
Housing    & \textbf{95.7}$\pm$4.2 & \textbf{96.2}$\pm$4.4 & \textbf{93.3}$\pm$9.3 & \textbf{91.4}$\pm$8.3 & \textbf{98.1}$\pm$3.3 & \textbf{97.1}$\pm$4.0 & \textbf{98.1}$\pm$2.5 & \textbf{96.2}$\pm$4.4\\
Energy     & \textbf{86.8}$\pm$6.2 & 83.5$\pm$9.7 & 85.8$\pm$6.1 & 83.2$\pm$10.2 & \textbf{88.1}$\pm$4.8 & \textbf{87.1}$\pm$5.9 & \textbf{87.7}$\pm$4.8 & \textbf{86.8}$\pm$5.8\\
Bike       & \textbf{91.8}$\pm$1.0 & \textbf{91.9}$\pm$2.1 & \textbf{90.9}$\pm$1.6 & \textbf{91.6}$\pm$2.7 & \textbf{92.0}$\pm$1.3 & \textbf{90.8}$\pm$1.5 & \textbf{91.6}$\pm$1.3 & \textbf{90.2}$\pm$1.7\\
Kin8nm     & \textbf{100.0}$\pm$0.0 & \textbf{100.0}$\pm$0.0 & \textbf{95.7}$\pm$1.6 & \textbf{95.0}$\pm$2.1 & \textbf{100.0}$\pm$0.0 & \textbf{100.0}$\pm$0.0 & \textbf{97.1}$\pm$1.5 & \textbf{97.2}$\pm$1.6\\
Concrete   & \textbf{99.0}$\pm$2.1 & \textbf{99.3}$\pm$1.6 & \textbf{93.9}$\pm$4.5 & \textbf{94.4}$\pm$3.6 & \textbf{99.5}$\pm$1.0 & \textbf{99.5}$\pm$1.5 & \textbf{96.8}$\pm$3.8 & \textbf{97.1}$\pm$3.4\\
QM8* & \textbf{94.6}$\pm$2.1 & \textbf{95.6}$\pm$2.4 & \textbf{90.4}$\pm$2.0 & \textbf{91.4}$\pm$2.6 & \textbf{94.9}$\pm$1.9 & \textbf{94.8}$\pm$2.2 & \textbf{91.2}$\pm$1.7 & \textbf{90.8}$\pm$1.9\\
QM9* & \textbf{94.4}$\pm$4.4 & \textbf{94.3}$\pm$4.5 & \textbf{88.5}$\pm$3.7 & \textbf{88.0}$\pm$3.7 & \textbf{94.9}$\pm$3.1 & \textbf{94.8}$\pm$3.4 & \textbf{90.1}$\pm$2.3 & \textbf{89.7}$\pm$2.5\\
\bottomrule
\end{tabular}
\label{appendix:table:exp:real:cov_variant}
\end{tiny}
\end{center}
\end{table}
}

\def \TableDiscVariants{
\begin{table}[ht]
\captionof{table}{
AUROC and MAD for different variants of \methodname.
Best AUROCs are in bold, and all are significantly higher than 50 (at $p=0.05$).
For MAD, the best for each task, if significantly better than the second-best (at $p=0.05$), are in bold.
}
\begin{center}
\begin{tiny}
\begin{tabular}{l|cccccccr}
\toprule
& \multicolumn{4}{c|}{$\hat{y}^{KR}$} & \multicolumn{4}{c|}{$\hat{y}^{NN}$} \\
& \multicolumn{2}{c|}{No-smooth} & \multicolumn{2}{c|}{Smooth} & \multicolumn{2}{c|}{No-smooth} & \multicolumn{2}{c|}{Smooth} \\
\midrule
AUROC        & MN & base & MN & base& MN & base& MN &\boxed{base}\\
\midrule
Yacht & 71.1$\pm$7.1 & 74.2$\pm$3.5 & 61.5$\pm$12.4 & 67.5$\pm$7.1 & 81.0$\pm$6.1 & \textbf{83.8}$\pm$5.4 & 80.9$\pm$6.1 & 83.5$\pm$5.8\\
Housing & 58.7$\pm$7.1 & 62.0$\pm$6.6 & 61.4$\pm$6.7 & \textbf{64.4}$\pm$5.6 & 62.6$\pm$7.9 & 60.0$\pm$7.0 & 62.1$\pm$9.0 & 59.2$\pm$8.5\\
Energy & 61.8$\pm$5.0 & 60.8$\pm$2.9 & 63.3$\pm$5.3 & 62.9$\pm$4.6 & 74.3$\pm$7.2 & 73.5$\pm$6.3 & \textbf{74.3}$\pm$7.2 & 73.5$\pm$6.3\\
Bike & 73.5$\pm$7.8 & 86.6$\pm$3.6 & 73.7$\pm$7.5 & \textbf{87.5}$\pm$3.2 & 72.3$\pm$8.5 & 68.1$\pm$11.1 & 72.4$\pm$8.5 & 68.2$\pm$11.0\\
Kin8nm & 55.7$\pm$1.8 & 55.9$\pm$2.1 & 60.5$\pm$1.9 & \textbf{61.8}$\pm$1.6 & 57.1$\pm$2.4 & 56.8$\pm$2.4 & 61.6$\pm$1.6 & 60.3$\pm$1.1\\
Concrete & 60.4$\pm$3.5 & 60.1$\pm$3.4 & 62.8$\pm$4.6 & 61.8$\pm$5.0 & 62.7$\pm$8.4 & 62.4$\pm$8.4 & \textbf{65.4}$\pm$6.1 & 64.0$\pm$6.1\\
QM8* & 73.2$\pm$9.6 & 72.8$\pm$11.9 & \textbf{75.9}$\pm$9.0 & 75.2$\pm$11.9 & 72.9$\pm$7.7 & 71.3$\pm$9.5 & 74.1$\pm$6.9 & 71.3$\pm$9.4\\
QM9* & \textbf{68.2}$\pm$7.6 & 67.3$\pm$8.9 & 66.5$\pm$3.5 & 66.4$\pm$5.4 & 66.2$\pm$3.5 & 64.1$\pm$3.7 & 66.3$\pm$3.5 & 62.7$\pm$3.6\\
\bottomrule%=============================AUROC=======================
\toprule%================================MAD=========================
MAD&\\
\midrule
Yacht      & \textbf{0.79}$\pm$0.09 & \textbf{0.79}$\pm$0.09 & 1.14$\pm$0.13 & 1.14$\pm$0.13 & 1.90$\pm$0.48 & 1.90$\pm$0.48 & 1.90$\pm$0.48 & 1.90$\pm$0.48\\
Housing    & 2.86$\pm$0.31 & 2.86$\pm$0.31 & 3.00$\pm$0.32 & 3.00$\pm$0.32 & 3.31$\pm$0.53 & 3.31$\pm$0.53 & 3.31$\pm$0.53 & 3.31$\pm$0.53\\
Energy     & 2.34$\pm$0.07 & 2.34$\pm$0.07 & 2.35$\pm$0.08 & 2.35$\pm$0.08 & 2.99$\pm$0.75 & 2.99$\pm$0.75 & 2.99$\pm$0.75 & 2.99$\pm$0.75\\
Bike       & 2.47$\pm$0.72 & 2.47$\pm$0.72 & 3.79$\pm$0.49 & 3.79$\pm$0.49 & \textbf{0.04}$\pm$0.03 & \textbf{0.04}$\pm$0.03 & \textbf{0.04}$\pm$0.03 & \textbf{0.04}$\pm$0.03\\
Kin8nm     & \textbf{0.06}$\pm$0.00 & \textbf{0.06}$\pm$0.00 & 0.07$\pm$0.00 & 0.07$\pm$0.00 & 0.07$\pm$0.00 & 0.07$\pm$0.00 & 0.07$\pm$0.00 & 0.07$\pm$0.00\\
Concrete   & \textbf{4.76}$\pm$0.26 & \textbf{4.76}$\pm$0.26 & 5.20$\pm$0.29 & 5.20$\pm$0.29 & 5.44$\pm$0.53 & 5.44$\pm$0.53 & 5.44$\pm$0.53 & 5.44$\pm$0.53\\
QM8* & 0.01$\pm$0.01 & 0.01$\pm$0.01 & 0.01$\pm$0.01 & 0.01$\pm$0.01 & 0.01$\pm$0.01 & 0.01$\pm$0.01 & 0.01$\pm$0.01 & 0.01$\pm$0.01\\
QM9* & 3.58$\pm$9.71 & 3.58$\pm$9.71 & 4.92$\pm$11.39 & 4.92$\pm$11.39 & 3.69$\pm$9.09 & 3.69$\pm$9.09 & 3.69$\pm$9.09 & 3.69$\pm$9.09\\
\bottomrule
\end{tabular}
\label{appendix:table:exp:real:disc_variant}
\end{tiny}
\end{center}
\end{table}
}

\def \TableFiftyVariants{
\begin{table}[ht]
\captionof{table}{
Counts of finite PIs and average width for different variants of \methodname (restricted to the subset for which all PIs are finite), with $\alpha=0.5$. 
In the count table, the size of the test set is included in the parenthesis, and the lowest count (which is used for width computation) is underscored.
}
\begin{center}
\begin{tiny}
\begin{tabular}{l|cccccccr}
\toprule
& \multicolumn{4}{c|}{$\hat{y}^{KR}$} & \multicolumn{4}{c|}{$\hat{y}^{NN}$} \\
& \multicolumn{2}{c|}{No-smooth} & \multicolumn{2}{c|}{Smooth} & \multicolumn{2}{c|}{No-smooth} & \multicolumn{2}{c|}{Smooth} \\
\midrule
\# finite @ 50\%        & MN & base & MN & base& MN & base& MN & \boxed{base}\\
\midrule
Yacht(62) & \underline{60.7}$\pm$1.9 & \underline{60.7}$\pm$1.9 & 61.9$\pm$0.3 & 61.9$\pm$0.3 & \underline{60.7}$\pm$1.9 & \underline{60.7}$\pm$1.9 & 61.9$\pm$0.3 & 61.9$\pm$0.3\\
Housing(101) & \underline{93.5}$\pm$5.7 & \underline{93.5}$\pm$5.7 & 98.3$\pm$3.1 & 98.3$\pm$3.1 & \underline{93.5}$\pm$5.7 & \underline{93.5}$\pm$5.7 & 98.3$\pm$3.1 & 98.3$\pm$3.1\\
Energy(154) & \underline{154.0}$\pm$0.0 & \underline{154.0}$\pm$0.0 & \underline{154.0}$\pm$0.0 & \underline{154.0}$\pm$0.0 & \underline{154.0}$\pm$0.0 & \underline{154.0}$\pm$0.0 & \underline{154.0}$\pm$0.0 & \underline{154.0}$\pm$0.0\\
Bike(3476) & \underline{3473.0}$\pm$2.8 & \underline{3473.0}$\pm$2.8 & 3475.2$\pm$1.2 & 3475.2$\pm$1.2 & \underline{3473.0}$\pm$2.8 & \underline{3473.0}$\pm$2.8 & 3475.2$\pm$1.2 & 3475.2$\pm$1.2\\
Kin8nm(1638) & \underline{844.3}$\pm$181.0 & \underline{844.3}$\pm$181.0 & 1610.1$\pm$10.2 & 1610.1$\pm$10.2 & \underline{844.3}$\pm$181.0 & \underline{844.3}$\pm$181.0 & 1610.1$\pm$10.2 & 1610.1$\pm$10.2\\
Concrete(206) & \underline{176.5}$\pm$21.7 & \underline{176.5}$\pm$21.7 & 200.1$\pm$4.0 & 200.1$\pm$4.0 & \underline{176.5}$\pm$21.7 & \underline{176.5}$\pm$21.7 & 200.1$\pm$4.0 & 200.1$\pm$4.0\\
QM8*(4357) & \underline{4002.4}$\pm$210.2 & \underline{4002.4}$\pm$210.2 & 4317.3$\pm$22.9 & 4317.3$\pm$22.9 & \underline{4002.4}$\pm$210.2 & \underline{4002.4}$\pm$210.2 & 4317.3$\pm$22.9 & 4317.3$\pm$22.9\\
QM9*(26744) & \underline{25376.7}$\pm$792.3 & \underline{25376.7}$\pm$792.3 & 26616.7$\pm$39.8 & 26616.7$\pm$39.8 & \underline{25376.7}$\pm$792.3 & \underline{25376.7}$\pm$792.3 & 26616.7$\pm$39.8 & 26616.7$\pm$39.8\\

\bottomrule%=============================AUROC=======================
\toprule%================================MAD=========================
Width @ 50\% &\\
\midrule
Yacht & 1.8$\pm$0.5 & \textbf{1.7}$\pm$0.5 & 2.2$\pm$0.5 & 2.1$\pm$0.4 & 4.5$\pm$0.9 & 4.4$\pm$0.9 & 4.1$\pm$0.9 & 3.8$\pm$0.9\\
Housing & 5.9$\pm$0.6 & 5.7$\pm$0.8 & 5.5$\pm$0.6 & \textbf{5.3}$\pm$0.7 & 7.2$\pm$1.1 & 7.0$\pm$1.1 & 6.7$\pm$1.3 & 6.4$\pm$1.0\\
Energy & 4.3$\pm$0.5 & \textbf{4.3}$\pm$0.3 & 4.3$\pm$0.4 & 4.3$\pm$0.3 & 6.2$\pm$1.5 & 6.1$\pm$1.4 & 6.2$\pm$1.5 & 6.1$\pm$1.4\\
Bike & 5.4$\pm$1.5 & 4.8$\pm$1.3 & 8.3$\pm$1.4 & 7.3$\pm$0.9 & 0.1$\pm$0.1 & 0.1$\pm$0.0 & 0.1$\pm$0.1 & \textbf{0.1}$\pm$0.0\\
Kin8nm & 0.2$\pm$0.0 & 0.2$\pm$0.0 & 0.1$\pm$0.0 & \textbf{0.1}$\pm$0.0 & 0.2$\pm$0.0 & 0.2$\pm$0.0 & 0.1$\pm$0.0 & 0.1$\pm$0.0\\
Concrete & 11.5$\pm$1.2 & 11.2$\pm$1.1 & 10.1$\pm$1.1 & \textbf{9.8}$\pm$1.1 & 13.1$\pm$2.1 & 12.8$\pm$2.2 & 10.8$\pm$1.8 & 10.5$\pm$1.8\\
QM8* & 0.0$\pm$0.0 & 0.0$\pm$0.0 & 0.0$\pm$0.0 & \textbf{0.0}$\pm$0.0 & 0.0$\pm$0.0 & 0.0$\pm$0.0 & 0.0$\pm$0.0 & 0.0$\pm$0.0\\
QM9* & 6.1$\pm$17.3 & 5.5$\pm$15.6 & 7.6$\pm$18.6 & 6.8$\pm$16.7 & 6.3$\pm$15.8 & 5.6$\pm$14.2 & 5.7$\pm$14.7 & \textbf{5.0}$\pm$12.9\\
\bottomrule
\end{tabular}
\label{appendix:table:exp:real:fifty_width_count}
\end{tiny}
\end{center}
\end{table}
}

\def \TableNintyVariants{
\begin{table}[ht]
\captionof{table}{
Same as Table \ref{appendix:table:exp:real:fifty_width_count}, but with $\alpha=0.1$. 
}
\begin{center}
\begin{tiny}
\begin{tabular}{l|cccccccr}
\toprule
& \multicolumn{4}{c|}{$\hat{y}^{KR}$} & \multicolumn{4}{c|}{$\hat{y}^{NN}$} \\
& \multicolumn{2}{c|}{No-smooth} & \multicolumn{2}{c|}{Smooth} & \multicolumn{2}{c|}{No-smooth} & \multicolumn{2}{c|}{Smooth} \\
\midrule
\# finite @ 90\%        & MN & base & MN & base& MN & base& MN & \boxed{base}\\
\midrule
Yacht(62) & \underline{35.1}$\pm$5.8 & \underline{35.1}$\pm$5.8 & 40.9$\pm$2.8 & 40.9$\pm$2.8 & \underline{35.1}$\pm$5.8 & \underline{35.1}$\pm$5.8 & 40.9$\pm$2.8 & 40.9$\pm$2.8\\
Housing(101) & \underline{54.5}$\pm$22.1 & \underline{54.5}$\pm$22.1 & 69.0$\pm$19.1 & 69.0$\pm$19.1 & \underline{54.5}$\pm$22.1 & \underline{54.5}$\pm$22.1 & 69.0$\pm$19.1 & 69.0$\pm$19.1\\
Energy(154) & \underline{145.1}$\pm$11.0 & \underline{145.1}$\pm$11.0 & \underline{145.1}$\pm$11.0 & \underline{145.1}$\pm$11.0 & \underline{145.1}$\pm$11.0 & \underline{145.1}$\pm$11.0 & \underline{145.1}$\pm$11.0 & \underline{145.1}$\pm$11.0\\
Bike(3476) & \underline{3458.5}$\pm$11.4 & \underline{3458.5}$\pm$11.4 & 3467.5$\pm$4.4 & 3467.5$\pm$4.4 & \underline{3458.5}$\pm$11.4 & \underline{3458.5}$\pm$11.4 & 3467.5$\pm$4.4 & 3467.5$\pm$4.4\\
Kin8nm(1638) & \underline{0.4}$\pm$1.0 & \underline{0.4}$\pm$1.0 & 938.0$\pm$123.7 & 938.0$\pm$123.7 & \underline{0.4}$\pm$1.0 & \underline{0.4}$\pm$1.0 & 938.0$\pm$123.7 & 938.0$\pm$123.7\\
Concrete(206) & \underline{28.8}$\pm$24.8 & \underline{28.8}$\pm$24.8 & 133.8$\pm$20.1 & 133.8$\pm$20.1 & \underline{28.8}$\pm$24.8 & \underline{28.8}$\pm$24.8 & 133.8$\pm$20.1 & 133.8$\pm$20.1\\
QM8*(4357) & \underline{2936.9}$\pm$587.9 & \underline{2936.9}$\pm$587.9 & 4041.6$\pm$136.8 & 4041.6$\pm$136.8 & \underline{2936.9}$\pm$587.9 & \underline{2936.9}$\pm$587.9 & 4041.6$\pm$136.8 & 4041.6$\pm$136.8\\
QM9*(26744) & \underline{21347.5}$\pm$2850.8 & \underline{21347.5}$\pm$2850.8 & 26147.0$\pm$151.6 & 26147.0$\pm$151.6 & \underline{21347.5}$\pm$2850.8 & \underline{21347.5}$\pm$2850.8 & 26147.0$\pm$151.6 & 26147.0$\pm$151.6\\
\bottomrule%=============================AUROC=======================
\toprule%================================MAD=========================
Width @ 90\% &\\
\midrule
Yacht & 5.69$\pm$2.98 & \textbf{2.21}$\pm$0.58 & 7.68$\pm$5.57 & 2.33$\pm$0.41 & 5.82$\pm$3.29 & 3.03$\pm$1.40 & 4.85$\pm$2.08 & 2.97$\pm$1.41\\
Housing & 32.33$\pm$34.63 & 14.14$\pm$2.09 & 22.67$\pm$21.69 & \textbf{13.29}$\pm$1.85 & 43.46$\pm$56.19 & 15.50$\pm$2.89 & 25.76$\pm$20.69 & 14.48$\pm$2.64\\
Energy & 14.61$\pm$7.54 & \textbf{12.47}$\pm$1.48 & 15.07$\pm$8.52 & 12.49$\pm$1.47 & 15.94$\pm$10.07 & 12.91$\pm$2.03 & 15.92$\pm$10.08 & 12.89$\pm$2.02\\
Bike & 21.46$\pm$11.96 & 8.33$\pm$2.43 & 35.62$\pm$21.29 & 11.78$\pm$1.72 & 0.19$\pm$0.11 & 0.15$\pm$0.13 & 0.19$\pm$0.12 & \textbf{0.15}$\pm$0.13\\
Kin8nm & 0.44$\pm$0.03 & 0.36$\pm$0.02 & 0.27$\pm$0.12 & 0.25$\pm$0.03 & 0.39$\pm$0.08 & 0.37$\pm$0.02 & 0.25$\pm$0.11 & \textbf{0.24}$\pm$0.02\\
Concrete & 27.25$\pm$4.77 & 26.30$\pm$4.71 & \textbf{20.55}$\pm$4.06 & 21.76$\pm$2.67 & 28.79$\pm$6.79 & 28.25$\pm$5.16 & 21.56$\pm$4.00 & 23.41$\pm$3.50\\
QM8* & 0.05$\pm$0.03 & 0.04$\pm$0.02 & 0.04$\pm$0.02 & \textbf{0.04}$\pm$0.02 & 0.05$\pm$0.03 & 0.04$\pm$0.02 & 0.04$\pm$0.02 & 0.04$\pm$0.02\\
QM9* & 14.87$\pm$41.31 & 15.01$\pm$42.57 & 17.86$\pm$43.59 & 17.20$\pm$43.19 & 14.84$\pm$38.27 & 15.22$\pm$40.44 & \textbf{13.65}$\pm$35.71 & 14.04$\pm$37.44\\
\bottomrule
\end{tabular}
\label{appendix:table:exp:real:ninty_width_count}
\end{tiny}
\end{center}
\end{table}
}

\def \TableWidth{
\begin{table}[ht]
\captionof{table}{
Average width of different conformal methods. 
Width significantly shorter than the second-best at $p=0.05$ are in bold.
}
\begin{center}
\begin{tiny}
\begin{tabular}{lccccccccr}
\toprule
  & \multicolumn{4}{c}{50\%-PI Width} & \multicolumn{4}{c}{90\%-PI Width}\\
Data (Count) & \# finite  & \methodname & MADSplit   & CQR        &  \# finite &  \methodname &  MADSplit  &  CQR      \\
\midrule
Yacht(62)  & 61.90$\pm$0.32 & 3.99$\pm$0.79 & 3.17$\pm$0.84 & 2.82$\pm$0.74 & 40.90$\pm$2.85 & 3.47$\pm$1.36 & 3.29$\pm$1.04 & 4.52$\pm$2.08\\
Housing(101) & 98.30$\pm$3.06 & 6.70$\pm$0.97 & 6.02$\pm$1.23 & \textbf{4.98}$\pm$0.72 & 69.00$\pm$19.11 & 15.94$\pm$2.62 & 16.81$\pm$7.44 & \textbf{13.70}$\pm$1.84\\
Energy(154) & 154.00$\pm$0.00 & 6.06$\pm$1.41 & 5.77$\pm$1.37 & 5.18$\pm$1.47 & 145.10$\pm$11.05 & 12.89$\pm$2.02 & 12.19$\pm$2.71 & 13.76$\pm$2.80\\
Bike(3476) & 3475.20$\pm$1.23 & 0.06$\pm$0.05 & 0.07$\pm$0.05 & 5.62$\pm$4.41 & 3467.50$\pm$4.40 & 0.15$\pm$0.13 & 0.19$\pm$0.11 & 33.65$\pm$21.52\\
Kin8nm(1638) & 1610.10$\pm$10.18 & 0.14$\pm$0.01 & 0.12$\pm$0.01 & 0.12$\pm$0.01 & 938.00$\pm$123.72 & 0.34$\pm$0.02 & 0.28$\pm$0.02 & 0.28$\pm$0.02\\
Concrete(206) & 200.10$\pm$4.01 & 10.92$\pm$1.85 & 9.77$\pm$1.66 & 9.35$\pm$2.84 & 133.80$\pm$20.13 & 27.93$\pm$3.59 & 21.79$\pm$2.93 & 22.87$\pm$3.48\\
QM8*(4357) & 4317.33$\pm$22.90 & 0.02$\pm$0.01 & 0.02$\pm$0.01 & 0.04$\pm$0.01 & 4041.63$\pm$136.85 & 0.05$\pm$0.03 & 0.05$\pm$0.03 & 0.11$\pm$0.03\\
QM9*(26744) & 26616.72$\pm$39.77 & 5.12$\pm$13.17 & 5.75$\pm$14.78 & 37.32$\pm$65.01 & 26146.95$\pm$151.58 & 15.06$\pm$38.94 & 14.77$\pm$37.04 & 129.63$\pm$207.46\\
\bottomrule
\end{tabular}
\label{table:exp:real:width}
\end{tiny}
\end{center}
\end{table}
}

\def \TableWidthInfAppendix{
\begin{table}[ht]
\captionof{table}{
Average width of all baselines methods, without restriction to the subsample for which \methodname gives finite PIs.
}
\begin{center}
\begin{scriptsize}
\begin{tabular}{lcccccccr}
\toprule
Width @ 50\% & MADSplit   & CQR        & DJ         & DE         & MCDP       & PBP       \\
\midrule
Yacht      & 3.18$\pm$0.82 & 2.83$\pm$0.72 & 19.10$\pm$1.26 & 5.26$\pm$0.78 & 14.34$\pm$0.71 & 2.16$\pm$0.31\\
Housing    & 6.06$\pm$1.22 & 5.00$\pm$0.71 & 11.50$\pm$1.41 & 10.23$\pm$1.50 & 30.58$\pm$0.33 & 0.74$\pm$0.08\\
Energy     & 5.77$\pm$1.37 & 5.18$\pm$1.47 & 9.83$\pm$1.39 & 10.52$\pm$1.59 & 30.14$\pm$0.24 & 0.78$\pm$0.04\\
Bike       & 0.07$\pm$0.05 & 5.62$\pm$4.41 & 0.14$\pm$0.07 & 13.62$\pm$6.61 & 115.47$\pm$0.84 & 0.84$\pm$0.27\\
Kin8nm     & 0.12$\pm$0.01 & 0.12$\pm$0.01 & 0.25$\pm$0.02 & 0.80$\pm$0.03 & 0.98$\pm$0.02 & 1.29$\pm$0.14\\
Concrete   & 9.84$\pm$1.70 & 9.39$\pm$2.82 & 47.98$\pm$84.01 & 18.33$\pm$2.96 & 47.82$\pm$0.30 & 0.78$\pm$0.06\\
QM8* &  0.05$\pm$0.03 & 0.11$\pm$0.03 & \textendash & 42.17$\pm$28.01 & \textendash & \textendash\\
QM9* &  14.77$\pm$37.04 & 129.63$\pm$207.46 & \textendash & 465.17$\pm$919.56 & \textendash & \textendash\\

\bottomrule
\toprule
Width @ 90\% & MADSplit   & CQR        & DJ         & DE         & MCDP       & PBP       \\
\midrule
Yacht      & 8.02$\pm$0.98 & 12.31$\pm$1.79 & 73.14$\pm$1.75 & 13.13$\pm$1.24 & 34.96$\pm$1.73 & 5.26$\pm$0.77\\
Housing    & 18.71$\pm$9.91 & 15.10$\pm$1.79 & 26.31$\pm$1.87 & 24.97$\pm$2.58 & 74.57$\pm$0.81 & 1.82$\pm$0.19\\
Energy     & 12.24$\pm$2.78 & 13.75$\pm$2.88 & 18.54$\pm$2.10 & 25.65$\pm$5.58 & 73.50$\pm$0.60 & 1.91$\pm$0.09\\
Bike       & 0.19$\pm$0.11 & 33.96$\pm$21.87 & 0.32$\pm$0.18 & 38.50$\pm$13.26 & 281.24$\pm$1.66 & 2.04$\pm$0.65\\
Kin8nm     & 0.31$\pm$0.02 & 0.32$\pm$0.01 & 0.48$\pm$0.03 & 1.89$\pm$0.15 & 2.39$\pm$0.06 & 3.15$\pm$0.35\\
Concrete   & 22.52$\pm$2.93 & 23.29$\pm$3.32 & 199.19$\pm$369.33 & 44.02$\pm$4.73 & 116.62$\pm$0.74 & 1.90$\pm$0.16\\
QM8* & 0.05$\pm$0.03 & 0.11$\pm$0.03 & \textendash & 42.17$\pm$28.01 & \textendash & \textendash\\
QM9* &  14.77$\pm$37.04 & 129.63$\pm$207.46 & \textendash & 465.17$\pm$919.56 & \textendash & \textendash\\
\bottomrule
\end{tabular}
\label{appendix:table:exp:real:widthinf}
\end{scriptsize}
\end{center}
\end{table}
}

\def \TableQMCoverage{
\begin{table}[ht]
\captionof{table}{
MCR and TCR for QM8 and QM9, for 90\% PI.
Numbers \textit{not} significantly lower than 90\% are in bold.
Like in the main text, MADSplit and CQR achieves 90\% marginal coverage rate empirically as expected, but fail to cover data with more extreme responses (in the tails).
Variants of \methodname almost always cover empirically, measured by both MCR and TCR.
}
\begin{center}
\begin{tiny}
\setlength\tabcolsep{2pt}
\begin{tabular}{l|cccccccc|cr}

\toprule

& \multicolumn{8}{c|}{\methodname} &  \multicolumn{2}{c}{Conformal Baselines}\\
& \multicolumn{4}{c|}{$\hat{y}^{KR}$} & \multicolumn{4}{c|}{$\hat{y}^{NN}$}  & & \\
& \multicolumn{2}{c|}{No-smooth} & \multicolumn{2}{c|}{Smooth} & \multicolumn{2}{c|}{No-smooth} & \multicolumn{2}{c|}{Smooth} & & \\
\midrule
MCR & MN & base & MN & base& MN & base& MN & \boxed{base}& MADSplit & CQR\\
\midrule
QM8(E1-CC2) & \textbf{96.3}$\pm$0.7 & \textbf{96.3}$\pm$0.8 & \textbf{92.6}$\pm$0.9 & \textbf{92.7}$\pm$0.6 & \textbf{96.2}$\pm$0.8 & \textbf{96.2}$\pm$0.6 & \textbf{92.3}$\pm$0.6 & \textbf{92.5}$\pm$0.5 & \textbf{90.0}$\pm$0.5 & \textbf{90.1}$\pm$0.7\\
QM8(E2-CC2) & \textbf{96.2}$\pm$1.0 & \textbf{96.2}$\pm$0.8 & \textbf{92.0}$\pm$0.5 & \textbf{92.2}$\pm$0.7 & \textbf{96.1}$\pm$1.0 & \textbf{96.2}$\pm$0.7 & \textbf{92.0}$\pm$0.5 & \textbf{92.2}$\pm$0.5 & \textbf{89.8}$\pm$0.6 & \textbf{90.1}$\pm$0.5\\
QM8(f1-CC2) & \textbf{93.9}$\pm$0.9 & \textbf{94.8}$\pm$0.9 & \textbf{92.3}$\pm$0.7 & \textbf{93.8}$\pm$0.8 & \textbf{93.7}$\pm$1.0 & \textbf{94.0}$\pm$1.2 & \textbf{92.2}$\pm$0.8 & \textbf{92.5}$\pm$1.1 & \textbf{90.1}$\pm$0.7 & \textbf{90.2}$\pm$0.7\\
QM8(f2-CC2) & \textbf{94.8}$\pm$1.3 & \textbf{95.1}$\pm$1.1 & \textbf{93.0}$\pm$0.6 & \textbf{93.6}$\pm$0.6 & \textbf{94.3}$\pm$1.3 & \textbf{95.1}$\pm$1.3 & \textbf{92.5}$\pm$0.8 & \textbf{93.6}$\pm$0.6 & \textbf{89.9}$\pm$0.7 & \textbf{89.9}$\pm$0.7\\
QM8(E1-PBE0) & \textbf{95.9}$\pm$0.5 & \textbf{96.0}$\pm$0.6 & \textbf{92.3}$\pm$0.7 & \textbf{92.7}$\pm$0.6 & \textbf{95.7}$\pm$0.7 & \textbf{95.9}$\pm$0.6 & \textbf{91.9}$\pm$0.8 & \textbf{92.5}$\pm$0.6 & \textbf{89.7}$\pm$0.5 & \textbf{90.2}$\pm$0.5\\
QM8(E2-PBE0) & \textbf{95.0}$\pm$0.9 & \textbf{95.4}$\pm$0.9 & \textbf{91.8}$\pm$0.5 & \textbf{92.4}$\pm$0.8 & \textbf{95.0}$\pm$0.9 & \textbf{95.3}$\pm$0.9 & \textbf{91.7}$\pm$0.6 & \textbf{92.3}$\pm$0.6 & \textbf{90.0}$\pm$0.5 & \textbf{90.0}$\pm$0.6\\
QM8(f1-PBE0) & \textbf{93.7}$\pm$1.1 & \textbf{94.3}$\pm$1.2 & \textbf{92.4}$\pm$1.0 & \textbf{93.1}$\pm$1.0 & \textbf{93.4}$\pm$1.2 & \textbf{93.9}$\pm$1.4 & \textbf{91.8}$\pm$1.3 & \textbf{92.4}$\pm$1.2 & \textbf{90.3}$\pm$0.8 & \textbf{90.2}$\pm$0.9\\
QM8(f2-PBE0) & \textbf{94.0}$\pm$1.8 & \textbf{94.6}$\pm$1.8 & \textbf{91.9}$\pm$1.1 & \textbf{92.5}$\pm$0.9 & \textbf{93.9}$\pm$1.9 & \textbf{94.8}$\pm$1.6 & \textbf{91.6}$\pm$1.2 & \textbf{92.8}$\pm$0.9 & \textbf{89.9}$\pm$0.9 & 89.7$\pm$0.5\\
QM8(E1-PBE0.1) & \textbf{95.8}$\pm$0.6 & \textbf{96.0}$\pm$0.6 & \textbf{92.2}$\pm$0.6 & \textbf{92.7}$\pm$0.6 & \textbf{95.6}$\pm$0.7 & \textbf{95.9}$\pm$0.5 & \textbf{91.9}$\pm$0.8 & \textbf{92.5}$\pm$0.7 & \textbf{89.8}$\pm$0.5 & \textbf{89.8}$\pm$0.6\\
QM8(E2-PBE0.1) & \textbf{94.9}$\pm$0.8 & \textbf{95.4}$\pm$0.9 & \textbf{91.7}$\pm$0.4 & \textbf{92.4}$\pm$0.8 & \textbf{95.0}$\pm$1.0 & \textbf{95.3}$\pm$0.9 & \textbf{91.8}$\pm$0.5 & \textbf{92.3}$\pm$0.6 & \textbf{90.2}$\pm$0.5 & \textbf{90.0}$\pm$0.4\\
QM8(f1-PBE0.1) & \textbf{93.6}$\pm$1.0 & \textbf{94.3}$\pm$1.2 & \textbf{92.3}$\pm$0.9 & \textbf{93.1}$\pm$1.0 & \textbf{93.1}$\pm$1.4 & \textbf{93.9}$\pm$1.3 & \textbf{91.6}$\pm$1.3 & \textbf{92.3}$\pm$1.4 & \textbf{89.9}$\pm$1.0 & \textbf{90.0}$\pm$0.4\\
QM8(f2-PBE0.1) & \textbf{94.2}$\pm$1.8 & \textbf{94.6}$\pm$1.8 & \textbf{92.0}$\pm$1.1 & \textbf{92.5}$\pm$0.9 & \textbf{94.0}$\pm$1.9 & \textbf{94.8}$\pm$1.7 & \textbf{91.8}$\pm$1.1 & \textbf{92.8}$\pm$0.9 & \textbf{89.9}$\pm$0.7 & \textbf{89.7}$\pm$0.8\\
QM8(E1-CAM) & \textbf{96.3}$\pm$1.1 & \textbf{96.5}$\pm$0.9 & \textbf{92.7}$\pm$0.9 & \textbf{93.2}$\pm$0.8 & \textbf{96.2}$\pm$1.2 & \textbf{96.4}$\pm$0.9 & \textbf{92.3}$\pm$0.8 & \textbf{92.8}$\pm$0.6 & \textbf{89.9}$\pm$0.5 & \textbf{90.2}$\pm$0.4\\
QM8(E2-CAM) & \textbf{95.3}$\pm$1.0 & \textbf{95.6}$\pm$1.1 & \textbf{92.1}$\pm$0.7 & \textbf{92.6}$\pm$0.8 & \textbf{95.4}$\pm$0.8 & \textbf{95.6}$\pm$0.9 & \textbf{92.1}$\pm$0.5 & \textbf{92.5}$\pm$0.7 & \textbf{90.0}$\pm$0.5 & \textbf{90.0}$\pm$0.6\\
QM8(f1-CAM) & \textbf{93.7}$\pm$1.0 & \textbf{94.7}$\pm$1.2 & \textbf{92.2}$\pm$0.6 & \textbf{93.5}$\pm$0.8 & \textbf{93.3}$\pm$0.9 & \textbf{93.8}$\pm$1.2 & \textbf{91.7}$\pm$0.6 & \textbf{92.2}$\pm$0.7 & \textbf{89.8}$\pm$0.7 & \textbf{89.9}$\pm$0.7\\
QM8(f2-CAM) & \textbf{94.7}$\pm$1.1 & \textbf{95.0}$\pm$0.8 & \textbf{92.7}$\pm$0.7 & \textbf{93.4}$\pm$0.6 & \textbf{94.5}$\pm$1.1 & \textbf{95.2}$\pm$0.9 & \textbf{92.5}$\pm$0.7 & \textbf{93.5}$\pm$0.7 & \textbf{90.0}$\pm$0.9 & \textbf{90.3}$\pm$0.8\\
QM9(mu) & \textbf{92.5}$\pm$1.7 & \textbf{93.2}$\pm$1.6 & \textbf{90.4}$\pm$0.4 & \textbf{91.3}$\pm$0.5 & \textbf{92.6}$\pm$1.7 & \textbf{93.3}$\pm$1.5 & \textbf{90.6}$\pm$0.4 & \textbf{91.4}$\pm$0.4 & \textbf{90.1}$\pm$0.2 & \textbf{90.0}$\pm$0.3\\
QM9(alpha) & \textbf{94.7}$\pm$1.5 & \textbf{94.6}$\pm$1.6 & \textbf{90.5}$\pm$0.4 & 89.7$\pm$0.4 & \textbf{94.5}$\pm$1.5 & \textbf{94.6}$\pm$1.6 & \textbf{90.1}$\pm$0.3 & 89.7$\pm$0.4 & \textbf{90.0}$\pm$0.2 & \textbf{89.9}$\pm$0.2\\
QM9(homo) & \textbf{92.3}$\pm$0.6 & \textbf{92.5}$\pm$0.7 & \textbf{90.4}$\pm$0.4 & \textbf{90.4}$\pm$0.2 & \textbf{92.4}$\pm$0.6 & \textbf{92.6}$\pm$0.7 & \textbf{90.6}$\pm$0.3 & \textbf{90.7}$\pm$0.3 & \textbf{90.0}$\pm$0.3 & \textbf{89.9}$\pm$0.3\\
QM9(lumo) & \textbf{93.5}$\pm$0.9 & \textbf{93.5}$\pm$1.0 & \textbf{90.5}$\pm$0.3 & \textbf{90.4}$\pm$0.3 & \textbf{93.5}$\pm$0.9 & \textbf{93.6}$\pm$0.9 & \textbf{90.7}$\pm$0.4 & \textbf{90.6}$\pm$0.3 & \textbf{90.1}$\pm$0.3 & \textbf{89.9}$\pm$0.4\\
QM9(gap) & \textbf{93.0}$\pm$1.2 & \textbf{93.1}$\pm$1.2 & \textbf{90.5}$\pm$0.3 & \textbf{90.4}$\pm$0.2 & \textbf{93.0}$\pm$1.1 & \textbf{93.2}$\pm$1.1 & \textbf{90.6}$\pm$0.3 & \textbf{90.7}$\pm$0.3 & \textbf{90.1}$\pm$0.2 & \textbf{89.9}$\pm$0.3\\
QM9(r2) & \textbf{92.6}$\pm$1.1 & \textbf{93.0}$\pm$1.2 & \textbf{90.1}$\pm$0.3 & \textbf{90.5}$\pm$0.3 & \textbf{92.6}$\pm$1.2 & \textbf{93.1}$\pm$1.1 & \textbf{90.4}$\pm$0.4 & \textbf{90.8}$\pm$0.5 & \textbf{89.9}$\pm$0.4 & \textbf{90.1}$\pm$0.3\\
QM9(zpve) & \textbf{95.3}$\pm$1.2 & \textbf{95.4}$\pm$1.2 & \textbf{90.5}$\pm$0.1 & \textbf{90.1}$\pm$0.3 & \textbf{95.2}$\pm$1.3 & \textbf{95.3}$\pm$1.4 & \textbf{90.3}$\pm$0.2 & \textbf{89.9}$\pm$0.2 & \textbf{90.0}$\pm$0.2 & \textbf{90.0}$\pm$0.2\\
QM9(u0) & \textbf{94.9}$\pm$1.5 & \textbf{94.8}$\pm$1.6 & \textbf{90.9}$\pm$0.5 & \textbf{90.6}$\pm$0.5 & \textbf{93.3}$\pm$1.3 & \textbf{93.4}$\pm$1.3 & \textbf{90.1}$\pm$0.2 & \textbf{90.0}$\pm$0.3 & \textbf{89.9}$\pm$0.2 & \textbf{90.1}$\pm$0.2\\
QM9(u298) & \textbf{95.1}$\pm$1.8 & \textbf{95.0}$\pm$1.9 & \textbf{90.9}$\pm$0.5 & \textbf{90.6}$\pm$0.5 & \textbf{93.6}$\pm$1.7 & \textbf{93.7}$\pm$1.8 & \textbf{90.0}$\pm$0.2 & \textbf{89.9}$\pm$0.3 & \textbf{89.9}$\pm$0.2 & \textbf{90.1}$\pm$0.2\\
QM9(h298) & \textbf{95.0}$\pm$1.6 & \textbf{94.9}$\pm$1.7 & \textbf{90.9}$\pm$0.5 & \textbf{90.6}$\pm$0.5 & \textbf{93.5}$\pm$1.4 & \textbf{93.6}$\pm$1.4 & \textbf{90.1}$\pm$0.2 & \textbf{90.0}$\pm$0.3 & \textbf{89.9}$\pm$0.2 & \textbf{90.1}$\pm$0.2\\
QM9(g298) & \textbf{94.9}$\pm$1.5 & \textbf{94.8}$\pm$1.6 & \textbf{90.8}$\pm$0.5 & \textbf{90.6}$\pm$0.5 & \textbf{93.3}$\pm$1.4 & \textbf{93.4}$\pm$1.3 & \textbf{90.0}$\pm$0.2 & \textbf{89.9}$\pm$0.3 & \textbf{89.9}$\pm$0.2 & \textbf{90.1}$\pm$0.2\\
QM9(cv) & \textbf{94.3}$\pm$1.4 & \textbf{94.4}$\pm$1.5 & \textbf{90.4}$\pm$0.3 & \textbf{89.9}$\pm$0.5 & \textbf{94.2}$\pm$1.5 & \textbf{94.4}$\pm$1.5 & \textbf{90.4}$\pm$0.4 & \textbf{90.1}$\pm$0.4 & \textbf{90.0}$\pm$0.2 & \textbf{90.0}$\pm$0.4\\

\bottomrule

\toprule
TCR&\\
\midrule
QM8(E1-CC2) & \textbf{94.4}$\pm$1.2 & \textbf{94.3}$\pm$1.4 & \textbf{89.6}$\pm$1.9 & 89.3$\pm$0.9 & \textbf{95.2}$\pm$0.9 & \textbf{94.3}$\pm$1.5 & \textbf{91.3}$\pm$1.2 & \textbf{90.2}$\pm$1.3 & 88.1$\pm$1.5 & 83.5$\pm$6.4\\
QM8(E2-CC2) & \textbf{95.5}$\pm$1.5 & \textbf{94.9}$\pm$1.7 & \textbf{89.6}$\pm$1.2 & 88.6$\pm$1.4 & \textbf{95.8}$\pm$1.6 & \textbf{95.6}$\pm$1.5 & \textbf{90.7}$\pm$1.7 & \textbf{90.1}$\pm$1.7 & 87.4$\pm$2.2 & 84.4$\pm$4.4\\
QM8(f1-CC2) & \textbf{94.5}$\pm$2.2 & \textbf{97.0}$\pm$1.7 & \textbf{90.4}$\pm$1.5 & \textbf{93.6}$\pm$1.4 & \textbf{94.9}$\pm$1.3 & \textbf{94.1}$\pm$2.0 & \textbf{91.4}$\pm$1.2 & \textbf{90.5}$\pm$1.8 & 85.0$\pm$1.8 & 80.1$\pm$2.8\\
QM8(f2-CC2) & \textbf{95.6}$\pm$2.5 & \textbf{96.2}$\pm$2.7 & \textbf{91.8}$\pm$1.6 & \textbf{92.3}$\pm$1.7 & \textbf{95.7}$\pm$2.6 & \textbf{95.2}$\pm$2.9 & \textbf{92.4}$\pm$1.6 & \textbf{91.4}$\pm$1.5 & 84.9$\pm$2.1 & 73.3$\pm$4.8\\
QM8(E1-PBE0) & \textbf{94.0}$\pm$1.4 & \textbf{94.2}$\pm$1.2 & \textbf{89.9}$\pm$1.8 & \textbf{89.9}$\pm$1.5 & \textbf{94.4}$\pm$0.8 & \textbf{94.2}$\pm$1.5 & \textbf{90.9}$\pm$1.2 & \textbf{90.3}$\pm$1.5 & 87.8$\pm$1.8 & 80.9$\pm$4.6\\
QM8(E2-PBE0) & \textbf{94.1}$\pm$1.6 & \textbf{93.8}$\pm$2.0 & \textbf{90.0}$\pm$1.6 & \textbf{89.7}$\pm$2.0 & \textbf{94.8}$\pm$1.4 & \textbf{94.2}$\pm$1.9 & \textbf{90.9}$\pm$1.7 & \textbf{90.5}$\pm$2.0 & 87.5$\pm$1.8 & 82.1$\pm$5.8\\
QM8(f1-PBE0) & \textbf{94.3}$\pm$2.3 & \textbf{97.3}$\pm$1.5 & \textbf{90.9}$\pm$3.0 & \textbf{94.4}$\pm$1.8 & \textbf{94.5}$\pm$2.0 & \textbf{95.2}$\pm$1.9 & \textbf{91.1}$\pm$2.7 & \textbf{91.8}$\pm$2.1 & 85.1$\pm$1.9 & 80.4$\pm$2.5\\
QM8(f2-PBE0) & \textbf{94.5}$\pm$3.4 & \textbf{95.6}$\pm$3.4 & \textbf{90.1}$\pm$2.4 & \textbf{90.9}$\pm$1.5 & \textbf{95.1}$\pm$3.2 & \textbf{95.1}$\pm$3.6 & \textbf{90.9}$\pm$2.4 & \textbf{90.2}$\pm$1.8 & 84.4$\pm$2.6 & 74.6$\pm$6.2\\
QM8(E1-PBE0.1) & \textbf{94.0}$\pm$1.3 & \textbf{94.2}$\pm$1.2 & \textbf{89.8}$\pm$1.5 & \textbf{89.9}$\pm$1.5 & \textbf{94.3}$\pm$1.2 & \textbf{93.9}$\pm$1.6 & \textbf{90.7}$\pm$1.4 & \textbf{90.3}$\pm$1.9 & 87.8$\pm$2.0 & 83.4$\pm$4.4\\
QM8(E2-PBE0.1) & \textbf{94.2}$\pm$1.4 & \textbf{93.8}$\pm$2.0 & \textbf{90.1}$\pm$1.6 & \textbf{89.7}$\pm$2.0 & \textbf{94.9}$\pm$1.5 & \textbf{94.3}$\pm$2.1 & \textbf{91.0}$\pm$1.5 & \textbf{90.4}$\pm$2.2 & 87.9$\pm$1.8 & 82.0$\pm$5.5\\
QM8(f1-PBE0.1) & \textbf{94.2}$\pm$2.3 & \textbf{97.3}$\pm$1.5 & \textbf{90.7}$\pm$2.8 & \textbf{94.4}$\pm$1.8 & \textbf{93.9}$\pm$2.1 & \textbf{95.2}$\pm$1.8 & \textbf{90.6}$\pm$2.7 & \textbf{91.7}$\pm$2.5 & 84.1$\pm$2.0 & 81.1$\pm$2.4\\
QM8(f2-PBE0.1) & \textbf{94.7}$\pm$3.2 & \textbf{95.6}$\pm$3.4 & \textbf{90.0}$\pm$2.1 & \textbf{90.9}$\pm$1.5 & \textbf{95.0}$\pm$3.2 & \textbf{95.0}$\pm$3.7 & \textbf{90.8}$\pm$2.3 & \textbf{90.3}$\pm$1.9 & 84.2$\pm$2.4 & 73.3$\pm$3.8\\
QM8(E1-CAM) & \textbf{94.0}$\pm$1.7 & \textbf{94.1}$\pm$1.7 & \textbf{89.1}$\pm$2.4 & \textbf{90.0}$\pm$1.3 & \textbf{94.3}$\pm$1.5 & \textbf{93.8}$\pm$1.9 & \textbf{90.2}$\pm$1.4 & \textbf{89.8}$\pm$1.7 & 86.5$\pm$1.4 & 81.4$\pm$5.8\\
QM8(E2-CAM) & \textbf{95.1}$\pm$1.4 & \textbf{95.0}$\pm$1.8 & \textbf{90.7}$\pm$1.2 & \textbf{90.4}$\pm$1.5 & \textbf{95.4}$\pm$1.4 & \textbf{95.4}$\pm$1.9 & \textbf{91.8}$\pm$1.0 & \textbf{91.1}$\pm$1.5 & 88.1$\pm$1.7 & 83.5$\pm$5.0\\
QM8(f1-CAM) & \textbf{95.3}$\pm$2.3 & \textbf{98.8}$\pm$1.2 & \textbf{92.1}$\pm$1.4 & \textbf{96.2}$\pm$0.7 & \textbf{95.1}$\pm$1.6 & \textbf{95.9}$\pm$1.4 & \textbf{92.3}$\pm$0.8 & \textbf{93.0}$\pm$1.3 & 86.7$\pm$1.1 & 81.9$\pm$3.3\\
QM8(f2-CAM) & \textbf{95.7}$\pm$1.7 & \textbf{96.8}$\pm$1.8 & \textbf{91.5}$\pm$0.9 & \textbf{92.8}$\pm$0.9 & \textbf{95.9}$\pm$1.7 & \textbf{96.0}$\pm$1.9 & \textbf{91.7}$\pm$1.5 & \textbf{91.2}$\pm$1.5 & 85.0$\pm$2.0 & 73.5$\pm$4.6\\
QM9(mu) & 83.2$\pm$4.7 & 82.9$\pm$4.6 & 77.4$\pm$1.5 & 77.1$\pm$1.7 & 87.0$\pm$3.5 & 86.0$\pm$3.4 & 83.1$\pm$1.5 & 81.9$\pm$1.1 & 79.8$\pm$1.5 & 64.7$\pm$4.2\\
QM9(alpha) & \textbf{96.7}$\pm$1.3 & \textbf{96.7}$\pm$1.5 & 88.9$\pm$0.7 & 87.9$\pm$0.7 & \textbf{97.4}$\pm$1.1 & \textbf{97.5}$\pm$1.2 & \textbf{90.7}$\pm$0.9 & \textbf{90.3}$\pm$0.9 & 87.3$\pm$0.8 & 69.9$\pm$2.3\\
QM9(homo) & \textbf{91.7}$\pm$1.4 & \textbf{91.4}$\pm$1.5 & 87.2$\pm$0.8 & 86.7$\pm$1.0 & \textbf{93.0}$\pm$1.0 & \textbf{92.9}$\pm$0.9 & \textbf{89.7}$\pm$0.7 & \textbf{89.4}$\pm$1.0 & 86.5$\pm$0.5 & \textbf{89.8}$\pm$2.7\\
QM9(lumo) & \textbf{93.4}$\pm$1.3 & \textbf{93.5}$\pm$1.2 & 88.2$\pm$0.8 & 88.1$\pm$0.7 & \textbf{94.7}$\pm$0.7 & \textbf{94.7}$\pm$0.8 & \textbf{91.1}$\pm$0.9 & \textbf{90.7}$\pm$1.0 & \textbf{89.7}$\pm$0.9 & 87.5$\pm$1.9\\
QM9(gap) & \textbf{92.0}$\pm$1.8 & \textbf{91.9}$\pm$2.0 & 87.6$\pm$0.6 & 86.9$\pm$0.9 & \textbf{93.7}$\pm$1.4 & \textbf{93.6}$\pm$1.5 & \textbf{90.3}$\pm$0.8 & \textbf{90.0}$\pm$1.0 & 87.9$\pm$0.8 & 88.1$\pm$1.3\\
QM9(r2) & \textbf{94.1}$\pm$2.1 & \textbf{93.7}$\pm$2.0 & 88.2$\pm$0.8 & 88.0$\pm$1.2 & \textbf{95.3}$\pm$1.7 & \textbf{95.0}$\pm$1.7 & \textbf{90.9}$\pm$0.7 & \textbf{90.6}$\pm$1.2 & 87.6$\pm$0.8 & 67.0$\pm$4.3\\
QM9(zpve) & \textbf{97.0}$\pm$1.0 & \textbf{97.1}$\pm$0.9 & \textbf{90.4}$\pm$0.7 & \textbf{90.1}$\pm$0.7 & \textbf{96.7}$\pm$1.2 & \textbf{96.9}$\pm$1.2 & \textbf{91.2}$\pm$0.9 & \textbf{90.9}$\pm$0.8 & \textbf{90.5}$\pm$0.6 & \textbf{88.3}$\pm$4.5\\
QM9(u0) & \textbf{97.3}$\pm$1.5 & \textbf{97.2}$\pm$1.6 & \textbf{91.2}$\pm$0.9 & \textbf{90.7}$\pm$0.9 & \textbf{95.9}$\pm$1.5 & \textbf{96.1}$\pm$1.7 & \textbf{90.7}$\pm$0.5 & \textbf{90.4}$\pm$0.8 & 84.1$\pm$1.0 & 80.1$\pm$3.2\\
QM9(u298) & \textbf{97.4}$\pm$1.7 & \textbf{97.4}$\pm$1.8 & \textbf{91.2}$\pm$0.8 & \textbf{90.7}$\pm$0.9 & \textbf{96.2}$\pm$1.8 & \textbf{96.2}$\pm$1.9 & \textbf{90.6}$\pm$0.4 & \textbf{90.4}$\pm$0.8 & 84.0$\pm$0.9 & 79.8$\pm$3.3\\
QM9(h298) & \textbf{97.4}$\pm$1.7 & \textbf{97.3}$\pm$1.7 & \textbf{91.2}$\pm$0.8 & \textbf{90.7}$\pm$0.9 & \textbf{96.1}$\pm$1.8 & \textbf{96.1}$\pm$1.9 & \textbf{90.7}$\pm$0.4 & \textbf{90.4}$\pm$0.7 & 84.0$\pm$1.1 & 80.3$\pm$2.9\\
QM9(g298) & \textbf{97.3}$\pm$1.6 & \textbf{97.2}$\pm$1.7 & \textbf{91.1}$\pm$0.7 & \textbf{90.7}$\pm$0.9 & \textbf{96.0}$\pm$1.6 & \textbf{96.0}$\pm$1.7 & \textbf{90.7}$\pm$0.4 & \textbf{90.3}$\pm$0.7 & 84.0$\pm$1.1 & 80.0$\pm$2.3\\
QM9(cv) & \textbf{95.9}$\pm$1.8 & \textbf{95.9}$\pm$2.0 & 89.2$\pm$0.7 & 88.2$\pm$1.0 & \textbf{96.5}$\pm$1.5 & \textbf{96.6}$\pm$1.5 & \textbf{91.0}$\pm$0.7 & \textbf{90.8}$\pm$0.7 & 87.8$\pm$0.6 & 80.4$\pm$7.7\\

\bottomrule
\end{tabular}
\label{appendix:table:exp:real:qm_cov}
\end{tiny}
\end{center}
\end{table}
}

\def \TableQMDisc{
\begin{table}[ht]
\captionof{table}{
AUROC and MAD for QM8 and QM9 sub-tasks.
The best AUROCs are in bold, and AUROCs \textbf{not} significantly higher than 50 at $p=0.05$ are underscored.
For MAD, the best MADs, if significantly better than the second baseline at $p=0.05$, are in bold.
}
\begin{center}
\begin{tiny}
\setlength\tabcolsep{1.5pt}
\begin{tabular}{@{}l|cccccccc|cr@{}}

\toprule
& \multicolumn{8}{c|}{\methodname} &  \multicolumn{2}{c}{Conformal Baselines}\\
& \multicolumn{4}{c|}{$\hat{y}^{KR}$} & \multicolumn{4}{c|}{$\hat{y}^{NN}$}  & & \\
& \multicolumn{2}{c|}{No-smooth} & \multicolumn{2}{c|}{Smooth} & \multicolumn{2}{c|}{No-smooth} & \multicolumn{2}{c|}{Smooth} & & \\
\midrule
AUROC & MN & base & MN & base& MN & base& MN & \boxed{base}& MADSplit & CQR\\
\midrule
QM8(E1-CC2) & 64.3$\pm$1.4 & 61.8$\pm$1.5 & 67.9$\pm$0.9 & 64.9$\pm$1.0 & 66.2$\pm$1.0 & 63.2$\pm$1.2 & \textbf{68.2}$\pm$0.9 & 62.9$\pm$1.4 & 67.7$\pm$0.8 & 57.9$\pm$3.8\\
QM8(E2-CC2) & 62.5$\pm$1.5 & 60.2$\pm$1.3 & 66.7$\pm$0.8 & 63.2$\pm$1.2 & 64.3$\pm$1.2 & 61.1$\pm$1.2 & \textbf{66.8}$\pm$1.1 & 61.7$\pm$1.2 & 66.2$\pm$1.0 & 56.8$\pm$3.9\\
QM8(f1-CC2) & 83.8$\pm$2.1 & 85.7$\pm$2.5 & 85.7$\pm$1.0 & \textbf{87.9}$\pm$0.8 & 80.2$\pm$1.0 & 80.5$\pm$1.2 & 80.5$\pm$1.2 & 80.1$\pm$1.4 & 80.0$\pm$1.0 & 70.5$\pm$2.9\\
QM8(f2-CC2) & 81.4$\pm$2.6 & 82.9$\pm$2.2 & 84.3$\pm$1.3 & \textbf{86.0}$\pm$0.7 & 81.7$\pm$1.0 & 82.1$\pm$0.7 & 82.1$\pm$0.8 & 82.3$\pm$0.5 & 80.9$\pm$1.1 & 78.5$\pm$5.7\\
QM8(E1-PBE0) & 64.7$\pm$1.6 & 61.8$\pm$1.6 & 67.6$\pm$0.8 & 63.9$\pm$1.6 & 66.2$\pm$1.1 & 62.6$\pm$1.1 & \textbf{68.3}$\pm$0.9 & 62.6$\pm$1.1 & 67.6$\pm$0.9 & 55.2$\pm$5.0\\
QM8(E2-PBE0) & 63.9$\pm$1.4 & 60.2$\pm$1.2 & 66.0$\pm$1.0 & 61.8$\pm$1.2 & 65.1$\pm$1.3 & 61.3$\pm$1.4 & \textbf{66.6}$\pm$1.2 & 61.2$\pm$1.5 & 66.3$\pm$1.2 & 55.9$\pm$2.2\\
QM8(f1-PBE0) & 83.8$\pm$2.2 & 86.0$\pm$2.0 & 85.2$\pm$1.4 & \textbf{87.8}$\pm$0.7 & 79.0$\pm$1.1 & 78.8$\pm$1.0 & 79.1$\pm$1.1 & 78.1$\pm$1.7 & 79.0$\pm$1.0 & 67.2$\pm$2.7\\
QM8(f2-PBE0) & 80.1$\pm$2.3 & 82.1$\pm$2.8 & 82.8$\pm$1.1 & \textbf{85.2}$\pm$0.9 & 80.8$\pm$0.9 & 81.0$\pm$0.8 & 81.2$\pm$1.2 & 81.1$\pm$1.1 & 80.3$\pm$1.0 & 79.9$\pm$3.1\\
QM8(E1-PBE0.1) & 64.8$\pm$1.6 & 61.8$\pm$1.6 & 67.6$\pm$0.9 & 63.9$\pm$1.6 & 66.3$\pm$1.4 & 62.6$\pm$1.1 & \textbf{68.2}$\pm$0.7 & 62.4$\pm$0.8 & 67.4$\pm$0.8 & 56.9$\pm$2.5\\
QM8(E2-PBE0.1) & 63.8$\pm$1.4 & 60.2$\pm$1.2 & 65.8$\pm$1.0 & 61.8$\pm$1.2 & 64.6$\pm$1.3 & 60.8$\pm$1.5 & \textbf{66.1}$\pm$0.9 & 60.8$\pm$1.5 & 66.0$\pm$1.1 & 55.2$\pm$2.3\\
QM8(f1-PBE0.1) & 83.8$\pm$2.2 & 86.0$\pm$2.0 & 85.1$\pm$1.4 & \textbf{87.8}$\pm$0.7 & 79.5$\pm$1.5 & 79.3$\pm$1.2 & 79.7$\pm$1.4 & 78.7$\pm$1.8 & 79.4$\pm$1.3 & 68.1$\pm$3.8\\
QM8(f2-PBE0.1) & 80.1$\pm$2.3 & 82.1$\pm$2.8 & 82.9$\pm$1.1 & \textbf{85.2}$\pm$0.9 & 81.1$\pm$0.7 & 81.3$\pm$0.8 & 81.6$\pm$1.0 & 81.4$\pm$1.1 & 80.6$\pm$0.9 & 76.7$\pm$6.9\\
QM8(E1-CAM) & 63.4$\pm$1.6 & 61.4$\pm$1.2 & 67.7$\pm$1.1 & 64.8$\pm$1.0 & 65.4$\pm$1.2 & 62.2$\pm$1.0 & \textbf{68.2}$\pm$1.0 & 63.2$\pm$0.7 & 67.4$\pm$1.0 & 58.0$\pm$3.9\\
QM8(E2-CAM) & 63.9$\pm$2.2 & 61.2$\pm$1.7 & 66.6$\pm$0.8 & 63.3$\pm$1.8 & 65.1$\pm$2.1 & 61.7$\pm$1.6 & \textbf{66.9}$\pm$1.4 & 62.4$\pm$2.1 & 66.5$\pm$1.5 & 56.8$\pm$3.3\\
QM8(f1-CAM) & 85.8$\pm$1.5 & 87.7$\pm$1.5 & 87.2$\pm$1.0 & \textbf{89.5}$\pm$0.5 & 78.7$\pm$1.2 & 79.2$\pm$1.2 & 78.7$\pm$1.3 & 78.8$\pm$1.2 & 78.4$\pm$1.2 & 73.1$\pm$2.0\\
QM8(f2-CAM) & 81.5$\pm$2.1 & 83.0$\pm$2.2 & 84.8$\pm$1.0 & \textbf{86.8}$\pm$0.8 & 82.7$\pm$1.1 & 82.9$\pm$1.0 & 83.2$\pm$1.0 & 83.3$\pm$1.1 & 82.1$\pm$1.0 & 81.4$\pm$2.3\\
QM9(mu) & 71.7$\pm$0.6 & 67.6$\pm$0.5 & 71.8$\pm$1.0 & 66.6$\pm$1.7 & 72.6$\pm$1.1 & 68.2$\pm$0.5 & \textbf{73.9}$\pm$0.7 & 68.0$\pm$1.1 & 73.7$\pm$0.7 & 57.4$\pm$3.4\\
QM9(alpha) & 61.6$\pm$1.2 & 60.3$\pm$1.3 & \textbf{66.2}$\pm$1.3 & 65.9$\pm$1.9 & 65.3$\pm$0.8 & 63.2$\pm$0.9 & 65.3$\pm$0.6 & 61.3$\pm$1.4 & 64.0$\pm$0.5 & \underline{45.9}$\pm$7.7\\
QM9(homo) & 61.2$\pm$0.9 & 58.3$\pm$0.5 & 61.9$\pm$0.5 & 58.4$\pm$0.8 & 62.2$\pm$0.4 & 58.6$\pm$0.5 & \textbf{62.9}$\pm$0.3 & 58.1$\pm$0.9 & 62.6$\pm$0.4 & \underline{38.7}$\pm$20.8\\
QM9(lumo) & 60.6$\pm$0.7 & 58.6$\pm$0.7 & 61.5$\pm$0.7 & 59.3$\pm$0.7 & 61.2$\pm$0.8 & 58.5$\pm$0.4 & \textbf{62.4}$\pm$0.4 & 58.1$\pm$0.4 & 62.2$\pm$0.4 & \underline{58.4}$\pm$20.1\\
QM9(gap) & 62.3$\pm$1.0 & 60.4$\pm$1.0 & 62.8$\pm$0.5 & 60.5$\pm$0.8 & 62.8$\pm$0.9 & 60.1$\pm$0.7 & \textbf{63.5}$\pm$0.5 & 59.6$\pm$0.6 & 63.1$\pm$0.5 & \underline{60.2}$\pm$25.0\\
QM9(r2) & 67.8$\pm$1.1 & 64.1$\pm$1.1 & 68.5$\pm$1.2 & 65.0$\pm$1.0 & 69.3$\pm$0.7 & 64.7$\pm$0.6 & \textbf{69.8}$\pm$0.5 & 63.1$\pm$0.9 & 69.5$\pm$0.5 & 63.8$\pm$0.9\\
QM9(zpve) & 60.5$\pm$1.7 & 60.7$\pm$1.7 & 65.4$\pm$0.9 & 65.6$\pm$1.4 & 62.7$\pm$0.5 & 61.8$\pm$0.7 & 61.3$\pm$0.3 & 58.8$\pm$0.6 & 60.4$\pm$0.4 & \textbf{67.5}$\pm$19.1\\
QM9(u0) & 77.6$\pm$3.4 & \textbf{79.2}$\pm$2.7 & 68.9$\pm$2.4 & 72.9$\pm$1.6 & 68.3$\pm$0.5 & 67.7$\pm$0.4 & 67.8$\pm$0.6 & 66.1$\pm$1.0 & 64.6$\pm$0.9 & 54.9$\pm$2.0\\
QM9(u298) & 77.6$\pm$3.3 & \textbf{79.1}$\pm$2.6 & 68.8$\pm$2.5 & 72.9$\pm$1.7 & 68.4$\pm$0.6 & 67.9$\pm$0.6 & 67.9$\pm$0.8 & 66.0$\pm$1.2 & 64.7$\pm$1.0 & 55.8$\pm$2.2\\
QM9(h298) & 77.5$\pm$3.2 & \textbf{79.1}$\pm$2.6 & 68.9$\pm$2.5 & 72.9$\pm$1.6 & 68.4$\pm$0.7 & 67.9$\pm$0.7 & 67.9$\pm$0.8 & 66.2$\pm$1.2 & 64.7$\pm$0.9 & 55.7$\pm$2.0\\
QM9(g298) & 77.6$\pm$3.3 & \textbf{79.2}$\pm$2.7 & 68.9$\pm$2.4 & 72.9$\pm$1.7 & 68.6$\pm$0.7 & 68.0$\pm$0.7 & 68.0$\pm$0.8 & 66.1$\pm$1.1 & 64.8$\pm$0.9 & 54.5$\pm$1.8\\
QM9(cv) & 62.6$\pm$1.0 & 60.8$\pm$0.9 & 65.0$\pm$1.1 & 64.0$\pm$0.9 & 64.6$\pm$1.1 & 62.3$\pm$0.8 & \textbf{65.3}$\pm$0.7 & 60.9$\pm$0.8 & 64.6$\pm$0.7 & \underline{47.5}$\pm$9.1\\

\bottomrule

\toprule
MAD&\\
\midrule
QM8(E1-CC2) & \textbf{0.01}$\pm$0.00 & \textbf{0.01}$\pm$0.00 & 0.01$\pm$0.00 & 0.01$\pm$0.00 & 0.01$\pm$0.00 & 0.01$\pm$0.00 & 0.01$\pm$0.00 & 0.01$\pm$0.00 & 0.01$\pm$0.00 & 0.02$\pm$0.00\\
QM8(E2-CC2) & \textbf{0.01}$\pm$0.00 & \textbf{0.01}$\pm$0.00 & 0.01$\pm$0.00 & 0.01$\pm$0.00 & 0.01$\pm$0.00 & 0.01$\pm$0.00 & 0.01$\pm$0.00 & 0.01$\pm$0.00 & 0.01$\pm$0.00 & 0.02$\pm$0.00\\
QM8(f1-CC2) & \textbf{0.01}$\pm$0.00 & \textbf{0.01}$\pm$0.00 & 0.01$\pm$0.00 & 0.01$\pm$0.00 & 0.01$\pm$0.00 & 0.01$\pm$0.00 & 0.01$\pm$0.00 & 0.01$\pm$0.00 & 0.01$\pm$0.00 & 0.03$\pm$0.00\\
QM8(f2-CC2) & \textbf{0.03}$\pm$0.00 & \textbf{0.03}$\pm$0.00 & 0.03$\pm$0.00 & 0.03$\pm$0.00 & 0.03$\pm$0.00 & 0.03$\pm$0.00 & 0.03$\pm$0.00 & 0.03$\pm$0.00 & 0.03$\pm$0.00 & 0.06$\pm$0.01\\
QM8(E1-PBE0) & \textbf{0.01}$\pm$0.00 & \textbf{0.01}$\pm$0.00 & 0.01$\pm$0.00 & 0.01$\pm$0.00 & 0.01$\pm$0.00 & 0.01$\pm$0.00 & 0.01$\pm$0.00 & 0.01$\pm$0.00 & 0.01$\pm$0.00 & 0.02$\pm$0.00\\
QM8(E2-PBE0) & \textbf{0.01}$\pm$0.00 & \textbf{0.01}$\pm$0.00 & 0.01$\pm$0.00 & 0.01$\pm$0.00 & 0.01$\pm$0.00 & 0.01$\pm$0.00 & 0.01$\pm$0.00 & 0.01$\pm$0.00 & 0.01$\pm$0.00 & 0.02$\pm$0.00\\
QM8(f1-PBE0) & \textbf{0.01}$\pm$0.00 & \textbf{0.01}$\pm$0.00 & 0.01$\pm$0.00 & 0.01$\pm$0.00 & 0.01$\pm$0.00 & 0.01$\pm$0.00 & 0.01$\pm$0.00 & 0.01$\pm$0.00 & 0.01$\pm$0.00 & 0.03$\pm$0.00\\
QM8(f2-PBE0) & 0.02$\pm$0.00 & 0.02$\pm$0.00 & 0.02$\pm$0.00 & 0.02$\pm$0.00 & 0.02$\pm$0.00 & 0.02$\pm$0.00 & 0.02$\pm$0.00 & 0.02$\pm$0.00 & 0.02$\pm$0.00 & 0.05$\pm$0.01\\
QM8(E1-PBE0.1) & \textbf{0.01}$\pm$0.00 & \textbf{0.01}$\pm$0.00 & 0.01$\pm$0.00 & 0.01$\pm$0.00 & 0.01$\pm$0.00 & 0.01$\pm$0.00 & 0.01$\pm$0.00 & 0.01$\pm$0.00 & 0.01$\pm$0.00 & 0.02$\pm$0.00\\
QM8(E2-PBE0.1) & \textbf{0.01}$\pm$0.00 & \textbf{0.01}$\pm$0.00 & 0.01$\pm$0.00 & 0.01$\pm$0.00 & 0.01$\pm$0.00 & 0.01$\pm$0.00 & 0.01$\pm$0.00 & 0.01$\pm$0.00 & 0.01$\pm$0.00 & 0.02$\pm$0.00\\
QM8(f1-PBE0.1) & \textbf{0.01}$\pm$0.00 & \textbf{0.01}$\pm$0.00 & 0.01$\pm$0.00 & 0.01$\pm$0.00 & 0.01$\pm$0.00 & 0.01$\pm$0.00 & 0.01$\pm$0.00 & 0.01$\pm$0.00 & 0.01$\pm$0.00 & 0.03$\pm$0.01\\
QM8(f2-PBE0.1) & 0.02$\pm$0.00 & 0.02$\pm$0.00 & 0.02$\pm$0.00 & 0.02$\pm$0.00 & 0.02$\pm$0.00 & 0.02$\pm$0.00 & 0.02$\pm$0.00 & 0.02$\pm$0.00 & 0.02$\pm$0.00 & 0.05$\pm$0.01\\
QM8(E1-CAM) & \textbf{0.01}$\pm$0.00 & \textbf{0.01}$\pm$0.00 & 0.01$\pm$0.00 & 0.01$\pm$0.00 & 0.01$\pm$0.00 & 0.01$\pm$0.00 & 0.01$\pm$0.00 & 0.01$\pm$0.00 & 0.01$\pm$0.00 & 0.02$\pm$0.01\\
QM8(E2-CAM) & \textbf{0.01}$\pm$0.00 & \textbf{0.01}$\pm$0.00 & 0.01$\pm$0.00 & 0.01$\pm$0.00 & 0.01$\pm$0.00 & 0.01$\pm$0.00 & 0.01$\pm$0.00 & 0.01$\pm$0.00 & 0.01$\pm$0.00 & 0.02$\pm$0.00\\
QM8(f1-CAM) & \textbf{0.01}$\pm$0.00 & \textbf{0.01}$\pm$0.00 & 0.01$\pm$0.00 & 0.01$\pm$0.00 & 0.01$\pm$0.00 & 0.01$\pm$0.00 & 0.01$\pm$0.00 & 0.01$\pm$0.00 & 0.01$\pm$0.00 & 0.04$\pm$0.01\\
QM8(f2-CAM) & \textbf{0.02}$\pm$0.00 & \textbf{0.02}$\pm$0.00 & 0.02$\pm$0.00 & 0.02$\pm$0.00 & 0.02$\pm$0.00 & 0.02$\pm$0.00 & 0.02$\pm$0.00 & 0.02$\pm$0.00 & 0.02$\pm$0.00 & 0.05$\pm$0.01\\
QM9(mu) & 0.49$\pm$0.01 & 0.49$\pm$0.01 & 0.51$\pm$0.01 & 0.51$\pm$0.01 & \textbf{0.48}$\pm$0.00 & \textbf{0.48}$\pm$0.00 & \textbf{0.48}$\pm$0.00 & \textbf{0.48}$\pm$0.00 & \textbf{0.48}$\pm$0.00 & 2.01$\pm$1.24\\
QM9(alpha) & 0.69$\pm$0.03 & 0.69$\pm$0.03 & 1.10$\pm$0.07 & 1.10$\pm$0.07 & 0.70$\pm$0.01 & 0.70$\pm$0.01 & 0.70$\pm$0.01 & 0.70$\pm$0.01 & 0.70$\pm$0.01 & 9.79$\pm$0.83\\
QM9(homo) & 0.00$\pm$0.00 & 0.00$\pm$0.00 & 0.00$\pm$0.00 & 0.00$\pm$0.00 & \textbf{0.00}$\pm$0.00 & \textbf{0.00}$\pm$0.00 & \textbf{0.00}$\pm$0.00 & \textbf{0.00}$\pm$0.00 & \textbf{0.00}$\pm$0.00 & 2.13$\pm$2.01\\
QM9(lumo) & 0.00$\pm$0.00 & 0.00$\pm$0.00 & 0.01$\pm$0.00 & 0.01$\pm$0.00 & \textbf{0.00}$\pm$0.00 & \textbf{0.00}$\pm$0.00 & \textbf{0.00}$\pm$0.00 & \textbf{0.00}$\pm$0.00 & \textbf{0.00}$\pm$0.00 & 5.65$\pm$3.33\\
QM9(gap) & 0.01$\pm$0.00 & 0.01$\pm$0.00 & 0.01$\pm$0.00 & 0.01$\pm$0.00 & \textbf{0.01}$\pm$0.00 & \textbf{0.01}$\pm$0.00 & \textbf{0.01}$\pm$0.00 & \textbf{0.01}$\pm$0.00 & \textbf{0.01}$\pm$0.00 & 4.34$\pm$3.35\\
QM9(r2) & 35.56$\pm$1.29 & 35.56$\pm$1.29 & 42.12$\pm$1.56 & 42.12$\pm$1.56 & \textbf{33.53}$\pm$0.36 & \textbf{33.53}$\pm$0.36 & \textbf{33.53}$\pm$0.36 & \textbf{33.53}$\pm$0.36 & \textbf{33.53}$\pm$0.36 & 188.64$\pm$9.50\\
QM9(zpve) & \textbf{0.00}$\pm$0.00 & \textbf{0.00}$\pm$0.00 & 0.00$\pm$0.00 & 0.00$\pm$0.00 & 0.00$\pm$0.00 & 0.00$\pm$0.00 & 0.00$\pm$0.00 & 0.00$\pm$0.00 & 0.00$\pm$0.00 & 4.42$\pm$3.93\\
QM9(u0) & \textbf{1.46}$\pm$0.18 & \textbf{1.46}$\pm$0.18 & 3.69$\pm$0.33 & 3.69$\pm$0.33 & 2.29$\pm$0.07 & 2.29$\pm$0.07 & 2.29$\pm$0.07 & 2.29$\pm$0.07 & 2.29$\pm$0.07 & 40.85$\pm$3.49\\
QM9(u298) & \textbf{1.46}$\pm$0.18 & \textbf{1.46}$\pm$0.18 & 3.69$\pm$0.33 & 3.69$\pm$0.33 & 2.29$\pm$0.06 & 2.29$\pm$0.06 & 2.29$\pm$0.06 & 2.29$\pm$0.06 & 2.29$\pm$0.06 & 40.64$\pm$3.81\\
QM9(h298) & \textbf{1.45}$\pm$0.18 & \textbf{1.45}$\pm$0.18 & 3.69$\pm$0.33 & 3.69$\pm$0.33 & 2.29$\pm$0.06 & 2.29$\pm$0.06 & 2.29$\pm$0.06 & 2.29$\pm$0.06 & 2.29$\pm$0.06 & 40.84$\pm$3.03\\
QM9(g298) & \textbf{1.46}$\pm$0.18 & \textbf{1.46}$\pm$0.18 & 3.70$\pm$0.33 & 3.70$\pm$0.33 & 2.29$\pm$0.07 & 2.29$\pm$0.07 & 2.29$\pm$0.07 & 2.29$\pm$0.07 & 2.29$\pm$0.07 & 41.02$\pm$3.81\\
QM9(cv) & 0.34$\pm$0.01 & 0.34$\pm$0.01 & 0.47$\pm$0.02 & 0.47$\pm$0.02 & \textbf{0.33}$\pm$0.01 & \textbf{0.33}$\pm$0.01 & \textbf{0.33}$\pm$0.01 & \textbf{0.33}$\pm$0.01 & \textbf{0.33}$\pm$0.01 & 5.03$\pm$1.55\\
\bottomrule
\end{tabular}
\label{appendix:table:exp:real:qm_disc}
\end{tiny}
\end{center}
\end{table}
}

\def \TableQMWidth{
\begin{sidewaystable}[ht]
\captionof{table}{
Average width for 50\% and 90\% PIs of different methods. 
Narrowest PIs are in bold, and further underscored if significantly narrower than the second best (at $p=0.05$).
As a reminder, there are 4357 data in QM8's test set and 26744 in QM9's.
Among these valid methods, \methodname's 50\% PIs are the narrowest, and still competitive at 90\% despite satisfying a stronger coverage requirement.
}
\begin{center}
\begin{tiny}
\setlength\tabcolsep{2pt}
\begin{tabular}{lc|cccccccc|cr}

\toprule

& & \multicolumn{8}{c|}{\methodname} &  \multicolumn{2}{c}{Conformal Baselines}\\
& & \multicolumn{4}{c|}{$\hat{y}^{KR}$} & \multicolumn{4}{c|}{$\hat{y}^{NN}$}  & & \\
& & \multicolumn{2}{c|}{No-smooth} & \multicolumn{2}{c|}{Smooth} & \multicolumn{2}{c|}{No-smooth} & \multicolumn{2}{c|}{Smooth} & & \\
\midrule
Width @ 50\% & \# finite & MN & base & MN & base& MN & base& MN & \boxed{base}& MADSplit & CQR\\
\midrule
QM8(E1-CC2) & 3791.5$\pm$189.4 & 1.2e-02$\pm$4.0e-04 & 1.1e-02$\pm$3.0e-04 & 1.1e-02$\pm$6.0e-04 & 9.8e-03$\pm$5.0e-04 & 1.3e-02$\pm$5.0e-04 & 1.2e-02$\pm$4.0e-04 & 1.1e-02$\pm$5.0e-04 & \textbf{9.6e-03}$\pm$4.0e-04 & 1.0e-02$\pm$4.0e-04 & 3.2e-02$\pm$4.1e-03\\
QM8(E2-CC2) & 3829.8$\pm$126.6 & 1.4e-02$\pm$7.0e-04 & 1.3e-02$\pm$5.0e-04 & 1.3e-02$\pm$4.0e-04 & 1.2e-02$\pm$4.0e-04 & 1.5e-02$\pm$9.0e-04 & 1.4e-02$\pm$7.0e-04 & 1.2e-02$\pm$3.0e-04 & \underline{\textbf{1.1e-02}}$\pm$2.0e-04 & 1.2e-02$\pm$3.0e-04 & 2.8e-02$\pm$3.3e-03\\
QM8(f1-CC2) & 4066.0$\pm$179.7 & 2.1e-02$\pm$1.6e-03 & 1.9e-02$\pm$1.2e-03 & 1.8e-02$\pm$2.1e-03 & \textbf{1.6e-02}$\pm$1.6e-03 & 2.4e-02$\pm$1.2e-03 & 2.1e-02$\pm$1.0e-03 & 1.9e-02$\pm$2.1e-03 & 1.6e-02$\pm$1.4e-03 & 1.8e-02$\pm$1.6e-03 & 3.0e-02$\pm$5.4e-03\\
QM8(f2-CC2) & 3998.1$\pm$231.6 & 5.2e-02$\pm$3.1e-03 & 4.6e-02$\pm$2.5e-03 & 4.6e-02$\pm$4.9e-03 & \textbf{4.1e-02}$\pm$4.3e-03 & 5.5e-02$\pm$3.2e-03 & 4.9e-02$\pm$3.2e-03 & 4.7e-02$\pm$5.1e-03 & 4.2e-02$\pm$4.7e-03 & 4.1e-02$\pm$3.2e-03 & 5.7e-02$\pm$1.1e-02\\
QM8(E1-PBE0) & 3932.1$\pm$133.8 & 1.2e-02$\pm$3.0e-04 & 1.1e-02$\pm$2.0e-04 & 1.1e-02$\pm$6.0e-04 & \textbf{9.6e-03}$\pm$3.0e-04 & 1.3e-02$\pm$5.0e-04 & 1.1e-02$\pm$4.0e-04 & 1.1e-02$\pm$3.0e-04 & 9.6e-03$\pm$2.0e-04 & 1.1e-02$\pm$3.0e-04 & 3.4e-02$\pm$9.1e-03\\
QM8(E2-PBE0) & 4063.2$\pm$133.4 & 1.3e-02$\pm$4.0e-04 & 1.2e-02$\pm$3.0e-04 & 1.2e-02$\pm$3.0e-04 & 1.1e-02$\pm$3.0e-04 & 1.3e-02$\pm$6.0e-04 & 1.2e-02$\pm$6.0e-04 & 1.2e-02$\pm$3.0e-04 & \textbf{1.1e-02}$\pm$3.0e-04 & 1.2e-02$\pm$2.0e-04 & 2.9e-02$\pm$7.9e-03\\
QM8(f1-PBE0) & 4152.1$\pm$159.3 & 1.8e-02$\pm$9.0e-04 & 1.5e-02$\pm$9.0e-04 & 1.6e-02$\pm$2.0e-03 & \textbf{1.4e-02}$\pm$1.7e-03 & 2.0e-02$\pm$1.0e-03 & 1.7e-02$\pm$1.0e-03 & 1.6e-02$\pm$2.3e-03 & 1.4e-02$\pm$2.0e-03 & 1.4e-02$\pm$1.6e-03 & 3.2e-02$\pm$6.7e-03\\
QM8(f2-PBE0) & 4059.8$\pm$240.7 & 3.9e-02$\pm$1.0e-03 & 3.4e-02$\pm$1.0e-03 & 3.4e-02$\pm$3.5e-03 & \textbf{3.0e-02}$\pm$2.9e-03 & 4.1e-02$\pm$1.6e-03 & 3.6e-02$\pm$1.7e-03 & 3.4e-02$\pm$3.5e-03 & 3.0e-02$\pm$2.8e-03 & 3.0e-02$\pm$2.3e-03 & 4.2e-02$\pm$5.9e-03\\
QM8(E1-PBE0.1) & 3932.1$\pm$133.8 & 1.2e-02$\pm$2.0e-04 & 1.1e-02$\pm$2.0e-04 & 1.1e-02$\pm$4.0e-04 & \textbf{9.6e-03}$\pm$3.0e-04 & 1.3e-02$\pm$5.0e-04 & 1.1e-02$\pm$4.0e-04 & 1.1e-02$\pm$4.0e-04 & 9.6e-03$\pm$2.0e-04 & 1.1e-02$\pm$3.0e-04 & 3.1e-02$\pm$4.8e-03\\
QM8(E2-PBE0.1) & 4063.2$\pm$133.4 & 1.3e-02$\pm$4.0e-04 & 1.2e-02$\pm$3.0e-04 & 1.2e-02$\pm$3.0e-04 & 1.1e-02$\pm$3.0e-04 & 1.3e-02$\pm$5.0e-04 & 1.2e-02$\pm$5.0e-04 & 1.2e-02$\pm$3.0e-04 & \textbf{1.1e-02}$\pm$3.0e-04 & 1.2e-02$\pm$3.0e-04 & 2.9e-02$\pm$3.3e-03\\
QM8(f1-PBE0.1) & 4152.1$\pm$159.3 & 1.8e-02$\pm$1.1e-03 & 1.5e-02$\pm$9.0e-04 & 1.6e-02$\pm$2.0e-03 & \textbf{1.4e-02}$\pm$1.7e-03 & 1.9e-02$\pm$1.1e-03 & 1.7e-02$\pm$1.0e-03 & 1.6e-02$\pm$2.2e-03 & 1.4e-02$\pm$2.0e-03 & 1.4e-02$\pm$1.6e-03 & 3.5e-02$\pm$8.0e-03\\
QM8(f2-PBE0.1) & 4059.8$\pm$240.7 & 3.9e-02$\pm$1.3e-03 & 3.4e-02$\pm$1.0e-03 & 3.3e-02$\pm$3.5e-03 & \textbf{3.0e-02}$\pm$2.9e-03 & 4.1e-02$\pm$1.9e-03 & 3.6e-02$\pm$1.8e-03 & 3.4e-02$\pm$3.4e-03 & 3.0e-02$\pm$3.0e-03 & 3.0e-02$\pm$2.5e-03 & 4.3e-02$\pm$7.6e-03\\
QM8(E1-CAM) & 3782.7$\pm$252.3 & 1.1e-02$\pm$4.0e-04 & 1.0e-02$\pm$3.0e-04 & 1.0e-02$\pm$3.0e-04 & \textbf{9.1e-03}$\pm$2.0e-04 & 1.2e-02$\pm$5.0e-04 & 1.1e-02$\pm$6.0e-04 & 1.0e-02$\pm$4.0e-04 & 9.3e-03$\pm$3.0e-04 & 1.0e-02$\pm$3.0e-04 & 2.9e-02$\pm$3.1e-03\\
QM8(E2-CAM) & 4004.4$\pm$151.5 & 1.2e-02$\pm$3.0e-04 & 1.1e-02$\pm$3.0e-04 & 1.1e-02$\pm$3.0e-04 & 1.0e-02$\pm$3.0e-04 & 1.3e-02$\pm$6.0e-04 & 1.2e-02$\pm$7.0e-04 & 1.1e-02$\pm$3.0e-04 & \textbf{1.0e-02}$\pm$3.0e-04 & 1.1e-02$\pm$3.0e-04 & 2.7e-02$\pm$3.2e-03\\
QM8(f1-CAM) & 4125.5$\pm$161.7 & 1.9e-02$\pm$1.2e-03 & 1.6e-02$\pm$1.1e-03 & 1.6e-02$\pm$2.0e-03 & \underline{\textbf{1.4e-02}}$\pm$1.7e-03 & 2.2e-02$\pm$1.0e-03 & 1.9e-02$\pm$1.0e-03 & 1.8e-02$\pm$1.4e-03 & 1.6e-02$\pm$1.2e-03 & 1.6e-02$\pm$9.0e-04 & 3.1e-02$\pm$4.1e-03\\
QM8(f2-CAM) & 4001.1$\pm$224.0 & 4.1e-02$\pm$1.8e-03 & 3.7e-02$\pm$1.3e-03 & 3.6e-02$\pm$3.0e-03 & \textbf{3.2e-02}$\pm$2.6e-03 & 4.4e-02$\pm$1.2e-03 & 4.0e-02$\pm$1.3e-03 & 3.8e-02$\pm$3.0e-03 & 3.3e-02$\pm$2.8e-03 & 3.3e-02$\pm$2.3e-03 & 5.1e-02$\pm$5.8e-03\\
QM9(mu) & 25845.8$\pm$660.3 & 8.9e-01$\pm$5.7e-02 & 8.1e-01$\pm$3.1e-02 & 8.4e-01$\pm$2.7e-02 & 7.6e-01$\pm$1.9e-02 & 8.7e-01$\pm$5.3e-02 & 7.8e-01$\pm$4.2e-02 & 7.9e-01$\pm$2.1e-02 & \underline{\textbf{7.1e-01}}$\pm$1.7e-02 & 7.7e-01$\pm$2.0e-02 & 2.1e+00$\pm$4.2e-01\\
QM9(alpha) & 24665.0$\pm$1191.1 & 1.2e+00$\pm$3.3e-02 & 1.1e+00$\pm$4.1e-02 & 1.6e+00$\pm$1.3e-01 & 1.4e+00$\pm$1.2e-01 & 1.2e+00$\pm$3.2e-02 & 1.1e+00$\pm$4.8e-02 & 9.9e-01$\pm$4.2e-02 & \underline{\textbf{8.8e-01}}$\pm$2.2e-02 & 1.0e+00$\pm$3.3e-02 & 1.4e+01$\pm$8.4e-01\\
QM9(homo) & 26024.3$\pm$221.8 & 7.5e-03$\pm$3.0e-04 & 7.1e-03$\pm$3.0e-04 & 7.4e-03$\pm$2.0e-04 & 7.0e-03$\pm$2.0e-04 & 7.0e-03$\pm$1.0e-04 & 6.6e-03$\pm$1.0e-04 & 6.6e-03$\pm$1.0e-04 & \underline{\textbf{6.2e-03}}$\pm$1.0e-04 & 6.6e-03$\pm$1.0e-04 & 2.2e+00$\pm$1.7e+00\\
QM9(lumo) & 25608.8$\pm$416.6 & 8.5e-03$\pm$2.0e-04 & 8.1e-03$\pm$2.0e-04 & 8.3e-03$\pm$2.0e-04 & 7.9e-03$\pm$3.0e-04 & 7.9e-03$\pm$2.0e-04 & 7.5e-03$\pm$2.0e-04 & 7.2e-03$\pm$1.0e-04 & \underline{\textbf{6.8e-03}}$\pm$1.0e-04 & 7.2e-03$\pm$1.0e-04 & 2.1e+00$\pm$1.5e+00\\
QM9(gap) & 25862.9$\pm$476.1 & 1.1e-02$\pm$2.0e-04 & 1.0e-02$\pm$2.0e-04 & 1.1e-02$\pm$4.0e-04 & 9.8e-03$\pm$3.0e-04 & 1.0e-02$\pm$5.0e-04 & 9.6e-03$\pm$6.0e-04 & 9.5e-03$\pm$2.0e-04 & \underline{\textbf{8.8e-03}}$\pm$3.0e-04 & 9.4e-03$\pm$2.0e-04 & 1.1e+00$\pm$5.9e-01\\
QM9(r2) & 25736.3$\pm$537.4 & 6.3e+01$\pm$2.0e+00 & 5.7e+01$\pm$1.6e+00 & 6.9e+01$\pm$2.5e+00 & 6.2e+01$\pm$2.7e+00 & 5.8e+01$\pm$1.9e+00 & 5.2e+01$\pm$1.9e+00 & 5.4e+01$\pm$7.8e-01 & \underline{\textbf{4.8e+01}}$\pm$6.1e-01 & 5.4e+01$\pm$9.5e-01 & 2.4e+02$\pm$1.1e+01\\
QM9(zpve) & 24486.6$\pm$870.8 & 1.6e-03$\pm$2.0e-04 & \textbf{1.5e-03}$\pm$1.0e-04 & 3.8e-03$\pm$4.0e-04 & 3.6e-03$\pm$4.0e-04 & 2.8e-03$\pm$1.0e-04 & 2.7e-03$\pm$1.0e-04 & 2.4e-03$\pm$1.0e-04 & 2.2e-03$\pm$1.0e-04 & 2.4e-03$\pm$1.0e-04 & 1.9e+00$\pm$1.6e+00\\
QM9(u0) & 25375.4$\pm$499.9 & 1.9e+00$\pm$2.4e-01 & \underline{\textbf{1.7e+00}}$\pm$2.1e-01 & 4.9e+00$\pm$6.5e-01 & 4.2e+00$\pm$5.2e-01 & 3.7e+00$\pm$1.4e-01 & 3.2e+00$\pm$1.4e-01 & 3.0e+00$\pm$1.3e-01 & 2.5e+00$\pm$9.2e-02 & 2.9e+00$\pm$1.3e-01 & 4.4e+01$\pm$3.7e+00\\
QM9(u298) & 25204.3$\pm$754.3 & 1.9e+00$\pm$2.5e-01 & \textbf{1.7e+00}$\pm$2.0e-01 & 4.9e+00$\pm$6.9e-01 & 4.2e+00$\pm$5.3e-01 & 3.7e+00$\pm$1.3e-01 & 3.2e+00$\pm$1.5e-01 & 3.0e+00$\pm$1.5e-01 & 2.5e+00$\pm$8.5e-02 & 2.9e+00$\pm$1.3e-01 & 4.5e+01$\pm$3.7e+00\\
QM9(h298) & 25288.9$\pm$581.3 & 1.9e+00$\pm$2.4e-01 & \underline{\textbf{1.7e+00}}$\pm$2.0e-01 & 4.9e+00$\pm$6.8e-01 & 4.2e+00$\pm$5.3e-01 & 3.7e+00$\pm$1.3e-01 & 3.2e+00$\pm$1.4e-01 & 3.0e+00$\pm$1.3e-01 & 2.5e+00$\pm$8.1e-02 & 2.9e+00$\pm$1.1e-01 & 4.4e+01$\pm$3.8e+00\\
QM9(g298) & 25345.2$\pm$543.2 & 1.9e+00$\pm$2.5e-01 & \textbf{1.7e+00}$\pm$2.1e-01 & 4.9e+00$\pm$6.5e-01 & 4.2e+00$\pm$5.2e-01 & 3.7e+00$\pm$1.3e-01 & 3.2e+00$\pm$1.4e-01 & 3.1e+00$\pm$1.3e-01 & 2.5e+00$\pm$8.8e-02 & 2.9e+00$\pm$1.3e-01 & 4.4e+01$\pm$3.7e+00\\
QM9(cv) & 25051.2$\pm$829.4 & 6.0e-01$\pm$2.1e-02 & 5.5e-01$\pm$2.0e-02 & 6.9e-01$\pm$5.5e-02 & 6.3e-01$\pm$4.7e-02 & 5.7e-01$\pm$1.8e-02 & 5.3e-01$\pm$2.0e-02 & 4.9e-01$\pm$1.5e-02 & \underline{\textbf{4.4e-01}}$\pm$8.6e-03 & 4.9e-01$\pm$1.5e-02 & 5.2e+00$\pm$1.3e+00\\

\bottomrule

\toprule
Width @ 90\% &\\
\midrule
QM8(E1-CC2) & 2373.2$\pm$351.0 & 2.8e-02$\pm$2.1e-03 & 2.7e-02$\pm$8.0e-04 & 2.6e-02$\pm$1.4e-03 & 2.5e-02$\pm$9.0e-04 & 2.9e-02$\pm$2.6e-03 & 2.7e-02$\pm$8.0e-04 & 2.5e-02$\pm$1.2e-03 & \textbf{2.4e-02}$\pm$7.0e-04 & 2.5e-02$\pm$1.4e-03 & 8.4e-02$\pm$1.4e-02\\
QM8(E2-CC2) & 2356.4$\pm$368.3 & 3.3e-02$\pm$1.2e-03 & 3.3e-02$\pm$8.0e-04 & 3.0e-02$\pm$1.2e-03 & 2.9e-02$\pm$1.0e-03 & 3.3e-02$\pm$1.2e-03 & 3.3e-02$\pm$1.0e-03 & 2.9e-02$\pm$2.0e-04 & 2.9e-02$\pm$5.0e-04 & \textbf{2.8e-02}$\pm$6.0e-04 & 8.8e-02$\pm$1.5e-02\\
QM8(f1-CC2) & 3239.5$\pm$414.0 & 4.4e-02$\pm$7.7e-03 & 3.2e-02$\pm$5.9e-03 & 3.6e-02$\pm$7.0e-03 & \textbf{2.9e-02}$\pm$5.0e-03 & 5.6e-02$\pm$8.8e-03 & 3.6e-02$\pm$6.1e-03 & 4.0e-02$\pm$8.0e-03 & 3.2e-02$\pm$5.4e-03 & 4.0e-02$\pm$6.9e-03 & 8.8e-02$\pm$9.8e-03\\
QM8(f2-CC2) & 3061.2$\pm$504.8 & 1.1e-01$\pm$1.5e-02 & 8.6e-02$\pm$1.3e-02 & 9.3e-02$\pm$1.8e-02 & \textbf{7.7e-02}$\pm$1.5e-02 & 1.1e-01$\pm$1.9e-02 & 8.8e-02$\pm$1.3e-02 & 9.2e-02$\pm$1.7e-02 & 8.1e-02$\pm$1.4e-02 & 8.5e-02$\pm$1.2e-02 & 1.5e-01$\pm$3.2e-02\\
QM8(E1-PBE0) & 2557.4$\pm$218.2 & 2.9e-02$\pm$2.1e-03 & 2.8e-02$\pm$8.0e-04 & 2.6e-02$\pm$1.5e-03 & 2.6e-02$\pm$8.0e-04 & 2.9e-02$\pm$2.6e-03 & 2.8e-02$\pm$9.0e-04 & \textbf{2.5e-02}$\pm$1.2e-03 & 2.6e-02$\pm$8.0e-04 & 2.6e-02$\pm$1.1e-03 & 9.1e-02$\pm$5.9e-03\\
QM8(E2-PBE0) & 2956.5$\pm$408.3 & 3.2e-02$\pm$1.0e-03 & 3.2e-02$\pm$7.0e-04 & 2.9e-02$\pm$1.1e-03 & 2.9e-02$\pm$1.0e-03 & 3.2e-02$\pm$9.0e-04 & 3.2e-02$\pm$6.0e-04 & 2.8e-02$\pm$9.0e-04 & 2.9e-02$\pm$7.0e-04 & \textbf{2.8e-02}$\pm$4.0e-04 & 7.5e-02$\pm$7.2e-03\\
QM8(f1-PBE0) & 3493.5$\pm$488.2 & 3.6e-02$\pm$6.8e-03 & 2.8e-02$\pm$6.9e-03 & 3.3e-02$\pm$7.6e-03 & \textbf{2.6e-02}$\pm$7.3e-03 & 4.0e-02$\pm$6.0e-03 & 3.1e-02$\pm$6.9e-03 & 3.4e-02$\pm$8.7e-03 & 3.0e-02$\pm$7.2e-03 & 3.2e-02$\pm$5.3e-03 & 9.1e-02$\pm$1.3e-02\\
QM8(f2-PBE0) & 3155.5$\pm$646.3 & 8.4e-02$\pm$1.8e-02 & 6.8e-02$\pm$1.2e-02 & 7.1e-02$\pm$1.9e-02 & \textbf{5.9e-02}$\pm$1.4e-02 & 8.3e-02$\pm$1.6e-02 & 6.9e-02$\pm$1.2e-02 & 6.9e-02$\pm$1.8e-02 & 6.2e-02$\pm$1.4e-02 & 6.8e-02$\pm$1.3e-02 & 1.3e-01$\pm$2.6e-02\\
QM8(E1-PBE0.1) & 2557.4$\pm$218.2 & 3.0e-02$\pm$5.1e-03 & 2.8e-02$\pm$8.0e-04 & 2.6e-02$\pm$1.5e-03 & 2.6e-02$\pm$8.0e-04 & 2.9e-02$\pm$2.7e-03 & 2.8e-02$\pm$9.0e-04 & \textbf{2.5e-02}$\pm$1.2e-03 & 2.6e-02$\pm$7.0e-04 & 2.6e-02$\pm$1.1e-03 & 8.1e-02$\pm$1.2e-02\\
QM8(E2-PBE0.1) & 2956.5$\pm$408.3 & 3.2e-02$\pm$9.0e-04 & 3.2e-02$\pm$7.0e-04 & 2.9e-02$\pm$1.1e-03 & 2.9e-02$\pm$1.0e-03 & 3.2e-02$\pm$9.0e-04 & 3.2e-02$\pm$6.0e-04 & 2.9e-02$\pm$9.0e-04 & 2.9e-02$\pm$7.0e-04 & \textbf{2.8e-02}$\pm$5.0e-04 & 8.4e-02$\pm$7.6e-03\\
QM8(f1-PBE0.1) & 3493.5$\pm$488.2 & 3.5e-02$\pm$8.4e-03 & 2.8e-02$\pm$6.9e-03 & 3.3e-02$\pm$7.9e-03 & \textbf{2.6e-02}$\pm$7.3e-03 & 3.8e-02$\pm$7.5e-03 & 3.1e-02$\pm$6.9e-03 & 3.4e-02$\pm$8.3e-03 & 2.9e-02$\pm$7.2e-03 & 3.2e-02$\pm$5.6e-03 & 8.9e-02$\pm$1.7e-02\\
QM8(f2-PBE0.1) & 3155.5$\pm$646.3 & 8.4e-02$\pm$2.1e-02 & 6.8e-02$\pm$1.2e-02 & 7.2e-02$\pm$2.0e-02 & \textbf{5.9e-02}$\pm$1.4e-02 & 8.3e-02$\pm$1.7e-02 & 6.9e-02$\pm$1.1e-02 & 6.9e-02$\pm$1.8e-02 & 6.1e-02$\pm$1.4e-02 & 6.8e-02$\pm$1.3e-02 & 1.2e-01$\pm$2.0e-02\\
QM8(E1-CAM) & 2317.5$\pm$472.6 & 2.8e-02$\pm$2.3e-03 & 2.6e-02$\pm$9.0e-04 & 2.5e-02$\pm$1.1e-03 & 2.4e-02$\pm$7.0e-04 & 2.8e-02$\pm$2.0e-03 & 2.7e-02$\pm$9.0e-04 & 2.4e-02$\pm$1.2e-03 & 2.4e-02$\pm$8.0e-04 & \textbf{2.4e-02}$\pm$1.1e-03 & 9.0e-02$\pm$2.5e-02\\
QM8(E2-CAM) & 2791.8$\pm$491.5 & 2.9e-02$\pm$1.0e-03 & 3.0e-02$\pm$7.0e-04 & 2.7e-02$\pm$1.2e-03 & 2.7e-02$\pm$9.0e-04 & 2.9e-02$\pm$1.0e-03 & 3.0e-02$\pm$8.0e-04 & 2.7e-02$\pm$1.2e-03 & 2.7e-02$\pm$8.0e-04 & \textbf{2.6e-02}$\pm$6.0e-04 & 7.6e-02$\pm$1.0e-02\\
QM8(f1-CAM) & 3366.6$\pm$448.1 & 3.4e-02$\pm$9.3e-03 & 2.8e-02$\pm$7.5e-03 & 3.0e-02$\pm$8.9e-03 & \textbf{2.5e-02}$\pm$7.2e-03 & 4.3e-02$\pm$9.2e-03 & 3.2e-02$\pm$7.0e-03 & 3.4e-02$\pm$1.0e-02 & 2.9e-02$\pm$6.9e-03 & 3.4e-02$\pm$8.1e-03 & 1.0e-01$\pm$2.5e-02\\
QM8(f2-CAM) & 3059.9$\pm$482.0 & 8.8e-02$\pm$1.6e-02 & 7.0e-02$\pm$1.3e-02 & 7.4e-02$\pm$1.6e-02 & \textbf{6.2e-02}$\pm$1.4e-02 & 9.0e-02$\pm$1.4e-02 & 7.1e-02$\pm$1.3e-02 & 7.3e-02$\pm$1.7e-02 & 6.4e-02$\pm$1.3e-02 & 6.9e-02$\pm$1.3e-02 & 1.4e-01$\pm$1.3e-02\\
QM9(mu) & 23224.8$\pm$2625.8 & 2.2e+00$\pm$6.9e-02 & 2.2e+00$\pm$5.7e-02 & 2.1e+00$\pm$5.7e-02 & 2.1e+00$\pm$4.1e-02 & 2.1e+00$\pm$1.1e-01 & 2.2e+00$\pm$7.5e-02 & \textbf{1.9e+00}$\pm$4.8e-02 & 2.1e+00$\pm$3.7e-02 & 2.0e+00$\pm$3.7e-02 & 7.1e+00$\pm$2.8e+00\\
QM9(alpha) & 19039.0$\pm$3410.6 & 2.7e+00$\pm$1.1e-01 & 2.6e+00$\pm$7.5e-02 & 3.5e+00$\pm$3.6e-01 & 3.4e+00$\pm$2.8e-01 & 2.6e+00$\pm$9.3e-02 & 2.5e+00$\pm$7.7e-02 & 2.3e+00$\pm$1.2e-01 & \textbf{2.3e+00}$\pm$7.0e-02 & 2.4e+00$\pm$9.0e-02 & 4.0e+01$\pm$2.7e+00\\
QM9(homo) & 23865.0$\pm$834.2 & 1.8e-02$\pm$6.0e-04 & 1.8e-02$\pm$4.0e-04 & 1.8e-02$\pm$4.0e-04 & 1.8e-02$\pm$3.0e-04 & 1.7e-02$\pm$3.0e-04 & 1.8e-02$\pm$3.0e-04 & \textbf{1.7e-02}$\pm$3.0e-04 & 1.7e-02$\pm$1.0e-04 & 1.7e-02$\pm$2.0e-04 & 6.7e+00$\pm$3.8e+00\\
QM9(lumo) & 21867.7$\pm$1906.7 & 2.1e-02$\pm$5.0e-04 & 2.1e-02$\pm$5.0e-04 & 2.0e-02$\pm$5.0e-04 & 2.0e-02$\pm$5.0e-04 & 1.9e-02$\pm$4.0e-04 & 1.9e-02$\pm$5.0e-04 & \textbf{1.8e-02}$\pm$3.0e-04 & 1.8e-02$\pm$2.0e-04 & 1.8e-02$\pm$2.0e-04 & 1.3e+01$\pm$7.1e+00\\
QM9(gap) & 22868.0$\pm$2105.0 & 2.7e-02$\pm$7.0e-04 & 2.7e-02$\pm$8.0e-04 & 2.6e-02$\pm$7.0e-04 & 2.5e-02$\pm$6.0e-04 & 2.5e-02$\pm$1.0e-03 & 2.5e-02$\pm$1.1e-03 & \textbf{2.3e-02}$\pm$4.0e-04 & 2.3e-02$\pm$5.0e-04 & 2.4e-02$\pm$4.0e-04 & 1.1e+01$\pm$6.1e+00\\
QM9(r2) & 22850.8$\pm$1979.8 & 1.5e+02$\pm$4.8e+00 & 1.6e+02$\pm$3.4e+00 & 1.6e+02$\pm$8.0e+00 & 1.6e+02$\pm$6.6e+00 & 1.4e+02$\pm$2.6e+00 & 1.5e+02$\pm$3.0e+00 & \textbf{1.3e+02}$\pm$2.7e+00 & 1.4e+02$\pm$2.4e+00 & 1.3e+02$\pm$2.5e+00 & 7.6e+02$\pm$2.9e+01\\
QM9(zpve) & 17884.2$\pm$2703.3 & 3.6e-03$\pm$5.0e-04 & \textbf{3.5e-03}$\pm$4.0e-04 & 8.8e-03$\pm$9.0e-04 & 8.7e-03$\pm$8.0e-04 & 6.0e-03$\pm$2.0e-04 & 6.0e-03$\pm$2.0e-04 & 5.6e-03$\pm$2.0e-04 & 5.6e-03$\pm$2.0e-04 & 5.8e-03$\pm$2.0e-04 & 1.1e+01$\pm$7.8e+00\\
QM9(u0) & 21369.4$\pm$1980.2 & 5.3e+00$\pm$9.9e-01 & \textbf{4.6e+00}$\pm$7.7e-01 & 1.2e+01$\pm$1.4e+00 & 1.0e+01$\pm$1.3e+00 & 7.8e+00$\pm$5.4e-01 & 7.0e+00$\pm$3.1e-01 & 6.8e+00$\pm$5.5e-01 & 6.4e+00$\pm$4.6e-01 & 7.6e+00$\pm$3.0e-01 & 1.7e+02$\pm$1.5e+01\\
QM9(u298) & 20728.2$\pm$2926.0 & 5.3e+00$\pm$1.0e+00 & \textbf{4.6e+00}$\pm$7.8e-01 & 1.2e+01$\pm$1.6e+00 & 1.0e+01$\pm$1.4e+00 & 7.8e+00$\pm$5.5e-01 & 7.1e+00$\pm$3.0e-01 & 6.8e+00$\pm$6.1e-01 & 6.4e+00$\pm$5.1e-01 & 7.5e+00$\pm$3.1e-01 & 1.7e+02$\pm$1.5e+01\\
QM9(h298) & 21053.4$\pm$2274.2 & 5.2e+00$\pm$1.0e+00 & \textbf{4.6e+00}$\pm$8.1e-01 & 1.2e+01$\pm$1.5e+00 & 1.0e+01$\pm$1.4e+00 & 7.8e+00$\pm$5.6e-01 & 7.0e+00$\pm$3.4e-01 & 6.8e+00$\pm$5.6e-01 & 6.4e+00$\pm$4.8e-01 & 7.5e+00$\pm$2.9e-01 & 1.7e+02$\pm$1.3e+01\\
QM9(g298) & 21269.7$\pm$2092.8 & 5.3e+00$\pm$9.8e-01 & \textbf{4.6e+00}$\pm$7.7e-01 & 1.2e+01$\pm$1.4e+00 & 1.0e+01$\pm$1.3e+00 & 7.8e+00$\pm$5.4e-01 & 7.0e+00$\pm$3.1e-01 & 6.8e+00$\pm$5.5e-01 & 6.4e+00$\pm$4.6e-01 & 7.6e+00$\pm$3.0e-01 & 1.7e+02$\pm$1.5e+01\\
QM9(cv) & 20084.8$\pm$2942.8 & 1.4e+00$\pm$6.2e-02 & 1.4e+00$\pm$6.0e-02 & 1.6e+00$\pm$1.5e-01 & 1.5e+00$\pm$1.1e-01 & 1.3e+00$\pm$2.5e-02 & 1.3e+00$\pm$3.4e-02 & 1.2e+00$\pm$3.7e-02 & \textbf{1.2e+00}$\pm$2.8e-02 & 1.2e+00$\pm$3.3e-02 & 2.1e+01$\pm$5.5e+00\\

\bottomrule
\end{tabular}
\label{appendix:table:exp:real:qm_width}
\end{tiny}
\end{center}
\end{sidewaystable}
}

\def \TableFunctionalGroups{
\begin{table}[ht]
\captionof{table}{
(On QM9) MCR conditioning on the presence of certain functional groups in the original SMILES representation.
The list of groups are taken from the OPENSMILES project,  as implemented in~\cite{GlobalChem}.
The (pooled) mean coverage rate are computed over 10 random re-splits of the full QM9 dataset like in other parts of this paper.
We keep only the functional groups with at least 200 appearances in all ten randomly sampled test set.
All numbers not significantly lower than 90\% at $p=0.05$ are in bold.
}
\begin{center}
\begin{tiny}
\setlength\tabcolsep{2pt}
\begin{tabular}{lcccccccccccr}

\toprule
Conditional Coverage Rate & alpha & cv & g298 & gap & h298 & homo & lumo & mu & r2 & u0 & u298 & zpve\\
\midrule
2-butyne & 88.4$\pm$0.4 & \textbf{94.6}$\pm$0.4 & \textbf{93.1}$\pm$0.4 & \textbf{94.9}$\pm$0.4 & \textbf{92.8}$\pm$0.4 & \textbf{95.6}$\pm$0.4 & \textbf{94.4}$\pm$0.4 & \textbf{94.3}$\pm$0.4 & \textbf{93.2}$\pm$0.4 & \textbf{92.8}$\pm$0.4 & \textbf{93.0}$\pm$0.4 & \textbf{92.2}$\pm$0.4\\
aldehyde & \textbf{93.3}$\pm$0.4 & \textbf{94.4}$\pm$0.4 & \textbf{94.8}$\pm$0.4 & \textbf{92.6}$\pm$0.4 & \textbf{94.4}$\pm$0.4 & \textbf{94.7}$\pm$0.4 & \textbf{91.1}$\pm$0.4 & \textbf{90.7}$\pm$0.4 & \textbf{90.4}$\pm$0.4 & \textbf{94.4}$\pm$0.4 & \textbf{94.6}$\pm$0.4 & \textbf{93.6}$\pm$0.4\\
amide & \textbf{89.7}$\pm$0.3 & \textbf{89.7}$\pm$0.3 & \textbf{90.0}$\pm$0.3 & \textbf{91.1}$\pm$0.3 & \textbf{90.6}$\pm$0.3 & \textbf{92.2}$\pm$0.3 & \textbf{89.9}$\pm$0.3 & 88.0$\pm$0.3 & \textbf{91.4}$\pm$0.3 & \textbf{90.3}$\pm$0.3 & \textbf{90.2}$\pm$0.3 & \textbf{90.3}$\pm$0.3\\
carboxylic acid & \textbf{91.9}$\pm$0.4 & \textbf{93.2}$\pm$0.4 & \textbf{90.0}$\pm$0.4 & \textbf{91.4}$\pm$0.4 & \textbf{89.7}$\pm$0.4 & \textbf{92.3}$\pm$0.4 & \textbf{92.4}$\pm$0.4 & \textbf{92.7}$\pm$0.4 & \textbf{91.8}$\pm$0.4 & \textbf{90.0}$\pm$0.4 & \textbf{90.0}$\pm$0.4 & \textbf{91.1}$\pm$0.4\\
cyclopropane & \textbf{89.9}$\pm$0.3 & \textbf{90.5}$\pm$0.3 & \textbf{93.5}$\pm$0.3 & \textbf{90.9}$\pm$0.3 & \textbf{93.7}$\pm$0.3 & \textbf{89.8}$\pm$0.3 & \textbf{91.8}$\pm$0.3 & \textbf{93.1}$\pm$0.3 & 88.7$\pm$0.3 & \textbf{93.6}$\pm$0.3 & \textbf{93.8}$\pm$0.3 & \textbf{91.2}$\pm$0.3\\
dimethyl ether & \textbf{90.4}$\pm$0.1 & \textbf{91.9}$\pm$0.1 & 89.4$\pm$0.1 & \textbf{91.6}$\pm$0.1 & 89.3$\pm$0.1 & \textbf{91.9}$\pm$0.1 & \textbf{91.5}$\pm$0.1 & \textbf{92.2}$\pm$0.1 & 88.7$\pm$0.1 & 89.3$\pm$0.1 & 89.4$\pm$0.1 & \textbf{90.6}$\pm$0.1\\
ester & \textbf{92.5}$\pm$0.6 & \textbf{94.0}$\pm$0.6 & \textbf{90.8}$\pm$0.6 & \textbf{90.2}$\pm$0.6 & \textbf{90.2}$\pm$0.6 & \textbf{91.5}$\pm$0.6 & \textbf{91.4}$\pm$0.6 & \textbf{91.9}$\pm$0.6 & \textbf{92.3}$\pm$0.6 & \textbf{90.3}$\pm$0.6 & \textbf{90.6}$\pm$0.6 & \textbf{92.3}$\pm$0.6\\
ethanol & \textbf{90.0}$\pm$0.2 & \textbf{92.0}$\pm$0.2 & \textbf{90.0}$\pm$0.2 & \textbf{93.1}$\pm$0.2 & \textbf{90.1}$\pm$0.2 & \textbf{92.9}$\pm$0.2 & \textbf{92.5}$\pm$0.2 & \textbf{91.8}$\pm$0.2 & 88.1$\pm$0.2 & \textbf{90.0}$\pm$0.2 & \textbf{90.1}$\pm$0.2 & \textbf{90.8}$\pm$0.2\\
ethene & 84.1$\pm$0.2 & 89.5$\pm$0.2 & 86.7$\pm$0.2 & \textbf{90.3}$\pm$0.2 & 86.9$\pm$0.2 & \textbf{91.4}$\pm$0.2 & 89.3$\pm$0.2 & \textbf{92.8}$\pm$0.2 & \textbf{90.0}$\pm$0.2 & 86.7$\pm$0.2 & 86.8$\pm$0.2 & 87.0$\pm$0.2\\
ether & \textbf{90.4}$\pm$0.1 & \textbf{91.9}$\pm$0.1 & 89.4$\pm$0.1 & \textbf{91.6}$\pm$0.1 & 89.3$\pm$0.1 & \textbf{91.9}$\pm$0.1 & \textbf{91.5}$\pm$0.1 & \textbf{92.2}$\pm$0.1 & 88.7$\pm$0.1 & 89.3$\pm$0.1 & 89.4$\pm$0.1 & \textbf{90.6}$\pm$0.1\\
formaldehyde & \textbf{90.3}$\pm$0.2 & \textbf{91.5}$\pm$0.2 & \textbf{91.9}$\pm$0.2 & \textbf{90.9}$\pm$0.2 & \textbf{92.0}$\pm$0.2 & \textbf{91.7}$\pm$0.2 & \textbf{89.7}$\pm$0.2 & 86.7$\pm$0.2 & 88.3$\pm$0.2 & \textbf{91.9}$\pm$0.2 & \textbf{91.9}$\pm$0.2 & \textbf{91.5}$\pm$0.2\\
hydrogen cyanide & \textbf{93.8}$\pm$0.2 & \textbf{93.7}$\pm$0.2 & \textbf{93.4}$\pm$0.2 & \textbf{89.9}$\pm$0.2 & \textbf{93.4}$\pm$0.2 & 89.6$\pm$0.2 & \textbf{92.1}$\pm$0.2 & 88.3$\pm$0.2 & \textbf{92.2}$\pm$0.2 & \textbf{93.3}$\pm$0.2 & \textbf{93.1}$\pm$0.2 & \textbf{91.6}$\pm$0.2\\
ketone & 88.7$\pm$0.3 & \textbf{90.9}$\pm$0.3 & 89.4$\pm$0.3 & \textbf{92.2}$\pm$0.3 & 89.5$\pm$0.3 & \textbf{91.5}$\pm$0.3 & \textbf{90.7}$\pm$0.3 & \textbf{90.6}$\pm$0.3 & \textbf{90.7}$\pm$0.3 & \textbf{89.7}$\pm$0.3 & \textbf{89.8}$\pm$0.3 & \textbf{89.8}$\pm$0.3\\
prop-1-ene & 84.5$\pm$0.3 & \textbf{89.6}$\pm$0.3 & 85.2$\pm$0.3 & \textbf{90.4}$\pm$0.3 & 85.6$\pm$0.3 & \textbf{92.5}$\pm$0.3 & 89.3$\pm$0.3 & \textbf{94.1}$\pm$0.3 & 89.3$\pm$0.3 & 85.3$\pm$0.3 & 85.5$\pm$0.3 & 86.1$\pm$0.3\\
prop-1-yne & \textbf{89.8}$\pm$0.3 & \textbf{94.8}$\pm$0.3 & \textbf{94.0}$\pm$0.3 & \textbf{94.6}$\pm$0.3 & \textbf{93.6}$\pm$0.3 & \textbf{95.4}$\pm$0.3 & \textbf{94.0}$\pm$0.3 & \textbf{94.4}$\pm$0.3 & \textbf{92.2}$\pm$0.3 & \textbf{93.8}$\pm$0.3 & \textbf{93.8}$\pm$0.3 & \textbf{92.8}$\pm$0.3\\

\bottomrule
\end{tabular}
\label{appendix:table:exp:real:qm9_functional_groups}
\end{tiny}
\end{center}
\end{table}
}

\def \FigDJ{
\begin{figure}[ht]
    \centering
    \includegraphics[width=1\textwidth]{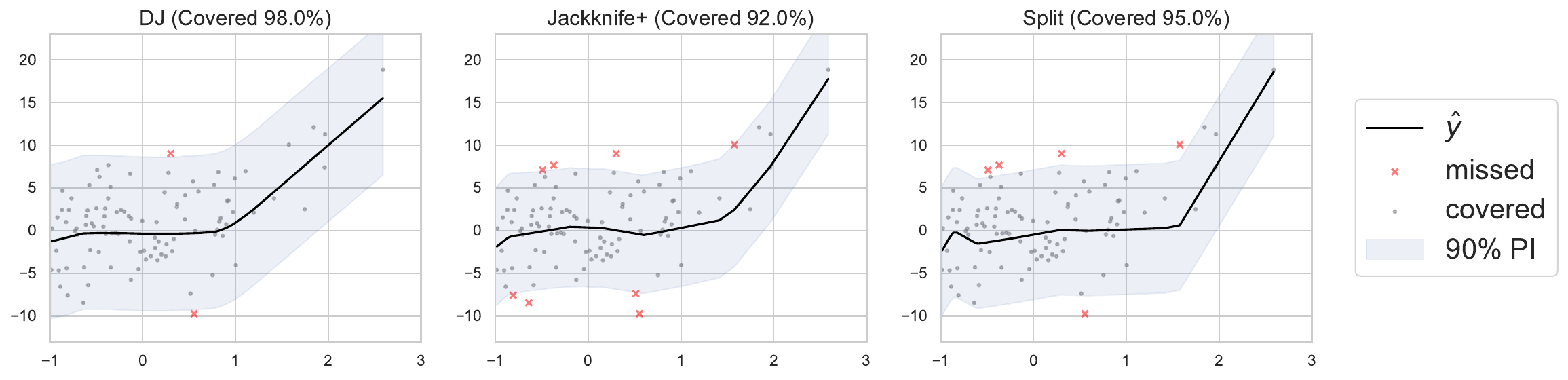}
\caption[ Synthetic Data]
{
DJ, Jackknife+ and Split conformal on the synthetic dataset.
\label{appendix:fig:exp:dj}
}
\end{figure}
}

\title{Locally Valid and Discriminative Prediction Intervals for Deep Learning Models}

\author{%
  Zhen Lin \\%\thanks{Contact: zhenlin4@illinois.edu} \\
  University of Illinois at Urbana-Champaign\\
  Urbana, IL 61801 \\
  \texttt{zhenlin4@illinois.edu} \\
  \And
  Shubhendu Trivedi \\
  MIT\\
  Cambridge, MA 02139 \\
  \texttt{shubhendu@csail.mit.edu} \\
  \And 
  Jimeng Sun \\
  University of Illinois at Urbana-Champaign\\
  Urbana, IL 61801 \\
  \texttt{jimeng@illinois.edu} \\
}

\begin{document}

\maketitle

\begin{abstract}
    Crucial for building trust in deep learning models for critical real-world applications is efficient and theoretically sound uncertainty quantification, a task that continues to be challenging. Useful uncertainty information is expected to have two key properties: It should be \textit{valid} (guaranteeing coverage) and \textit{discriminative} (more uncertain when the expected risk is high). 
    Moreover, when combined with deep learning (DL) methods, it should be \textit{scalable} and \textit{affect the DL model performance minimally}. 
    Most existing Bayesian methods lack frequentist coverage guarantees and usually affect model performance. 
    The few available frequentist methods are rarely discriminative and/or violate coverage guarantees due to unrealistic assumptions. 
    Moreover, many methods are expensive or require substantial modifications to the base neural network. 
    Building upon recent advances in conformal prediction \cite{guan2020conformal, covshift_tibshirani2020conformal} and leveraging the classical idea of kernel regression, we propose Locally Valid and Discriminative prediction intervals (\methodname), a simple, efficient and lightweight method to construct discriminative prediction intervals (PIs) for almost \textit{any} DL model.
    With no assumptions on the data distribution, such PIs also offer finite-sample local coverage guarantees (contrasted to the simpler marginal coverage).
    We empirically verify, using diverse datasets, that besides being the only locally valid method for DL, \methodname also exceeds or matches the performance (including coverage rate and prediction accuracy) of existing uncertainty quantification methods, while offering additional benefits in scalability and flexibility.
\vspace{-5pt}    
\end{abstract}

\section{Introduction}\label{sec:intro}
\vspace{-5pt}    
Consider a training set $\fullset = \{(X_i, Y_i)\}_{i=1}^{N}$ and a test example $(X_{N+1}, Y_{N+1})$, all drawn i.i.d from an arbitrary joint distribution $\mathcal{P}$, with $(X_i, Y_i) \in \mathcal{X} \times \mathcal{Y}$ for some $\mathcal{X}\subseteq \mathbb{R}^d$ and  $\mathcal{Y}\subseteq \mathbb{R}$. We are interested in the problem of predictive inference: On observing $\fullset$ and $X_{N+1}$, our task is to construct a prediction interval (PI)~\footnote{\noindent Several recent deep learning papers use ``Confidence Interval'' and ``Prediction Interval'' interchangeably. We stick to the conventional statistical usage.} estimate $\hat{C}(X_{N+1})$ that contains the true value of $Y_{N+1}$ with a (pre-specified) high probability. 

\begin{figure}[ht]
\vspace{-10pt}
    \centering
    \includegraphics[width=1\textwidth]{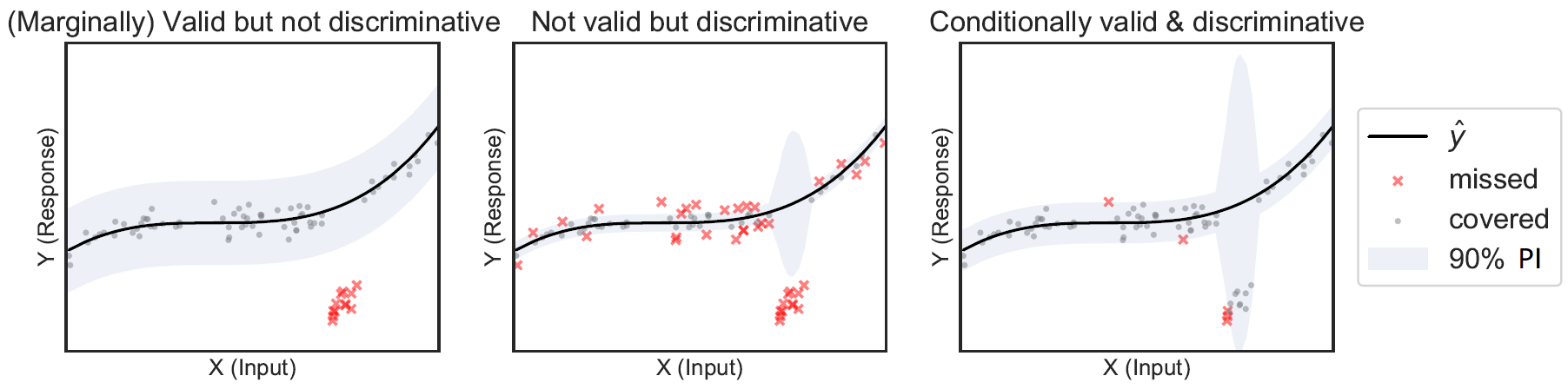}
    %\vskip\baselineskip
\caption[Example PIs]
{
Illustration of possible (good and bad) PIs. 
The PI on the left is valid, as it covers 90\% of the data. 
It is however only marginally valid, not reflecting the poor model prediction near the red cluster.
The middle PI is discriminative, and reflects the high error near the red cluster, but its coverage rate is much lower than the target (thus not valid). 
The PI on the right addresses both challenges by stretching ``just enough'' near the cluster, making it not only discriminative, but also \textit{conditionally} valid. We seek to construct PIs of the last type, but constructing exact conditionally valid PIs in a distribution-free setting is theoretically impossible. We thus relax this goal by instead aiming for local validity (more details in Section \ref{sec:prelim}).
\label{fig:intro:example}
}
\vspace{-15pt}
\end{figure}

The construction of actionable PIs involves two {\it general challenges}: 
First, $\hat{C}$ should be \textbf{valid}, meaning that if the specified probability is $1-\alpha$, we expect $\hat{C}(X_{N+1})$ to cover $Y_{N+1}$ at least $1-\alpha$ of the time.
Moreover, $\hat{C}$ should be \textbf{discriminative} i.e., we expect $\hat{C}(X_{N+1})$ to be narrower for confident cases and vice-versa. 
The width of the PI $\hat{C}(X_{N+1})$ is thus quantification of the uncertainty. Figure \ref{fig:intro:example} illustrates these notions, with more details in Section \ref{subsec:prelim:coverage} and \ref{subsec:prelim:discrimination}.

While deep learning (DL) models have demonstrated impressive performance over a range of complicated tasks and data modalities, it has remained difficult to quantify the uncertainty for their predictions. For DL predictions to be actionable, uncertainty information is however indispensable, especially in domains like medicine and finance \cite{amodei2016concrete}.
Apart from requiring validity and discrimination as discussed earlier, two additional challenges exist {\it specifically} for DL models.
Obviously, any uncertainty estimation method needs to finish reasonably fast to be useful, so the third challenge is {\it scalability}.
The fourth challenge is {\it accuracy}: The uncertainty estimation should not decrease the prediction accuracy of the DL model. 
\textit{Post-hoc} methods are ideal because they usually do not interfere with the base NN prediction at all.
These four requirements together constitute a set of essential desiderata for uncertainty quantification in DL.

Existing uncertainty estimation methods for DL rarely address more than one or two of the above requirements.
Credible intervals given by posteriors of approximate Bayesian methods such as \cite{Welling2011BayesianDynamics,DBLP:conf/icml/Hernandez-Lobato15b}, deep ensemble \cite{Lakshminarayanan2017SimpleEnsembles, AGDeepEnsemble} and Monte-Carlo dropout \cite{Gal2016DropoutLearning} are not valid in the frequentist sense \cite{BayesianFrequentist_10.1214/088342304000000116}.
Most existing methods also interfere with the original model design, loss function and/or training, which could be expensive and decrease the model performance (as verified in our experiments)~\cite{Gal2016DropoutLearning, pmlr-v119-alaa20a,BayesianFrequentist_10.1214/088342304000000116,Lakshminarayanan2017SimpleEnsembles}. 

%===============================================================
To address these requirements, we leverage recent advances in conformal prediction and the classical idea of kernel regression. 
Conformal prediction, pioneered by Vovk \cite{Vovk2005AlgorithmicWorld}, is a powerful approach for constructing valid PIs.
The most popular split conformal methods usually leverage prediction errors from a hold-out set to construct $\hat{C}(X_{N+1})$, which would be valid if a future data point $(X_{N+1}, Y_{N+1})$ follows the same distribution as data in the hold-out set.
This framework is particularly suitable for deep learning due to its distribution-free nature, and has motivated many recent uncertainty quantification efforts in deep learning for both classification and regression tasks \cite{pmlr-v119-alaa20a, linscrib, DBLP:journals/corr/abs-2009-14193,fisch2021efficient}.
However, most conformal methods are only marginally valid \cite{papadopoulos2002,Vovk2005AlgorithmicWorld,Lei2018Distribution-FreeRegression,Barber10.1214/20-AOS1965}.
Moreover, less-than-meticulous applications to DL can break distributional assumptions and theoretical validity, as in the case of \cite{pmlr-v119-alaa20a} (see discussion in Appendix). 
We however seek to construct a PI conditioning on the input (similar to the third PI in Fig. \ref{fig:intro:example}).
Some recent advances (\cite{guan2020conformal,covshift_tibshirani2020conformal}) examine the possibility of ``approximately'' conditionally valid PIs.
While these methods cannot be directly applied to DL due to efficiency and performance considerations, their methodological and theoretical contributions serve as major inspirations for us to develop a highly flexible and practical method.

\textbf{Summary of Contributions}:
We propose Locally Valid Discriminative Prediction Intervals (\methodname), a simple uncertainty estimation method for deep learning which combines recent advances in conformal prediction and the classical idea of kernel regression.
\methodname applies to almost all DL models, and is the first method that satisfies all four aforementioned requirements:
\vspace*{-2mm}
\begin{itemize}
    \item Validity: \methodname has frequentist coverage guarantee (not just marginal, but approximately conditional). 
    \item Discrimination: The width of the PIs given by \methodname adapts to the risk/uncertainty level of $X_{N+1}$.
    \item Scalability: \methodname is lightweight, adding limited overhead to the base DL model.
    \item Accuracy: \methodname is post-hoc without requiring model retraining, and does not affect the base performance of the DL model.
    \vspace*{-2mm}
\end{itemize}
We must note that while the theoretical foundation for guaranteeing "validity" is mostly based on \cite{guan2020conformal}, \methodname addresses several challenges to satisfy the other three requirements.
The code to replicate all our results can be found at \url{https://github.com/zlin7/LVD}.

%===============================================================
\section{Preliminaries}\label{sec:prelim}
\vspace{-5pt}

\subsection{Learning Setup and Assumptions}\label{subsec:prelim:learning}
We assume data and response pairs $(X, Y)\in \mathcal{X}\times \mathcal{Y}$ have a joint distribution denoted $\mathcal{P}$, with the marginal distributions of $Y$ and $X$ and the conditional distribution $Y|X$ denoted as $\mathcal{P}_Y$, $\mathcal{P}_X$, and $\mathcal{P}_{Y|X}$, respectively.
Further, we will define $Z_i \defeq (X_i, Y_i)$ for concision. 
%================================================================

Assuming that we already have an algorithm (with all the training protocols folded in), such as a Deep Neural Network (DNN), that provides a mean estimator $\hat{\mu}^{NN}(x): \mathcal{X}\mapsto\mathcal{Y}$.
Given a target coverage level $1-\alpha \in (0,1)$, our task is to also construct a prediction interval estimator function $\hat{C}_\alpha(x): \mathcal{X}\mapsto \{\text{subset of } \mathcal{Y}\}$ that has the \textbf{validity} and \textbf{discrimination} properties as defined below.

\subsection{Validity (Frequentist Coverage)}\label{subsec:prelim:coverage}
There are several (related) notions for a PI to be valid -- marginal, conditional, and local.
Given target level $1-\alpha$, we say $\hat{C}_\alpha$ has the \textbf{marginal coverage} guarantee (or, equivalently, is marginally valid) if
\begin{align}\label{eq:coverage:marginal} 
    \mathbb{P}\{Y_{N+1} \in \hat{C}_{\alpha}(X_{N+1})\} \geq 1-\alpha
\end{align}
where the probability is taken over the training data and (the unseen)  $(X_{N+1}, Y_{N+1})$. 

A limitation of marginal coverage is that it is not conditioned on $X_{N+1}$. 
A more desirable, albeit stronger, property would be \textbf{conditional coverage} at $1-\alpha$:
\begin{align}\label{eq:coverage:conditional} 
    \mathbb{P}\{Y_{N+1} \in \hat{C}_{\alpha}(X_{N+1}) | X_{N+1}=x\} \geq 1-\alpha \text{ for almost all $x\in\mathcal{X}$}.
\end{align}
Here the probability is taken over the training data and $Y_{N+1}$ (with $X_{N+1}$ fixed).
It is thus clear that conditional coverage implies marginal coverage but not the other way around. 
Indeed, a $\hat{C}_\alpha$ with marginal coverage property only implies a $1-\alpha$ chance of being accurate \emph{on average} across all data points (marginalizing over $X_{N+1}$) i.e. there might be a sub-population in the data for which the coverage is completely missed. 
Unfortunately, it is \textit{impossible} to achieve distribution-free finite-sample conditional coverage (Eq. \ref{eq:coverage:conditional}) in a non-trivial way. Indeed, it is known that a finite-sample estimated $\hat{C}_{\alpha}(x)$ cannot achieve conditional coverage, unless it produces infinitely wide prediction intervals in expectation under any non-discrete distribution $\mathcal{P}$ \cite{pmlr-v25-vovk12,lei2014, barber2020limits}. 

It is thus reasonable to instead seek \textbf{\emph{approximate}} conditional coverage.
As might be apparent, there is considerable freedom in defining an appropriate notion of ``approximate'', depending on specific tasks and domains.
However, a sufficiently general-purpose and natural notion involves using a kernel function $K:\mathcal{X}\times \mathcal{X}\mapsto\mathbb{R}$ and a center $x'\in\mathcal{X}$, like the relaxation given in \cite{covshift_tibshirani2020conformal}:
\begin{align}\label{eq:coverage:local_pre} 
    \frac{\int \mathbb{P}\{Y_{N+1} \in \hat{C}_{\alpha}(x') | X_{N+1} = x\}K(x, x') d\mathcal{P}_X(x)}{\int K(x, x') d\mathcal{P}_X(x)} \geq 1-\alpha
\end{align}
with the probability ($\mathbb{P}$ in the integral) taken over all training samples and $Y_{N+1}$, with $(X_{N+1}, Y_{N+1})\sim\tilde{\mathcal{P}}= \tilde{\mathcal{P}}_X \times \mathcal{P}_{Y|X}$.
Here $\tilde{\mathcal{P}}_X$ is just the distribution re-weighted by the kernel with a center $x'$, defined by $\frac{d\tilde{\mathcal{P}}_X(x)}{dx}\propto \frac{d\mathcal{P}_X(x)}{dx} K(x', x)$.
Instead of choosing $x'$ beforehand, if we let the center be $X_{N+1}$ and fold the integral into $\mathbb{P}$ like in \cite{guan2020conformal}, we arrive at the definition of \textbf{local coverage}:%\vspace{-8pt}
\begin{align}\label{eq:coverage:local} 
    \mathbb{P}\{\tilde{Y}_{N+1} \in \hat{C}_\alpha(X_{N+1})|X_{N+1}=x'\}  \geq 1-\alpha.
\end{align}
Here the probability integrates over all training data and an additional $(\tilde{X}_{N+1}, \tilde{Y}_{N+1})\sim\tilde{\mathcal{P}}$ defined above.
Intuitively, this definition means $\hat{C}_{\alpha}(X_{N+1})$ is valid ``on average'' within a small neighborhood of $X_{N+1}$.
Note that Eqs. \ref{eq:coverage:marginal} and \ref{eq:coverage:conditional} reduce to Eq. \ref{eq:coverage:local} with $K$ being constant and delta functions, respectively.
In the rest of the paper, we will call $\hat{C}_{\alpha}$ marginally/conditionally/locally valid if it satisfies Eq. \ref{eq:coverage:marginal}/\ref{eq:coverage:conditional}/\ref{eq:coverage:local} respectively, and we will pursue finite-sample \textbf{local validity}.

\subsection{Discrimination}\label{subsec:prelim:discrimination}
The idea of discrimination is simple: If the error of our prediction $\hat{\mu}(x)$ is high for an input $x$, the PI should be wide, and vice versa. 
Formally, following \cite{pmlr-v119-alaa20a}, we require
\begin{align}\label{eq:discrimination}
    \mathbb{E}[W(\hat{C}(x))] \geq \mathbb{E}[W(\hat{C}(x'))] \Leftrightarrow \mathbb{E}[\ell(y, \hat{\mu}(x))] \geq \mathbb{E}[\ell(y',\hat{\mu}(x'))].
\end{align}
Here the expectation is taken over the training data, $W$ is a measure of the width of the PI, and $\ell$ is a loss function such as MSE. 
This property can be verified (as shown in Section \ref{sec:exp}) by checking how well $W(\hat{C}(x))$ could predict the magnitude of the error.
Discrimination could be considered a measure of efficiency, as a good $\hat{C}$ could ``save'' some width when the expected risk is low.
However, it only makes sense to compare efficiency if all else is equal (i.e. two marginally valid PIs estimators with the same error).
Note that although discrimination could be related to conditional/local validity, they are not the same - e.g., a PI that is always infinitely wide is conditionally valid, but not discriminative. 

Our goal is to achieve both local validity and discrimination without making any assumptions about the underlying distribution $\mathcal{P}$ (i.e., in a distribution-free setting).
As noted in Section \ref{sec:intro}, our method should also run fast and not affect the performance of underlying neural network model $\hat{\mu}^{NN}$. 
%===================================================================================

%==============================================================
\section{Method: Locally Valid Disciminative Prediction Intervals (\methodname)}
\textbf{Overview}: %In essence, our proposed procedure is remarkably simple: 
We first train a deep neural network (DNN) $\hat{\mu}^{NN}$ (if not already given), followed by a post-hoc training of an appropriately chosen kernel function $K$. 
Specifically, we learn $K$ in a non-parametric kernel regression setting using embeddings from the deep learning model while optimizing for the underlying distance metric that the kernel function leverages. 
Both of these steps are explicated in more detail in Section \ref{subsec:method:training}.
Armed with $\hat{\mu}^{NN}$, we proceed to utilize a hold-out set to collect prediction residuals, which are used with the learned $K$ (along with its distance metric) to build the final PI for any datum at inference time (Section \ref{subsec:method:inference}). 
We then show the finite-sample local validity and asymptotic conditional validity in Section \ref{subsec:method:theory}.

\subsection{Training}\label{subsec:method:training}

At the onset, we partition $\fullset$ of $N$ data points into two sets - $\trainingset$ and $\validationset$. 
We will denote $\trainingset$ as $\{Z_i\}_{i=1}^n$ and $\validationset$ as $\{Z_{n+i}\}_{i=1}^m$, where $m = N-n$. $\trainingset$ is used to learn an embedding function $\embedder$ and a kernel $K$, and $\validationset$ is used for conformal prediction.

\textbf{[Optional] Training an Embedding Function}:
Instead of training a deep kernel in a kernel regression directly, which can be prohibitively expensive, we split the training task into two steps:
training the (expensive) DNN, and training the kernel $K$. 
Specifically, we first train a DNN mean estimator $\hat{\mu}^{NN}: \mathcal{X}\mapsto \mathcal{Y}$ to solve the supervised regression task with the mean squared error (MSE) loss. 
Note that $\hat{\mu}^{NN}$  can be based on any existing model. 
Moreover, this step could be skipped if we are already provided with a pre-trained $\hat{\mu}^{NN}$. 
Then, we remove the last layer of $\hat{\mu}^{NN}$ and produce an embedding function $\embedder: \mathcal{X} \mapsto \mathbb{R}^h$ for some positive integer $h$.
If the original model $\hat{\mu}^{NN}$ is good, usually such an embedding provides a rich and discriminative representation of the input (as will be verified empirically in Section \ref{sec:exp}).

\textbf{Training the Kernel}:
Fixing the embedding funtion $\embedder$, we perform leave-one-out Nadaraya-Watson \cite{Nadaraya}\cite{GyorfiNonParametric}\cite{WatsonKernel} kernel regression with a learnable Gaussian kernel on $\trainingset$:
\begin{align}\label{eq:kernreg}
    \hat{y}^{KR}_i  &= \frac{\sum_{j\neq i, j\in [n]} y_j K_\embedder(x_i, x_j)}{\sum_{j\neq i, j\in [n]} K_\embedder(x_i, x_j)} \\
    \text{where } K_\embedder(x_i, x_j) &= K(\embedder(x_i), \embedder(x_j)) = \frac{1}{\sigma \sqrt{2\pi}} e^{\frac{-d(\embedder(x_i), \embedder(x_j))}{\sigma^2}} 
    \text{ and } [n]\defeq\{1,\ldots,n\}.
\end{align}
Here $d(\cdot, \cdot)$ is a Mahalanobis distance parameterized by a positive-semidefinite matrix $\mathbf{W} \succeq 0$, which is learned.
To avoid solving an expensive semi-definite program, instead of working with $\mathbf{W}$ directly, we work with a low-rank matrix $\mathbf{A}\in\mathbb{R}^{h\times k}$ such that $\mathbf{W} = \mathbf{A}^T \mathbf{A}$, yielding the following equivalent distance formulation:
\vspace{-7pt}
\begin{align}\label{eq:dist}
    d(\embedder(x_i), \embedder(x_j)) = \|\mathbf{A}(\embedder(x_i) - \embedder(x_j))\|^2 .
\end{align}
This parameterization of $K$ is similar to that in \cite{pmlr-v2-weinberger07a}. Finally, to train $K$, we minimize the MSE loss.

\textbf{Residual Collection}: 
In this step, we take the trained embedding function and kernel, denoted as $K_\embedder$ for simplicity, and apply it on $\validationset$.
$\forall i\in[m]$, we compute and collect the absolute residual, 
\begin{align}
    R_i &= |y_{n+i} - \hat{y}_{n+i}|, 
\end{align}
the distribution of which is used for PI construction. It is important to remark that $\hat{y}$ does not have to be $\hat{y}^{KR}$.
The main purpose of the previous step is to train the $K$, and $\hat{y}$ could still be obtained through the original DNN $\hat{y} = \hat{\mu}^{NN}(x)$, or \textit{any} estimator not trained on $\validationset$.
As a result, the accuracy can only improve (if $\hat{y}^{KR}$ turns out to be a better mean estimator)\footnote{As will be shown in the Appendix, $\hat{y}^{KR}$ is often preferable because of the distance information it encodes.}.
\vspace{-5pt}
\subsection{Inference}\label{subsec:method:inference}
Before proceeding further, we recall a useful definition and fix some necessary notation.
For a distribution with cumulative density function (cdf) $F$ defined on the augmented real line $\mathbb{R}\cup\{-\infty, \infty\}$, the quantile function is defined as $Q(\alpha, F) = F^{-1}(\alpha)$.
This definition is the same for a finite distribution like the empirical distribution.
Suppose the empirical distribution consists of $R_1,\ldots,R_m$, then we denote the empirical distribution $\hat{F}$ and the empirical quantile $Q(\alpha, \hat{F})$ as:
\vspace{-5pt}
\begin{align}
   \hat{F} = \frac{1}{m}\sum_{i=1}^m \delta_{R_i} \qquad \text{and} \qquad Q(\alpha, \hat{F}) = \inf_r \hat{F}(r) \geq \alpha
\end{align}
where $\delta_{R}(r) = \mathbbm{1}\{ r \geq R\}$.
Note that we treat $\{R_i\}_{i=1}^m$ as an unordered list.
Besides, $R_i$ can be $\pm \infty$, and can repeat.
Finally, we can assign weights to $R_i$, and define the quantiles for a weighted distribution:
\vspace{-10pt}
\begin{align}
    \tilde{F} = \sum_{i=1}^m w_i \delta_{R_i} \qquad \text{where} \qquad \sum_{i=1}^m w_i = 1.
\vspace{-10pt}
\end{align}
\textbf{Split Conformal}:
Before presenting the detailed construction of the PI in \methodname, it would be particularly instructive to first consider a special case. 
Specifically, when $K$ returns a constant number for any $(x_i, x_j)$, we recover the well-known ``split conformal'' method~\cite{papadopoulos2002,Vovk2005AlgorithmicWorld,Lei2018Distribution-FreeRegression}, which uses the $1-\alpha$ quantile of the residuals as the PI width. 
Following our setup, the split conformal PI is given by:
\begin{align}\label{eq:split:PI}
    \hat{C}^{split}_{\alpha}(X_{N+1}) = \left\{y\in\mathbb{R}: |y - \hat{y}_{N+1}| \leq Q\left(1-\alpha, \frac{1}{m+1}\Bigg(\delta_{\infty} + \sum_{i=1}^m \delta_{R_i}\Bigg)\right)\right\}.
\end{align}
Because the residuals $\{R_i\}_{i\in[m]}\cup \{R_{N+1}\}$ are i.i.d., $R_{N+1}$'s ranking among them is uniformly distributed. 
We cannot know $R_{N+1}$, so we use $\infty$ instead to be ``safe'' ($\forall r\in\mathbb{R}, \delta_\infty(r) = 0$).
It follows that $\hat{C}^{split}_\alpha$ is ($1-\alpha$) marginally valid \cite{papadopoulos2002}. 

\textbf{Local Conformal}:
In order to achieve the local coverage property, all we need to do is to re-weigh the residuals. Following the approach in \cite{guan2020conformal}, we arrive at the following suitable notion of PI:
\begin{align}\label{eq:local:PI}
    \hat{C}^{\methodname}_{\alpha}(X_{N+1}) &= \left\{y\in\mathbb{R}: |y - \hat{y}_{N+1}| \leq Q\Bigg(1-\alpha, \Bigg( w_{N+1}\delta_{\infty} + \sum_{i=1}^m w_{n+i} \delta_{R_i}\Bigg)\Bigg)\right\}\\
    \text{where }\hfill w_j &= \frac{K_\embedder(x_{j}, x_{N+1})}{K_\embedder(x_{N+1}, x_{N+1})+\sum_{i=1}^m K_\embedder(x_{n+i}, x_{N+1})}.
\end{align}
In other words, we first assign weights to $\{R_i\}$ based on the similarity between $\{X_{n+i}\}$ and $X_{N+1}$ using $K_\embedder$, and then set the width to be the weighted quantile .
Note that with $\delta_{\infty}$, $\hat{C}^{\methodname}_{\alpha}(X_{N+1})$ will be infinitely wide if data is scarce around $X_{N+1}$.
However, as argued in \cite{guan2020conformal}, this is desired.

\subsection{Implementation Details}\label{subsec:method:implementation}
\textbf{Parameterization and Training}:
Since $\mathbf{A}$ is intricately linked to the computation of the weights assigned by the Gaussian kernel $K$ (Eq. \ref{eq:dist}), it is implemented as $K(\embedder(x_i), \embedder(x_j)) =e^{-||\mathbf{A}(\embedder(x_i)-\embedder(x_j))||^2}$. In order to optimize for $\mathbf{A}$, we treat it as a usual linear layer in a neural network and perform gradient descent.

\textbf{Smoothness Requirement}: 
In the context of obtaining locally valid prediction intervals, a potential drawback of using the Nadaraya-Watson kernel regression framework is that the $K_\embedder$ will not meaningfully learn the similarity of any input with \textit{itself}. 
For example, we can arbitrarily define $K_\embedder(x,x)$ to be \textit{any} value, including $\infty$. In the context of only regression, the fitted function's performance will not change as long as there are no two identical $x_i$.
With the Gaussian kernel, this issue is somewhat mitigated. 
However, during the training, the $K(x_i, x_i)$ can still be too high compared with $K(x_i, x_i+\epsilon)$, resulting in a less meaningful definition for local coverage.
We could then enforce a regularization by replacing the $\hat{y}_i$ in Eq. \ref{eq:kernreg} with
\begin{align}
    \hat{y}'^{KR}_i  &= \frac{ \overline{y}_{-i} K_\embedder(x_i,x_i) + \sum_{j\neq i, j\in [n]} y_j K_\embedder(x_i, x_j)}{K_\embedder(x_i,x_i)+\sum_{j\neq i, j\in [n]} K_\embedder(x_i, x_j)}
    & \text{where } & \overline{y}_{-i} = \frac{1}{n-1}\sum_{j\neq i, j\in[n]} y_j.
\end{align}
This can be considered an explicit bias term towards the (leave-one-out) sample mean. Empirically, we observe that enforcing this requirement is crucial to obtain meaningful and tight intervals. We direct the reader to the Appendix for a detailed ablation on its utility.

\textbf{Complexity}:
To facilitate training, we use stochastic gradient descent instead of gradient descent with batch size denoted as $B_1$. Furthermore, if the dataset size is prohibitively large, we can also randomly sample a subset of $B_2$ points  $\{x_j\}_{j\neq i}$ to predict $\hat{y}_i$. The total complexity is $O(B_1B_2 h k)$  where $h$ and $k$, defined earlier, denote the dimensionality of the embedding before/after it is multiplied by $\mathbf{A}$.
Note that $B_2 = O(1)$ or $o(N)$.
The inference time for each data point can be improved from $O(B_2 hk)$ to $O(B_2k + hk)$ by storing $\mathbf{A}x_j$ instead of $\mathbf{A}$ and $x_j$ separately. 

Denoting the number of parameter of the base NN as $P$, since DL models are usually overparameterized, the additional training time for each descent could be comparable or shorter than training the base NN (depending on the relation between $P$ and $B_2hk$)\footnote{
In practice, since $\embedder$ is already well-trained, the training of $K_\embedder$ converges very fast.
}, and the additional inference time would be much shorter than that of the base NN model.
In addition, most of these factors (especially $B_2$) can be easily parallelized.
The full procedure is summarized below in Algorithm \ref{alg:main}.
\AlgoMain
\vspace{-5pt}
\subsection{Theoretical Guarantees}\label{subsec:method:theory}
We conclude this section by showing that $\hat{C}^{\methodname}_\alpha$ provides the local coverage property.
We adapt  Theorem 5.1  in \cite{guan2020conformal} and results in \cite{covshift_tibshirani2020conformal} to our setting. The detailed proof is deferred to the Appendix:
\begin{theorem}\label{thm:coverage} %Thm 5.1 in \cite{guan2020conformal}
Conditional on $X_{N+1}$, the PI obtained from Algorithm \ref{alg:main}, $\hat{C}^{\methodname}_{\alpha}(X_{N+1})$, satisfies
\begin{align}
    \mathbb{P}\{\tilde{Y}_{N+1} \in \hat{C}^{\methodname}_{\alpha}(X_{N+1}) | X_{N+1} = x'\} \geq 1-\alpha \text{ for any } x'
\end{align}
where the probability is taken over all the training samples $\overset{i.i.d.}{\sim}\mathcal{P} = \mathcal{P}_{Y|X}\times \mathcal{P}_X$, and $(\tilde{X}_{N+1}, \tilde{Y}_{N+1})$ with distribution $\tilde{X}_{N+1}|X_{N+1}\sim \mathcal{P}_X^{X_{N+1}}$ and $\tilde{Y}_{N+1}|\tilde{X}_{N+1} \sim \mathcal{P}_{Y|X}$.
Here $\mathcal{P}_X^{X_{N+1}}$ means the localized distribution with $\frac{d\mathcal{P}_X^{X_{N+1}}(x)}{dx} \propto \frac{d\mathcal{P}_X(x)}{dx}K_\embedder(X_{N+1},x)$.
\end{theorem}

With some regularity assumptions like in \cite{lei2014}, we can also obtain asymptotic conditional coverage:
\begin{theorem}\label{thm:asymp_conditional} %Thm 5.1 in \cite{guan2020conformal}
With appropriate assumptions, $\hat{C}^{\methodname}$ is asymptotically conditional valid. 
\end{theorem}
The detailed assumptions, formal statement, and proof of Theorem \ref{thm:asymp_conditional} are deferred to the Appendix.

\textbf{Remark:}
Roughly speaking, Theorem \ref{thm:coverage} tells us that the response $Y$ of a new data point sampled ``near'' $X_{N+1}$ will fall in our PI with high probability. 
Theorem \ref{thm:asymp_conditional} further states that, under suitable assumptions and enough data, $\hat{C}^{\methodname}$ also covers $Y_{N+1}$ (i.e., no re-sampling) with high probability.

%===================================Experiment
\section{Experiments}\label{sec:exp}

\textbf{Baselines}:
We compare \methodname with the following baselines (with a qualitative comparison in Table \ref{table:exp:method_features}):  
\begin{enumerate}[leftmargin=*]
    \item \textit{Discriminative Jackknife (DJ)} \cite{pmlr-v119-alaa20a}, which claims to be both discriminative and marginally valid but is neither (See Appendix \ref{appendix:jkp}).
\item \textit{Deep Ensemble (DE)}~\cite{Lakshminarayanan2017SimpleEnsembles}, which trains an ensemble of networks to estimate variance and mean.
\item \textit{Monte-Carlo Dropout (MCDP)} \cite{Gal2016DropoutLearning}, a popular bayesian method for NN that performs Dropout \cite{Dropout_JMLR:v15:srivastava14a} at inference time for the predictive variance estimate.
\item \textit{Probabilistic Backpropagation (PBP)}~\cite{DBLP:conf/icml/Hernandez-Lobato15b}, a successful method to train Bayesian Neural Networks by computing a forward propagation of probabilities before a backward computation of gradients.
\item \textit{Conforamlized Quantile Regression (CQR)}~\cite{CQR_NEURIPS2019_5103c358}, an efficient (narrow PI) marginally valid conformal method that takes \textit{quantile} predictors instead of mean predictors.
This comes with a huge cost: one needs to retrain the predictor for \textit{each} $\alpha$ if more than one coverage level is desired. 
\item \textit{MAD-Normalized Split Conformal (MADSplit)}~\cite{Lei2018Distribution-FreeRegression, pmlr-v128-bellotti20a}, a variant of the well-known split-conformal method that requires an estimator for the mean absolute deviation (MAD), and performs conformal prediction on the MAD-normalized residuals.
\end{enumerate}\vspace{-5pt} 
In our experiments, PIs for non-valid methods are obtained from the quantile functions of the posterior for target coverage $1-\alpha$ like in \cite{pmlr-v119-alaa20a}. 
\TabFigMethodFeaturesAndSyntheticData

\subsection{Synthetic Data}\label{sec:exp:synthetic}

We will first examine the dynamics of different uncertainty methods with synthetic data.
The formula we use is the same as in \cite{DBLP:conf/icml/Hernandez-Lobato15b, pmlr-v119-alaa20a}: $y=x^3+\epsilon$. 
Here, $\epsilon \sim \mathcal{N}(0, 4^2)$, and $x$ comes from $ Unif[-1,1]$ with probability $0.9$, and half-normal distribution on $[1,\infty)$ with $\sigma=1$ with probability 0.1.
We used this $\mathcal{P}_X$ to illustrate local validity. The results are shown in Figure \ref{fig:exp:syn}. We observe that \methodname, CQR, MADSplit, and DJ all achieve close to 90\% coverage.
However, \methodname gives a more meaningful discriminative predictive band:
Specifically, near the boundaries, it will give us wider intervals (sometimes $\infty$) because there is little similar data around, which is desirable for local validity.
Although CQR and MADSplit can be discriminative, they are still only marginally valid, so we can see that despite the varying width, they actually get narrower when $x$ is more eccentric, which is clearly an issue.
DJ essentially gives PIs of constant width, as estimated from the quantile of the residuals.
DE also does not give meaningful uncertainty estimates, giving almost constant PIs that cover well below 90\%.
For Bayesian methods, MCDP behaves much like a Gaussian Process (as claimed in \cite{Gal2016DropoutLearning}), with low coverage rate, whereas PBP is mildly discriminative and not valid\footnote{Sometimes it may be possible to calibrate Bayesian methods~\cite{CQR_NEURIPS2019_5103c358}.
However, one needs to calibrate the entire posterior for the Bayesian method to makes sense.
Moreover, from our experiments, it is impossible to do this in MCDP, when it behaves like a Gaussian process and predicts zero variance near known data. }.

\subsection{Real Datasets}\label{sec:exp:real}
\TableCounts
We will be using a series of standard benchmark datasets in the uncertainty literature~\cite{pmlr-v119-alaa20a,CQR_NEURIPS2019_5103c358,DBLP:conf/icml/Hernandez-Lobato15b}, including:
UCI Yacht Hydrodynamics (Yacht)~\cite{UCI_Yacht}, 
UCI Bikesharing (Bike)~\cite{UCI_Bike},  
UCI Energy Efficiency (Energy)~\cite{UCI_Energy},
UCI Concrete Compressive Strength (Concrete)~\cite{UCI_Concrete}, 
Boston Housing (Housing)~\cite{Housing_data}, 
Kin8nm~\cite{Kin8nm}.%\footnote{https://www.openml.org/d/189}.
We also use QM8 (16 sub-tasks) and QM9 (12 sub-tasks)~\cite{QM8_doi:10.1063/1.4928757,QM9_1_doi:10.1021/ci300415d, QM9_2_ramakrishnan2014quantum} as examples of more complicated datasets.
In each experiment, 20\% of the data is used for testing. 
The sizes of datasets used are shown in Table \ref{table:exp:real:counts}.
We use the same DNN model for all baselines, which has 2 layers, 100 hidden nodes each layer, and ReLU activation for the non-QM datasets.
For QM8 and QM9, we use the molecule model implemented in \cite{chemproppaperdoi:10.1021/acs.jcim.9b00237} and apply applicable baselines.
Missing baselines (``\textendash'' in the tables) are either too expensive (i.e. time and/or memory) or require a significant redesign of the training and NN, which is beyond the scope of this paper.

\textbf{Evaluation Metrics}:
The evaluation is based on validity and discrimination. 
For validity, we check the marginal coverage rate (MCR) and the tail coverage rate (TCR), which is defined as the coverage rate for data whose $Y$ falls in the top and bottom 10\%. 
The motivation behind TCR is that if our local validity is very close to conditional validity, then \methodname's coverage rate would be above target in \textit{any} pre-defined sub-samples, including those with extreme $Y$s.
For discrimination, to verify Eq. \ref{eq:discrimination}, which is a prediction task, we compute the AUROC of using the PI width to predict whether the absolute residual is in the top half of all residuals.
AUROC alone is misleading, however, as a bad predictor can easily be discriminative (e.g., by randomly adding to both its prediction and PI width a huge constant).
Therefore, we also report the mean absolute deviation (MAD), defined as $\frac{\sum_{i=1}^M |\hat{y}_i - y_i|}{M}$.

\TableCoverage
\TableDisc

We repeat all experiments 10 times and report mean and standard deviations.
For QM8 and QM9, we report the average numbers across all sub-tasks, with a breakdown on each sub-task in the Appendix.

\textbf{Results}:
For \textbf{validity}, as shown in  Table \ref{table:exp:real:coverage}, \methodname achieves marginal coverage empirically, as well as MADSplit\footnote{It is worth noting that MADSplit, despite the theoretical guarantee, misses on the Yacht dataset, because the MAD-predictor predicts a ``negative'' absolute residual for some subset of the data, thus creating extremely narrow PIs, even after requiring the prediction to be positive and the ``practitioner's trick'' mentioned in \cite{CQR_NEURIPS2019_5103c358}.}
, CQR, and DJ. 
However, for tail coverage rate, only \methodname consistently covers at or above target coverage rates.
For the larger datasets, both coverage rates tend to get close to 90\% for \methodname.
DE, MCDP, and PBP do not achieve meaningful coverage (either too high or too low). \cite{DBLP:journals/corr/abs-2010-03039} also report mixed results on marginal validity with existing uncertainty quantification methods for DL. To further test for local validity, we also examine the average coverage rate conditioned on the presence of certain functional groups for the QM9 dataset (detailed results are relegated to the appendix). \methodname achieves empirical validity for these groups as well, even though functional groups define a kind of similarity that is never used in the uncertainty quantification process.

For \textbf{discrimination} (Table \ref{table:exp:real:discriminative}), \methodname is generally in the top two while maintaining the lowest MAD almost always.
MADSplit has the same MAD as \methodname (using the same $\hat{\mu}^{NN}$), and has a similar AUROC as \methodname despite explicitly modeling MAD.
Other baselines occasionally show a significant discriminative property, but usually have much higher MAD.
Despite training an ensemble of models, DE incurs huge prediction errors in many datasets.
As noted earlier, AUROC alone is misleading if the MAD is high: MCDP and CQR seem highly discriminative on the Bike dataset, mostly due to the high model error (epistemic uncertainty).

\noindent{\bf Scalability:}
For the largest dataset, QM9, the extra inference time of \methodname  vs. inference time of the original NN is 0.65 vs. 0.75 second per 1000 samples\footnote{
We use the full $\validationset$ for PI construction, as the inference time is short enough without sampling.
}
on an NVIDIA 2080Ti GPU.
MADSplit on the other hand takes 0.93 second overhead in the most optimized case.
That said, any method that finishes within $O(1)$ multiple of the original NN model is usable in practice.
For \methodname, extra vs. original training time is about 1.5 vs. 0.75 second per 1000 samples, but because the $\embedder$ is already highly informative,  the training of the kernel $K_\embedder$ finishes in very few iterations, resulting in $<$ 5\% overhead of the total training time.
Note that MADSplit will take strictly $\geq 1\times$ time in total because it needs to train a second model to predict residuals.
Like MADSplit, CQR needs to train at least one quantile predictor\footnote{That is, if one is willing to have mean estimate outside the PI occasionally and consider CQR post-hoc.}, but it needs to train a new predictor for every $\alpha$, which is a huge cost.
\vspace{-6pt}

\section{Conclusion}
\vspace{-6pt}
This paper introduces \methodname, the first locally valid and discriminative PI estimator for DL, which is also scalable and post-hoc. 
Because \methodname is both valid and discriminative, it can provide actionable uncertainty information for the real world application of DL regression models. 
Moreover, it is easy to apply \methodname to almost any DL model without any negative impact on the accuracy due to its post-hoc nature. 
Our experiments confirm that \methodname generates locally valid PIs that cover subgroups of data all other methods fail to.
It also exceeds or matches the performance in discriminative power while offering additional benefits in scalability and flexibility.
We foresee that \methodname can enable more real-world applications of DL models by providing users actionable uncertainty information.

\section*{Acknowledgments}
This work is in part supported by National Science Foundation award SCH-2014438, IIS-1418511, CCF-1533768, IIS-2034479, the National Institute of Health award NIH R01 1R01NS107291-01 and R56HL138415. The authors are also thankful to Andrew Gordon Wilson and three anonymous reviewers for their comments to help improve this work.

\newpage
\bibliography{main}

\begin{thebibliography}{10}

\bibitem{pmlr-v119-alaa20a}
Ahmed Alaa and Mihaela Van Der~Schaar.
\newblock {Discriminative Jackknife: Quantifying Uncertainty in Deep Learning
  via Higher-Order Influence Functions}.
\newblock In Hal~Daumé III and Aarti Singh, editors, {\em Proceedings of the
  37th International Conference on Machine Learning}, volume 119 of {\em
  Proceedings of Machine Learning Research}, pages 165--174. PMLR, 2020.

\bibitem{amodei2016concrete}
Dario Amodei, Chris Olah, Jacob Steinhardt, Paul Christiano, John Schulman, and
  Dan Mané.
\newblock Concrete problems in ai safety, 2016.

\bibitem{DBLP:journals/corr/abs-2009-14193}
Anastasios Angelopoulos, Stephen Bates, Jitendra Malik, and Michael~I. Jordan.
\newblock Uncertainty sets for image classifiers using conformal prediction.
\newblock {\em CoRR}, abs/2009.14193, 2020.

\bibitem{Barber10.1214/20-AOS1965}
Rina~Foygel Barber, Emmanuel~J Cand{\`{e}}s, Aaditya Ramdas, and Ryan~J
  Tibshirani.
\newblock {Predictive inference with the jackknife+}.
\newblock {\em The Annals of Statistics}, 49(1):486--507, 2021.

\bibitem{barber2020limits}
Rina~Foygel Barber, Emmanuel~J. Candès, Aaditya Ramdas, and Ryan~J.
  Tibshirani.
\newblock The limits of distribution-free conditional predictive inference.
\newblock {\em arXiv}, abs/1903.04684, 2020.

\bibitem{basu2021influence}
Samyadeep Basu, Philip Pope, and Soheil Feizi.
\newblock {Influence Functions in Deep Learning Are Fragile}, 2021.

\bibitem{BayesianFrequentist_10.1214/088342304000000116}
M.~J. Bayarri and J.~O. Berger.
\newblock {The Interplay of Bayesian and Frequentist Analysis}.
\newblock {\em Statistical Science}, 19(1):58 -- 80, 2004.

\bibitem{pmlr-v128-bellotti20a}
Anthony Bellotti.
\newblock Constructing normalized nonconformity measures based on maximizing
  predictive efficiency.
\newblock In Alexander Gammerman, Vladimir Vovk, Zhiyuan Luo, Evgueni Smirnov,
  and Giovanni Cherubin, editors, {\em Proceedings of the Ninth Symposium on
  Conformal and Probabilistic Prediction and Applications}, volume 128 of {\em
  Proceedings of Machine Learning Research}, pages 41--54. PMLR, 09--11 Sep
  2020.

\bibitem{Housing_data}
The boston housing dataset.
\newblock \url{http://lib.stat.cmu.edu/datasets/boston}.
\newblock Accessed: 2021-05-27.

\bibitem{UCI_Bike_paper}
Hadi Fanaee-T and Joao Gama.
\newblock Event labeling combining ensemble detectors and background knowledge.
\newblock {\em Progress in Artificial Intelligence}, pages 1--15, 2013.

\bibitem{fisch2021efficient}
Adam Fisch, Tal Schuster, Tommi~S. Jaakkola, and Regina Barzilay.
\newblock Efficient conformal prediction via cascaded inference with expanded
  admission.
\newblock In {\em International Conference on Learning Representations}, 2021.

\bibitem{Gal2016DropoutLearning}
Yarin Gal and Zoubin Ghahramani.
\newblock {Dropout as a Bayesian approximation: Representing model uncertainty
  in deep learning}.
\newblock In {\em 33rd International Conference on Machine Learning, ICML
  2016}, 2016.

\bibitem{guan2020conformal}
Leying Guan.
\newblock Conformal prediction with localization.
\newblock {\em arXiv}, abs/1908.08558, 2020.

\bibitem{GyorfiNonParametric}
L\'aszl\'o Gy\"orfi, Michael Kohler, Adam Krzyzak, and Harro Walk.
\newblock {\em {A Distribution-Free Theory of Nonparametric Regression}}.
\newblock Springer Science \& Business Media, 2006.

\bibitem{DBLP:conf/icml/Hernandez-Lobato15b}
Jos{\'{e}}~Miguel Hern{\'{a}}ndez{-}Lobato and Ryan~P. Adams.
\newblock Probabilistic backpropagation for scalable learning of bayesian
  neural networks.
\newblock In Francis~R. Bach and David~M. Blei, editors, {\em Proceedings of
  the 32nd International Conference on Machine Learning, {ICML} 2015, Lille,
  France, 6-11 July 2015}, volume~37 of {\em {JMLR} Workshop and Conference
  Proceedings}, pages 1861--1869. JMLR.org, 2015.

\bibitem{Kin8nm}
Kin family of datasets.
\newblock \url{http://www.cs.toronto.edu/~delve/data/kin/desc.html}.
\newblock Accessed: 2021-05-27.

\bibitem{ADAM_DBLP:journals/corr/KingmaB14}
Diederik~P. Kingma and Jimmy Ba.
\newblock Adam: {A} method for stochastic optimization.
\newblock In Yoshua Bengio and Yann LeCun, editors, {\em 3rd International
  Conference on Learning Representations, {ICLR} 2015, San Diego, CA, USA, May
  7-9, 2015, Conference Track Proceedings}, 2015.

\bibitem{influence_pmlr-v70-koh17a}
Pang~Wei Koh and Percy Liang.
\newblock Understanding black-box predictions via influence functions.
\newblock In Doina Precup and Yee~Whye Teh, editors, {\em Proceedings of the
  34th International Conference on Machine Learning}, volume~70 of {\em
  Proceedings of Machine Learning Research}, pages 1885--1894. PMLR, 06--11 Aug
  2017.

\bibitem{DBLP:journals/corr/abs-2010-03039}
Benjamin Kompa, Jasper Snoek, and Andrew Beam.
\newblock Empirical frequentist coverage of deep learning uncertainty
  quantification procedures.
\newblock {\em CoRR}, abs/2010.03039, 2020.

\bibitem{Lakshminarayanan2017SimpleEnsembles}
Balaji Lakshminarayanan, Alexander Pritzel, and Charles Blundell.
\newblock {Simple and scalable predictive uncertainty estimation using deep
  ensembles}.
\newblock In {\em Advances in Neural Information Processing Systems}, 2017.

\bibitem{Lei2018Distribution-FreeRegression}
Jing Lei, Max G’Sell, Alessandro Rinaldo, Ryan~J. Tibshirani, and Larry
  Wasserman.
\newblock {Distribution-Free Predictive Inference for Regression}.
\newblock {\em Journal of the American Statistical Association}, 2018.

\bibitem{lei2014}
Jing Lei and Larry Wasserman.
\newblock Distribution-free prediction bands for non-parametric regression.
\newblock {\em Journal of the Royal Statistical Society: Series B (Statistical
  Methodology)}, 76(1):71--96, 2014.

\bibitem{linscrib}
Zhen Lin, Cao Xiao, Lucas Glass, M.~Brandon Westover, and Jimeng Sun.
\newblock {SCRIB:} set-classifier with class-specific risk bounds for blackbox
  models.
\newblock {\em CoRR}, abs/2103.03945, 2021.

\bibitem{Nadaraya}
Elizbar Nadaraya.
\newblock {\em {Nonparametric Estimation of Probability Densities and
  Regression Curves}}.
\newblock Kluwer Academic Publishers, 1989.

\bibitem{papadopoulos2002}
Harris Papadopoulos, Kostas Proedrou, Volodya Vovk, and Alex Gammerman.
\newblock Inductive confidence machines for regression.
\newblock In Tapio Elomaa, Heikki Mannila, and Hannu Toivonen, editors, {\em
  Machine Learning: ECML 2002}, pages 345--356, Berlin, Heidelberg, 2002.
  Springer Berlin Heidelberg.

\bibitem{PyTorch_NEURIPS2019_9015}
Adam Paszke, Sam Gross, Francisco Massa, Adam Lerer, James Bradbury, Gregory
  Chanan, Trevor Killeen, Zeming Lin, Natalia Gimelshein, Luca Antiga, Alban
  Desmaison, Andreas Kopf, Edward Yang, Zachary DeVito, Martin Raison, Alykhan
  Tejani, Sasank Chilamkurthy, Benoit Steiner, Lu~Fang, Junjie Bai, and Soumith
  Chintala.
\newblock Pytorch: An imperative style, high-performance deep learning library.
\newblock In H.~Wallach, H.~Larochelle, A.~Beygelzimer, F.~d\textquotesingle
  Alch\'{e}-Buc, E.~Fox, and R.~Garnett, editors, {\em Advances in Neural
  Information Processing Systems 32}, pages 8024--8035. Curran Associates,
  Inc., 2019.

\bibitem{QM9_2_ramakrishnan2014quantum}
Raghunathan Ramakrishnan, Pavlo~O Dral, Matthias Rupp, and O~Anatole von
  Lilienfeld.
\newblock Quantum chemistry structures and properties of 134 kilo molecules.
\newblock {\em Scientific Data}, 1, 2014.

\bibitem{QM8_doi:10.1063/1.4928757}
Raghunathan Ramakrishnan, Mia Hartmann, Enrico Tapavicza, and O.~Anatole von
  Lilienfeld.
\newblock Electronic spectra from tddft and machine learning in chemical space.
\newblock {\em The Journal of Chemical Physics}, 143(8):084111, 2015.

\bibitem{CQR_NEURIPS2019_5103c358}
Yaniv Romano, Evan Patterson, and Emmanuel Candes.
\newblock Conformalized quantile regression.
\newblock In H.~Wallach, H.~Larochelle, A.~Beygelzimer, F.~d\textquotesingle
  Alch\'{e}-Buc, E.~Fox, and R.~Garnett, editors, {\em Advances in Neural
  Information Processing Systems}, volume~32. Curran Associates, Inc., 2019.

\bibitem{QM9_1_doi:10.1021/ci300415d}
Lars Ruddigkeit, Ruud van Deursen, Lorenz~C. Blum, and Jean-Louis Reymond.
\newblock Enumeration of 166 billion organic small molecules in the chemical
  universe database gdb-17.
\newblock {\em Journal of Chemical Information and Modeling},
  52(11):2864--2875, 2012.
\newblock PMID: 23088335.

\bibitem{Dropout_JMLR:v15:srivastava14a}
Nitish Srivastava, Geoffrey Hinton, Alex Krizhevsky, Ilya Sutskever, and Ruslan
  Salakhutdinov.
\newblock Dropout: A simple way to prevent neural networks from overfitting.
\newblock {\em Journal of Machine Learning Research}, 15(56):1929--1958, 2014.

\bibitem{GlobalChem}
Sul and Elena Chow.
\newblock Globalchem: A content variable store for chemistry!
\newblock \url{https://github.com/Sulstice/global-chem}, 2021.

\bibitem{covshift_tibshirani2020conformal}
Ryan~J. Tibshirani, Rina~Foygel Barber, Emmanuel~J. Candes, and Aaditya Ramdas.
\newblock Conformal prediction under covariate shift, 2020.

\bibitem{UCI_Energy_paper_TSANAS2012560}
Athanasios Tsanas and Angeliki Xifara.
\newblock Accurate quantitative estimation of energy performance of residential
  buildings using statistical machine learning tools.
\newblock {\em Energy and Buildings}, 49:560--567, 2012.

\bibitem{UCI_Bike}
Bike sharing data set.
\newblock \url{https://archive.ics.uci.edu/ml/datasets/Bike+Sharing+Dataset}.
\newblock Accessed: 2021-05-27.

\bibitem{UCI_Concrete}
Concrete compressive strength data set.
\newblock
  \url{http://archive.ics.uci.edu/ml/datasets/concrete+compressive+strength}.
\newblock Accessed: 2021-05-27.

\bibitem{UCI_Energy}
Energy efficiency data set.
\newblock \url{https://archive.ics.uci.edu/ml/datasets/energy+efficiency}.
\newblock Accessed: 2021-05-27.

\bibitem{UCI_Yacht}
Yacht hydrodynamics data set.
\newblock \url{http://archive.ics.uci.edu/ml/datasets/yacht+hydrodynamics}.
\newblock Accessed: 2021-05-27.

\bibitem{pmlr-v25-vovk12}
Vladimir Vovk.
\newblock Conditional validity of inductive conformal predictors.
\newblock In Steven C.~H. Hoi and Wray Buntine, editors, {\em Proceedings of
  the Asian Conference on Machine Learning}, volume~25 of {\em Proceedings of
  Machine Learning Research}, pages 475--490, Singapore Management University,
  Singapore, 04--06 Nov 2012. PMLR.

\bibitem{Vovk2005AlgorithmicWorld}
Vladimir Vovk, Alexander Gammerman, and Glenn Shafer.
\newblock {\em {Algorithmic learning in a random world}}.
\newblock Springer US, 2005.

\bibitem{WatsonKernel}
Geoffrey~S. Watson.
\newblock Smooth regression analysis.
\newblock {\em Sankhyā: The Indian Journal of Statistics, Series A
  (1961-2002)}, 26(4):359--372, 1964.

\bibitem{pmlr-v2-weinberger07a}
Kilian~Q. Weinberger and Gerald Tesauro.
\newblock Metric learning for kernel regression.
\newblock In Marina Meila and Xiaotong Shen, editors, {\em Proceedings of the
  Eleventh International Conference on Artificial Intelligence and Statistics},
  volume~2 of {\em Proceedings of Machine Learning Research}, pages 612--619,
  San Juan, Puerto Rico, 21--24 Mar 2007. PMLR.

\bibitem{Welling2011BayesianDynamics}
Max Welling and Yee~Whye Teh.
\newblock {Bayesian learning via stochastic gradient langevin dynamics}.
\newblock In {\em Proceedings of the 28th International Conference on Machine
  Learning, ICML 2011}, 2011.

\bibitem{AGDeepEnsemble}
Andrew~Gordon Wilson and Pavel Izmailov.
\newblock Bayesian deep learning and a probabilistic perspective of
  generalization.
\newblock In Hugo Larochelle, Marc'Aurelio Ranzato, Raia Hadsell,
  Maria{-}Florina Balcan, and Hsuan{-}Tien Lin, editors, {\em Advances in
  Neural Information Processing Systems 33: Annual Conference on Neural
  Information Processing Systems 2020, NeurIPS 2020, December 6-12, 2020,
  virtual}, 2020.

\bibitem{chemproppaperdoi:10.1021/acs.jcim.9b00237}
Kevin Yang, Kyle Swanson, Wengong Jin, Connor Coley, Philipp Eiden, Hua Gao,
  Angel Guzman-Perez, Timothy Hopper, Brian Kelley, Miriam Mathea, Andrew
  Palmer, Volker Settels, Tommi Jaakkola, Klavs Jensen, and Regina Barzilay.
\newblock {Analyzing Learned Molecular Representations for Property
  Prediction}.
\newblock {\em Journal of Chemical Information and Modeling}, 59(8):3370--3388,
  2019.

\bibitem{UCI_Concrete_paper_YEH19981797}
I.-C. Yeh.
\newblock Modeling of strength of high-performance concrete using artificial
  neural networks.
\newblock {\em Cement and Concrete Research}, 28(12):1797--1808, 1998.

\end{thebibliography}
\bibliographystyle{plain}

\newpage
\appendix
%================================================Proof
\section{Proofs}
\subsection{Proof for Theorem \ref{thm:coverage}}\label{appendix:subsec:proof:proof_local}
In this section, we will prove Theorem \ref{thm:coverage}. 
The key idea behind the proof is that since $\hat{\mu}$ (let it be $\hat{\mu}^{NN}$ or $\hat{\mu}^{KR}$) and $K_\embedder$ are independent of $\validationset$, the residuals ($R_i$) collected on $\validationset$ follow the same distribution as a new test residual. 
Thus, re-weighting $\mathcal{P}_X$ by $K_\embedder$ precisely fits into the covariate-shift setting studied by \cite{covshift_tibshirani2020conformal} (further studied in \cite{guan2020conformal}), which in turn implies that under the new (localized) distribution for $X$, the coverage guarantee holds (Theorem \ref{thm:coverage}). 

We first introduce a few definitions following the same notation as in \cite{guan2020conformal}.
To begin, we define the score function as $V(x,y) \defeq |y - \hat{\mu}(x)|$. 
Then, we write the localizer function as $H(x, x') \defeq K_\embedder(x, x')$.
For convenience, we also will rewrite subscripts of the data, so we have $Z'_1=(X'_1,Y'_1), \ldots,  Z'_{m+1}=(X'_{m+1}, Y'_{m+1})$, where $Z'_i$ is just $Z_{n+i}$ for $i\in[m]$ and $Z'_{m+1}$ is $Z_{N+1}$.
For $\{Z'_i\}_{i=1}^{m+1}$, both $V$ and $H$ would be considered \textit{fixed} because the training did not use any information from $\validationset$. 
\cite{guan2020conformal} allows for a more general form of $H$ that can depend on $\validationset$, which is not needed in our setting.

We proceed to define the weighted residual distributions like in \cite{guan2020conformal}: 
\begin{align}
    \hat{\mathcal{F}}_i &\defeq \sum_{j=1}^{m+1} p_{i,j}^H \delta_{V(X'_j, Y'_j)} \\
    \text{where } p^H_{i,j} &\defeq \frac{H(X'_i, X'_j)}{\sum_{k=1}^{m+1} H(X'_i, X'_k)}
\end{align}
Finally, $\hat{\mathcal{F}}$ is defined as $p^H_{m+1,m+1}\delta_\infty + \sum_{i=1}^m p^H_{m+1,i}\delta_{V(X'_i, Y'_i)}$.
$V(X'_{m+1}, Y'_{m+1})$ can be considered set to $\infty$, because we don't know the value of $Y'_{m+1}$ and want to be conservative.

Now, our construction of the PI could be rewritten in the following form:
\begin{align}
    \hat{C}_\alpha^{\methodname}(X'_{m+1}) &\defeq \{y: V(X'_{m+1}, y) \leq Q(1-\alpha, \hat{\mathcal{F}})\}
\end{align}
This is precisely the setup of Theorem 5.1 in \cite{guan2020conformal}, and Theorem \ref{thm:coverage} follows from Theorem 5.1 in \cite{guan2020conformal}.

%===================
\subsection{Asymptotic Conditional Validity (Theorem \ref{thm:asymp_conditional})}\label{appendix:subsec:proof:proof_asymp_cond}
Before we discuss the asymptotic property of $\hat{C}^{\methodname}$, we formally define asymptotic conditional validity (from \cite{lei2014}).
\begin{definition}
(Asymptotic Conditional Validity)
Given training data $(X_1, Y_1),\ldots,(X_m, Y_m)$, a PI estimator $\hat{C}_{m,\alpha}$ is asymptotically conditionally valid if 
\begin{align}
    \sup_{x} \Big[\mathbb{P}\{Y_{m+1} \not\in C_{m,\alpha}(x) | X_{m+1} = x\} - \alpha\Big]_{+} \overset{\mathcal{P}}{\to} 0 
\end{align}
as $m\to \infty$, where the sup is taken over the support of $\mathcal{P}_X$.
\end{definition}
Here, we add the subscript $m$ to $\hat{C}_{\alpha}$ to emphasize the dependence on the sample size.
If a PI estimator is asymptotically conditionally valid at level $1-\alpha$, then given enough samples (as $m\to\infty$), the probability of $\hat{C}_{m,\alpha}$ missing the next response $Y_{m+1}$ converges to $\alpha$ in probability. Note that LVD has an implicit assumption that the embedding function $\embedder$ (after some transformation) maps similar data close together. But with Theorem \ref{thm:asymp_conditional} such an assumption is not critical as the size of the dataset increases.

To facilitate the discussion, we add a subscript $m$ and denote the PI given by \methodname as $\hat{C}^{\methodname}_{m, \alpha}$. 
With the setup mentioned in Section \ref{appendix:subsec:proof:proof_local}, we obtain a result similar to Theorem 5.1 (b) in \cite{guan2020conformal} to for $\hat{C}^{\methodname}_{m, \alpha}$ as well. 

\textbf{Assumptions}:
We need to make the following assumptions:
\begin{enumerate}[label=(\arabic*)]
    \item 
    Denote $W \defeq \embedder(X)$ as a new random variable in $\mathbb{R}^h$. 
    $W$ is (assumed to be) on $[0,1]^h$ with marginal density bounded from two sides by two constants $b_1 < b_2$. 
    In other words, $0<b_1 \leq p_{W}(w) \leq b_2 < \infty$. 
    
    \item 
    The conditional density of $R$ (the residual) given $W$ is Lipschitz in $W$.
    In other words, $\forall w, w'$, $\|p_{R|W}(\cdot|w) - p_{R|W}(\cdot|w')\|_\infty \leq L\|w-w'\|$.
    
\end{enumerate}
As might be clear, (1) and (2) are standard regularity assumptions (as in \cite{guan2020conformal,lei2014}), but stated for our setting. 
For assumption (1), if $W$ does not fall in $[0,1]^h$, we can easily fix it by adding a normalization layer to $\embedder$. 
Compared with \cite{guan2020conformal,lei2014}, (2) is not any less likely to hold, as we usually only have one linear layer after $\embedder$ in $\hat{\mu}^{NN}$. 

\textbf{Bandwidth ($h$)}: 
To clearly state the theorem, we also need to decompose/unfold our transform matrix $\mathbf{A}$ into two steps - projection and rescaling:
$\mathbf{A}(w-w')\defeq \frac{1}{h} \mathbf{A_1}(w-w')$, where $\|\mathbf{A}_1\|_2 = 1$.
Note that in our learning, we are mostly learning $\mathbf{A}_1$, and $h$ is in fact \textit{chosen}.
In our experiment, we implicitly folded $h$ into $\mathbf{A}$, as changing $h$ entails making an explicit decision on how ``local'' one wants the coverage to be when the data is limited, and we do not have a strong prior on this.
However, for the sake of this discussion, as $N\to\infty$, if we keep the same ratio between $n=|\trainingset|$ and $m=|\validationset|$, then:
\begin{itemize}
    \item $\mathbf{A}_1$ would converge to some fixed unit-norm matrix in $\mathbb{R}^{h\times k}$, and
    \item we could let $h\to0$ like in \cite{guan2020conformal} and \cite{lei2014}, because if we have $m\to\infty$, then the number of validation residuals is large, so we could afford a much more ``local'' validity with few infinitely wide PIs.
\end{itemize}

%==========
With the assumptions stated above, we are in a position to state the following theorem regarding the asymptotic conditional validity of $\hat{C}^{\methodname}_{m,\alpha}$:
\begin{theorem}\label{appendix:thm:asymp_conditional} 
(Asymptotic Conditional Validity. Re-statement of Theorem \ref{thm:asymp_conditional}):
With assumptions (1) and (2), and $m\to\infty$, if we also let $h\to0$, then
\begin{align}
    \Big[\alpha - \mathbb{P}\{Y'_{m+1} \in \hat{C}^{\methodname}_{m, \alpha}(X'_{m+1}) \}\Big]_+ \overset{\mathcal{P}}{\to} 0.
\end{align}
\end{theorem}
The proof is essentially the same as that in \cite{guan2020conformal}, with the key difference that in \cite{guan2020conformal}, the Gaussian kernel only has one bandwidth $h$, which goes to 0 asymptotically. 
This has been discussed in the ``Bandwidth (h)'' section above.
Note the key difference between Theorem \ref{appendix:thm:asymp_conditional} and \ref{thm:coverage} is that the response $Y'_{m+1}$ now belongs to $X'_{m+1}$, which is used to construct the PI.

%================================================Experiment Details
\clearpage
\section{Additional Experimental Details}
\subsection{Training Details}
As noted in the paper, the DNN used for most datasets (except QM8 and QM9) has 2 layers, 100 hidden nodes, and uses ReLU for the activation function.
This is the same architecture as in \cite{pmlr-v119-alaa20a}, but with the difference that the activation is ReLU instead of tanh. We make this choice because the code accompanying \cite{pmlr-v119-alaa20a} uses ReLU, and tanh does not train for most of the datasets in our experiments.
Recall that the learnable matrix $\mathbf{A}$ reduces dimension from $h$ to $k$. 
For QM8 and QM9, please refer to \cite{chemproppaperdoi:10.1021/acs.jcim.9b00237} for a detailed description of the architecture and training protocols.
We make the following modifications in order to run some baselines:
\begin{itemize}
    \item MADSplit: We train a second model after the model in \cite{chemproppaperdoi:10.1021/acs.jcim.9b00237} that has the same architecture and training protocol, but tries to predict the absolute error of the first model.
    \item CQR: We replace the MSE loss with the ``pinball'' loss mentioned in \cite{CQR_NEURIPS2019_5103c358} and simultaneously train two quantiles for the same $\alpha$. For different $\alpha$, we re-train a model.
    \item DE: We replace the loss with the negative log-likelihood (NLL) loss as suggested in \cite{Lakshminarayanan2017SimpleEnsembles}, and train an ensemble of 5 models for each experiment.
    
\end{itemize}

For all experiments, $h$ is given by the DNN, and we set $k=10$. 
The training of the kernel follows the following protocol: we first take embedding from the training data, compute and fix the mean $\mu_i$ and standard deviation $s_i$ for each dimension $i\in[h]$. 
Dimensions with standard deviation $<$1e-3 are ignored as they are most likely dead nodes (due to ReLU). 
The embeddings are then always normalized using $\mu_i$ and $s_i$ before passing through $\mathbf{A}$.

$\mathbf{A}$ is implemented as a \texttt{torch.nn.Linear} layer using PyTorch\cite{PyTorch_NEURIPS2019_9015} and follows the default initialization. 
We restrict the kernel regression to use the top 3000 (or all) similar data points so the computation can be fast (like in \cite{pmlr-v2-weinberger07a}).
We use an Adam optimizer \cite{ADAM_DBLP:journals/corr/KingmaB14} implemented in PyTorch, with a learning rate set to 1e-2, and batch size 100. 
We repeat the process for 1000 up to batches, and stop early if the loss does not improve for 50 consecutive batches. 

For each setup, we repeat the experiment 10 times by randomly re-splitting training, validation, and test set with random seed from 0 to 9. 
For \methodname, MADSplit, and CQR (which require a hold-out set for conformal prediction), we use 60\% for training, 20\% for validation/hold-out set, and 20\% for test.
For all other methods, we use 80\% for training and 20\% for testing.

\subsection{Average PI Width}
It is hard to compare efficiency because \methodname achieves a much more demanding type of coverage, MADSplit and CQR achieve marginal coverage, and the rest of the methods are not valid (thus not comparable).
We thus restrict the comparison to only valid methods (\methodname, MADSplit, and CQR) and the subset of data for which all PIs are finite in Table \ref{table:exp:real:width}.
We can see that, as expected, \methodname tends to give infinite PI for small datasets at 90\% target level (``\# finite'' is low for a few datasets), because it requires some weighted observation in a local neighborhood.
(Note that the \# of finite PIs could be tuned by a bandwidth $h$ as discussed in Section \ref{appendix:subsec:proof:proof_asymp_cond}.)
However, despite providing a stronger coverage guarantee, \methodname still managed to be the most efficient on Bike and QM9. 

The most efficient method seems to be CQR, but the results are not very stable (very wide PIs for CQR in the Bike dataset, for example), and most of the time the difference in average width is not significant. 
However, as noted earlier in the main text, the potential efficiency of CQR comes with a huge cost: CQR requires re-training the model for each $\alpha$.
Moreover, there is no guarantee that the estimate of the lower bound of the PI is actually lower than the upper bound (``quantile crossing'', see \cite{CQR_NEURIPS2019_5103c358}), nor that a mean estimate actually falls in the PI either. 
In our experiments, we had to take the mean of the lower and upper bound as the mean estimator to ensure the mean estimator is always within the PI. 

We also include the average width of all baselines in Table  \ref{appendix:table:exp:real:widthinf} for reference, although it is not very meaningful to compare valid and non-valid methods.
\TableWidth
%\TableWidthAppendix
\TableWidthInfAppendix

\subsection{Additional Results of Different Variants of \methodname}
Although we consider MADSplit as a baseline, our method could be combined with it as well, by simply replacing $R_i$ with a normalized $R'_i\defeq \frac{y_{n+i}-\hat{y}_{n+i}}{\hat{\sigma}(x_{n+i})}$ like that in MADSplit.
One key observation is that using embedding given by a pre-trained DL model can simultaneously keep most of the performance of the base model and combine it with many conformal methods with acceptable overhead.

In this section, we will change different settings of \methodname and compare the effects.
Specifically, there are 3 independent choices:
\begin{itemize}
    \item Whether we use the kernel regression prediction $\hat{y}^{KR}$ or the base DNN predictor $\hat{\mu}^{NN}$ (KR vs. NN)
    \item Whether we apply the smoothness requirement as mentioned in Section \ref{subsec:method:implementation} (No-smooth vs. Smooth)
    \item Whether we normalize the residuals by an extra prediction of MAD or not. 
    We will denote the version described in the main text as ``base''.
    For the MAD-Normalized case (``MN''), similar to MADSplit \cite{Lei2018Distribution-FreeRegression, pmlr-v128-bellotti20a}, the non-conformity score, and the final PI construction, are replaced by 
    \begin{align}
        R'_i&\defeq \frac{y_{n+i}-\hat{y}_{n+i}}{\hat{\sigma}(x_{n+i})}\\
        \hat{C}^{MN}_\alpha(X_{N+1} &\defeq \left\{y\in\mathbb{R}: |y - \hat{y}_{N+1}| \leq \frac{1}{\hat{\sigma}(X_{N+1})}Q\left(1-\alpha, w_{N+1}\delta_{\infty} + \sum_{i=1}^m w_{n+i} \delta_{R'_i}\right)\right\}
    \end{align}
    This potentially can make the PI more discriminative by modeling the heteroscedasticity explicitly. 
\end{itemize}
As a reminder, all results shown in the main text are using $\hat{\mu}^{NN}$, with smoothing, and not normalized by MAD prediction (NN, Smooth, NM).
Also, all choices will not break any theoretical guarantees, including Theorem \ref{thm:coverage} and \ref{thm:asymp_conditional}. 

The results are presented in Table \ref{appendix:table:exp:real:cov_variant} and \ref{appendix:table:exp:real:disc_variant}, with the version shown in the main text boxed. 
All methods achieve target coverage rates as measured by MCR and TCR empirically.  
In general, we found that using $\hat{y}^{KR}$ tends to give higher AUROC, with similar or lower MAD. 
It should be noted that the MAD prediction in ``MN'' requires a base classifier, which is $\hat{y}^{NN}$ in our case. 
In other words, there is a mismatch in the ``MN'' version with $\hat{y}^{KR}$.
We conjecture that if the MAD predictor is properly trained for $\hat{y}^{KR}$, the AUROC for this combination would be even higher (at no cost to other metrics).
\TableCoverageVariants
\TableDiscVariants
We also include the average width and count of finite PIs in Table \ref{appendix:table:exp:real:fifty_width_count} and \ref{appendix:table:exp:real:ninty_width_count}. 
For most experiments adding smoothness requirement and using $\hat{y}^{KR}$ seems to achieve narrow PI, high AUROC, and low MAD. 
As noted earlier, training a separate model to model the residual of $\hat{y}^{KR}$ might give additional discrimination (and possibly narrower PIs as well).
%In general, we conclude that using $\hat{y}^{KR}$, with the smoothness requirement, and potentially adding ``MN'' using a MAD predictor trained for $\hat{y}^{KR}$ seems to achieve the best trade-off among all metrics.
\TableFiftyVariants
\TableNintyVariants

\subsection{Additional Results on QM8/QM9 sub-tasks}
Table \ref{appendix:table:exp:real:qm_cov} and \ref{appendix:table:exp:real:qm_disc} show the metrics for validity and discrimination, respectively, of different variants of \methodname, and the two valid baselines.
Table \ref{appendix:table:exp:real:qm_width} shows the number of widths of PIs by different methods on the QM subtasks. Table \ref{appendix:table:exp:real:qm9_functional_groups} shows the coverage rates for a list of functional groups from the  OPENSMILES project\footnote{\url{http://opensmiles.org/opensmiles.html}}. 
We keep only the subset of data whose original SMILES representation contains the corresponding functional group's SMILES representation, and compute the average coverage rate for each of the twelve targets of QM9 dataset\footnote{We only did this for QM9 because the size of QM8 is not enough for this task.}.
If \methodname is actually conditionally valid, then the conditional coverage rate should not be significantly lower than the target (90\%).
Again, it is worth noting that \methodname is only \textit{approximately} conditionally valid, and the raw SMILES functional groups were not used anywhere in the entire pipeline. 
However, \methodname is still almost always valid empirically.

\TableQMDisc
\TableQMCoverage
\TableQMWidth

\TableFunctionalGroups
%================================================Discussion of DJ
\clearpage
%\newpage
\section{Discussion on Discriminative Jackknife}\label{appendix:jkp}
Discriminative Jackknife (DJ) was recently proposed as a post-hoc method to construct prediction intervals for regression deep learning models \cite{pmlr-v119-alaa20a}.
\cite{pmlr-v119-alaa20a} claims that DJ is simultaneously marginally valid and discriminative.
Unfortunately, \textit{neither claim is true}, and it has other practical issues, as we will discuss in detail in this section. 

\subsection{Jackknife+ vs. DJ}
Although it is out-of-scope for this paper, we would like to briefly explain where the finite-sample coverage guarantee comes from, or rather \emph{should have} come from.
It is highly recommended that the readers read the original work of Jackknife+, \cite{Barber10.1214/20-AOS1965} which lays the theoretical foundation for \cite{pmlr-v119-alaa20a} more details.

Suppose we have training data $\{Z_i\}_{i=1}^n$ where $Z_i = (X_i, Y_i)$, and $(X, Y)\sim\mathcal{P}$ for some unknown distribution $\mathcal{P}$.
Suppose we have an \textit{order-invariant} algorithm $\mathcal{A}$ that trains a mean-estimator given some data.
We will denote the full estimator as $\hat{\mu}$, the leave-one-out (LOO) estimator as $\hat{\mu}_{-i}$, and the LOO residual as $R_i^{LOO}$, defined as:
\begin{align}
    \hat{\mu}&\defeq \mathcal{A}\big(\{ (X_{j}, Y_{j})\}_{j\in[n]}\big)\\
    \hat{\mu}_{-i} &\defeq \mathcal{A}\big(\{ (X_{j}, Y_{j})\}_{j\in[n]\setminus\{i\}}\big)\\
    R^{LOO}_i &\defeq |Y_i - \hat{\mu}_{-i}(X_i)|
\end{align}
We will also define $\hat{q}^{+}_{n,\beta}\{v_i\}$ as the $\ceil{(n+1)\beta}$-th smallest (close to the $\beta$-th quantile) of $v_1,\ldots,v_n$, and $\hat{q}^{-}_{n,\beta}\{v_i\}$ as the $\floor{(n+1)\beta}$ smallest value\footnote{Note the $+$ and $-$ signs are used to distinguish the $\ceil{\cdot}$ and $\floor{\cdot}$ operations.}.

The original Jackknife+ \cite{Barber10.1214/20-AOS1965} does the following to construct a PI with finite-sample coverage guarantee (at level $1-2\alpha$, but empirically usually covers $1-\alpha$ of the time):
\begin{itemize}
    \item Step 1: Train the LOO estimator $\hat{\mu}_{-i}$ for $i\in[n]$.
    \item Step 2: Collect the LOO residuals $R^{LOO}_i$ for $i\in[n]$.
    \item Step 3 (inference): For a new data point $(X_{n+1}, Y_{n+1})$, the Jackknife+ PI would be 
    \begin{align}
        \hat{C}_\alpha^{Jackknife+} (X_{n+1}) \defeq [\hat{q}^{-}_{n, \alpha}\{\hat{\mu}_{-i}(X_{n+1}) - R^{LOO}_i\}, \hat{q}^{+}_{n,1-\alpha}\{\hat{\mu}_{-i}(X_{n+1}) + R^{LOO}_i\}]
    \end{align}
\end{itemize}
Assuming exchangeability of $\{Z_i\}_{i=1}^{n+1}$, \cite{Barber10.1214/20-AOS1965} proves that 
\begin{align}
    \mathbb{P}\{Y_{n+1} \in \hat{C}_\alpha^{Jackknife+} (X_{n+1})\} \geq 1-2\alpha
\end{align}
Here the probability is taken over all training samples and the test data.

DJ aims to apply the above for deep learning algorithm $\mathcal{A}$. 
The only difference between DJ and Jackknife+ is that replaces step 1 with step 1b below:
\begin{itemize}
    \item Step 1b: replace $\hat{\mu}_{-i}$ with $\hat{\mu}^{HOIF}_{-i}$, which is estimated using $\hat{\mu}$ and higher-order influence function (HOIF) without actually retraining the deep learning algorithm $\mathcal{A}$.
\end{itemize}

\subsection{Validity}\label{appendix:subsec:DJ:validity}
Although using influence function (IF) to estimate $\hat{\mu}_{-i}$ is possible, in practice, there is almost no way to do this. 
For the coverage guarantee (Theorem 1 in \cite{Barber10.1214/20-AOS1965}) to hold, it is important that for $\hat{\mu}_{-i}$, $Z_i$ and $Z_{n+1}$ are also ``exchangeable''. 
In other words, $\hat{\mu}_{-i}$ cannot see $Z_i$ at all, which is crucial in Step 2 of the proof of Theorem 1 (Section 6 in \cite{Barber10.1214/20-AOS1965}). 
If $\hat{\mu}_{-i}$ actually ``remember'' $Z_i$ somehow, then the last step of Step 2 in the proof breaks.

Unfortunately, $\hat{\mu}^{HOIF}_{-i}$ \textit{does} ``remember'' the $Z_i$ it saw.
The original paper \cite{influence_pmlr-v70-koh17a} also uses IF to estimate the LOO models, but it only applies this to understand which training sample has more influence on the model, or in some qualitative assessment settings (as the name of the paper suggests). 
Even for such use case, \cite{basu2021influence} summarizes several issues with using IF in deep learning, one of which is the error in estimating just the ranking of the influences even with  first order IF estimated with \textit{exact} inverse-Hessian vector product (HVP).
In \cite{pmlr-v119-alaa20a}, the HOIFs are computed recursively, and every IF is computed with \textit{approximate} HVP, which means there is little understanding in the quality of such estimates\footnote{In fact, based on the experiment, the errors seem to build up, as we will discuss in Section \ref{appendix:subsec:DJ:discrimination}.}.
To actually achieve the theoretical guarantee in this setting, we need to eliminate completely the influence of $Z_i$ on the model parameters of $\hat{\mu}$, which requires infinite-order exact IF and is clearly unrealistic.

\subsection{Discrimination}\label{appendix:subsec:DJ:discrimination}
The short answer to this is DJ is actually not discriminative, or at least not in practice.
This can be found in our experiments in Section \ref{sec:exp:real}. 
\cite{pmlr-v119-alaa20a} reports high AUPRC due to a code error\footnote{\url{https://github.com/ahmedmalaa/discriminative-jackknife/blob/e012d0a359aa8dac16fe03a99fa586966cf86ffe/UCI\_experiments.py\#L82}}. 
It is worth noting that the exact version of Jackknife+ does not show discrimination in the way claimed in \cite{pmlr-v119-alaa20a} either (See the comparison in Figure \ref{appendix:fig:exp:dj}).
\FigDJ
The varying width of the PI is a by-product of the construction and proof, and is usually close to constant in practice. 
Intuitively, as $n\to\infty$,  $\hat{\mu}_{-i}\to \hat{\mu}$, and the variance of the PI width would $\to 0$. 
Here are two simple thought experiments:
\begin{enumerate}
    \item As $n\to\infty$,  $\hat{\mu}_{-i}\to \hat{\mu}$, and the variance of the PI width would $\to 0$. 
    \item Suppose $X$ follows a uniform distribution from $\{-10, -9,\ldots,0,\ldots,9,10\}$, and $Y = |X|$. 
    Suppose $\mathcal{A}$ is a linear regression algorithm without intercept. 
    As long as $n$ is big enough, the PI for any input $X_{n+1}$ would be $[-9,9]$.
    The error would however $\to |X_{n+1}|$ (because $\hat{\mu}(x) \to 0$), so there is not discrimination at all.
\end{enumerate}
As a result, DJ, the approximated version, could only potentially be discriminative due to some numerical instability and/or some effect that is orthogonal to the LOO procedure and the construction of the PI, which requires more exploration and detailed explanation. 

\subsection{Other Considerations}
\textbf{Order-invariance for $\mathcal{A}$} is rarely satisfied for the deep learning model.
This is because a deep learning model usually uses some variants of stochastic gradient descent (SGD) instead of gradient descent, which means permuting the input data would result in different $\hat{\mu}$.
However, it is also required for the proof in \cite{Barber10.1214/20-AOS1965}.
\cite{pmlr-v119-alaa20a} did not mention this at all, which results in an incomplete proof even if every stated above is fixed.
That said, the proof \cite{Barber10.1214/20-AOS1965} could easily be extended to training DNN with SGD as well; however it is out of the scope of this discussion.

\textbf{Scalability} of the proposed method in DJ is not practical, even the employed approximations. 
At training time, at least for the experiments in \cite{pmlr-v119-alaa20a}, directly performing the LOO procedure is faster than actually computing influence functions and estimating $\hat{\mu}_{-i}$. 
This of course depends on the number of training data points vs. the number of parameters of the DNN. 
However, as we will discuss in Section \ref{appendix:subsec:DJ:conclusion}, there is no strong argument for using DJ in any scenario.
Moreover, if we do not store all the LOO model weights (which has a large space requirement), we would need to compute the IFs on the fly for each test data, which is prohibitively expensive.

\textbf{Stability} is another concern. 
In using the influence function, inverting Hessian is very expensive, so DJ follows \cite{influence_pmlr-v70-koh17a} in using a stochastic Hessian Vector Product (HVP) method.
However, one would also need to get a good estimate of the eigenvalue of the Hessian\footnote{which can be very large and thus unstable to estimate according to \cite{basu2021influence}} for the HVP estimation process to converge meaningfully.
In our experiment (and in the code published by the authors of \cite{pmlr-v119-alaa20a}), exact Hessian with small NNs have to be used, instead of HVP, due to stability issues. 

\subsection{Conclusion}\label{appendix:subsec:DJ:conclusion}
If we take a step back, Jackknife+ was proposed as an improved version of the classical Jackknife with a finite-sample marginal coverage guarantee.
The question it tries to address however is not just concerning finite-sample marginal coverage, but also about data scarcity: 
As noted in the original Jackknife+ paper \cite{Barber10.1214/20-AOS1965}, split conformal already has a finite-sample guarantee (at $1-\alpha$ level as opposed to $1-2\alpha$ of Jackknife+), but the limitation is that it requires reserving a hold-out set.
When the model requires more data to train, this might result in a poor fit. 
Of course, it is desirable to use all the data we have to train the base model.
However, in many cases we only need a small portion of the data as the validation/calibration set.
If data is abundant, this is not a concern, so one could use split conformal (or CQR, MADSplit, \methodname, etc.).
If the data is actually very scarce, then usually the model cannot be too complicated, so directly performing the LOO cross-validation with Jackknife+ would not be too expensive and will keep the theoretical guarantee.
If we use DJ, we might spend more time while breaking the theoretical guarantee.

\newpage
\section{Data}
In this section, we will try our best to list the licenses of the public datasets we use and details about how the consent was obtained.

\begin{itemize}
    \item UCI Yacht Hydrodynamics (Yacht)\cite{UCI_Yacht}: 
    We could not find the license for this dataset. 
    The dataset was created by ``Ship Hydromechanics Laboratory, Maritime and Transport Technology Department, Technical University of Delft'', and donated by ``Dr Roberto Lopez'' per \cite{UCI_Yacht}.
    
    \item UCI Bikesharing (Bike) \cite{UCI_Bike, UCI_Bike_paper}: 
    The original data was provided according to the Capital Bikeshare Data License Agreement \url{https://www.capitalbikeshare.com/data-license-agreement}.
    We could not find details on how the data was obtained.

    \item UCI Energy Efficiency (Energy)\cite{UCI_Energy,UCI_Energy_paper_TSANAS2012560}:
    We could not find the license for this dataset. 
    The dataset was created by Angeliki Xifara (angxifara '@' gmail.com, Civil/Structural Engineer) and was processed by Athanasios Tsanas (tsanasthanasis '@' gmail.com, Oxford Centre for Industrial and Applied Mathematics, University of Oxford, UK).
    
    \item UCI Concrete Compressive Strength (Concrete)\cite{UCI_Concrete,UCI_Concrete_paper_YEH19981797}:
    We could not find the license for this dataset. 
    The dataset was original owned and donated by Prof. I-Cheng Yeh at Department of Information Management at Chung-Hua University, Taiwan, R.O.C.

    \item Boston Housing (Housing)\cite{Housing_data}: 
    We could not find the license for this dataset. 
    This dataset contains information collected by the U.S Census Service concerning housing in the area of Boston Mass\footnote{https://www.cs.toronto.edu/~delve/data/boston/bostonDetail.html}.
    
    \item Kin8nm\cite{Kin8nm}:
    We could not find the license for this dataset. 
    The original parent dataset (the ``kin'' dataset) was contributed by Zoubin Ghahramani\footnote{https://www.cs.toronto.edu/~delve/data/kin/desc.html}.

    \item QM8 \cite{QM8_doi:10.1063/1.4928757,QM9_1_doi:10.1021/ci300415d} and QM9 \cite{QM9_1_doi:10.1021/ci300415d, QM9_2_ramakrishnan2014quantum}:
    We could not find the original license for these datasets, but they are discributed under CC By 4.0 \footnote{https://tdcommons.ai/single\_pred\_tasks/qm/}.
    They are obtained in \cite{QM8_doi:10.1063/1.4928757} and \cite{QM9_2_ramakrishnan2014quantum}.
    
\end{itemize}
\end{document}